\documentclass[runningheads]{llncs}

\usepackage{eccv}

\usepackage{eccvabbrv}
\usepackage{graphicx}
\usepackage[accsupp]{axessibility}  

\usepackage{hyperref}

\usepackage{orcidlink}
\usepackage{booktabs}
\usepackage{makecell}
\usepackage{multirow}
\usepackage{circledsteps}
\usepackage{tabularx}
\usepackage[table]{xcolor} 
\usepackage[normalem]{ulem}

\begin{document}

\title{RESOLVE: A Multi-Resolution and Multi-Modal Dataset for Roadside Cooperative Perception} 
\titlerunning{RESOLVE: A Multi-Resolution and Multi-Modal Roadside Dataset}

\author{Shaozu Ding\inst{1}\orcidlink{0009-0000-0720-176X} \and
Linan Song\inst{1}\orcidlink{0009-0004-4617-6295} \and
Marco De Vincenzi\inst{1,2}\orcidlink{0000-0002-2706-2936} \and
Dajiang Suo\inst{1}\orcidlink{0000-0003-3748-6115}\thanks{Corresponding author.}}

\authorrunning{S.~Ding et al.}

\institute{The Polytechnic School, Arizona State University, USA \\
\email{\{sding32,lsong124,mdevinc2,dsuo2\}@asu.edu}
\and
Department of Computer Science and Engineering, New York University, USA
}
\maketitle

\begin{abstract}
LiDAR has increasingly been integrated into traffic cameras to expand coverage and mitigate occlusion in roadside cooperative perception. However, how unimodal and camera–LiDAR fusion architectures behave under variations in LiDAR point sparsity induced by sensor configurations and scene-dependent sensing conditions remains underexplored. We introduce RESOLVE, a large-scale real-world benchmark dataset featuring multi-resolution roadside LiDAR and synchronized camera-LiDAR sensing for systematic evaluation of unimodal and fusion-based architectures in roadside 3D detection and tracking. RESOLVE contains over 100k images and 26k point cloud frames with 220k manually annotated bounding boxes, captured at a real-world urban intersection across diverse lighting and weather conditions and spanning 10 classes of traffic participants. In particular, RESOLVE enables controlled evaluation across three LiDAR resolution levels while keeping all other sensing and environmental factors fixed. This allows fair cross-architecture comparisons under point cloud distribution shifts resulting from resolution variations, sensing distance, and training–inference resolution mismatches. Results from extensive benchmark experiments reveal insights into how multimodal fusion can compensate for LiDAR point sparsity, offering clues for designing cost-efficient roadside multimodal perception. The dataset and benchmark codes are available at \url{https://github.com/ASU-Suo-Lab/RESOLVE}. 
\end{abstract}

\begin{figure}[t]
  \centering
  \includegraphics[width=\linewidth]{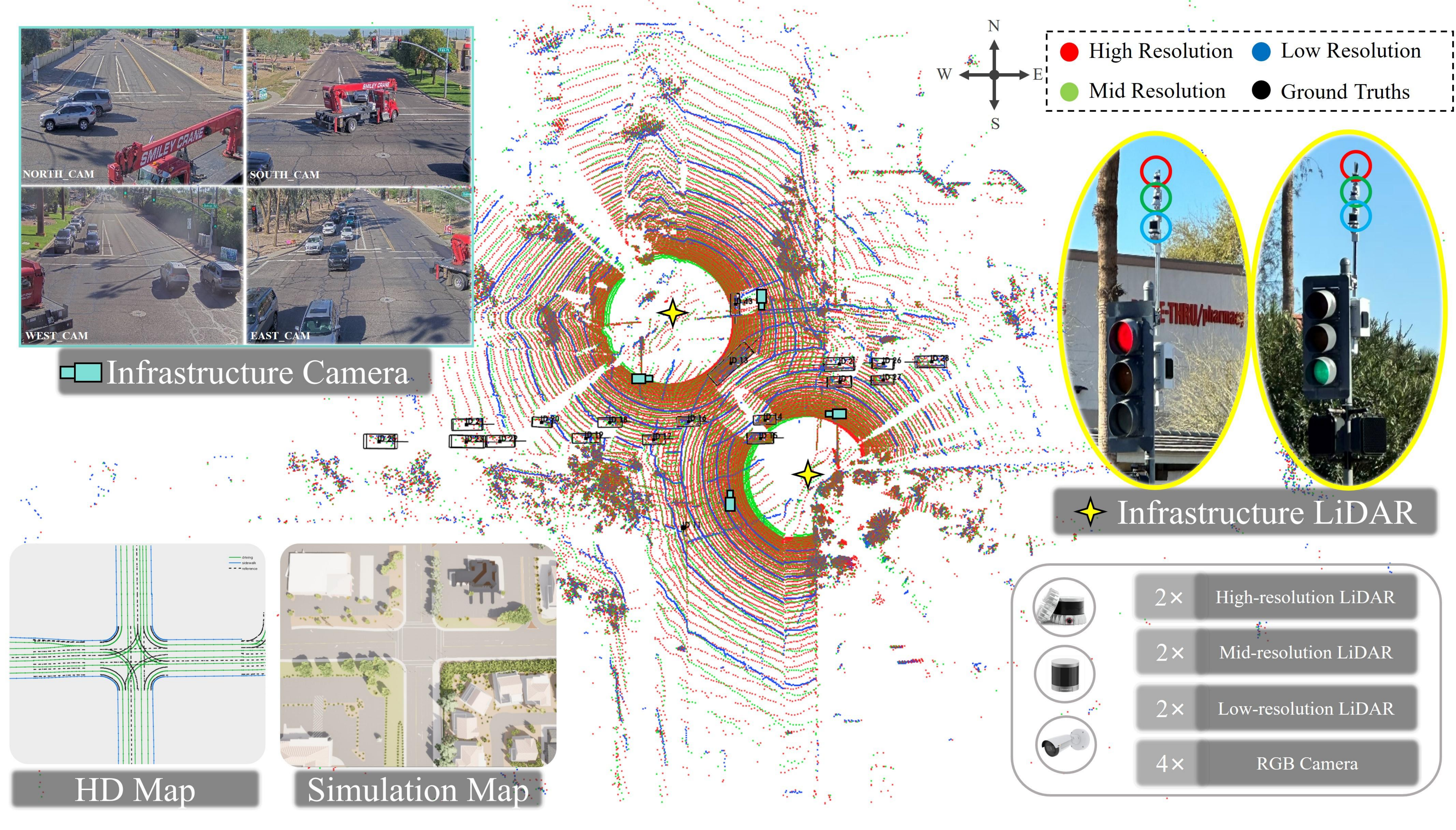}
  \caption{This illustration provides an comprehensive overview of our RESOLVE dataset. We visualize globally fused point clouds of a representative scene, with point clouds from LiDARs of different resolutions encoded in different colors. The sensor installation locations on the infrastructure side are highlighted. Both the high-fidelity simulation map based on CARLA and the HD map is displayed.
  }
  \label{fig:overview}
\end{figure}

\section{Introduction}
\label{sec:intro}
In the context of intelligent transportation systems (ITS)~\cite{its_roadside} and autonomous driving~\cite{av_roadside}, cooperative perception refers to merging data from distributed sensors (e.g., Camera, LiDAR) to mitigate occlusions and expand sensing coverage. In parallel with vehicle-centric cooperative perception, which relies on vehicle-to-vehicle (V2V)~\cite{CoCa3D} or vehicle-to-infrastructure (V2I)~\cite{pillargrid,VINet} data sharing, another paradigm focuses on sharing data among infrastructure-mounted sensors through infrastructure-to-infrastructure (I2I)~\cite{i2iperception} communication mechanisms. In recent years, this infrastructure-centric roadside cooperative perception has emerged as an effective approach to improving traffic safety and operational efficiency in complex urban environments, which is the focus of this paper.

However, existing roadside cooperative perception datasets~\cite{urbanv2x,rcooper} are collected under a fixed LiDAR configuration. In real-world settings, LiDAR is an emerging technology that is incrementally integrated into existing traffic camera-based infrastructure. Within the lifecycle of a single deployment site such as an intersection, LiDAR configurations can vary due to budget constraints, hardware upgrades, procurement cycles, and installation conditions~\cite{SEIP,lidar_placement,lidar_place_infra}. In addition, LiDAR measurements become sparser as objects move farther from the sensing location due to LiDAR's angular sampling characteristics~\cite{Range_based,BeamDeployment}. These variations lead to shifts in the point density and geometric representation, creating point distribution changes even when the underlying scene remains the same.

As a result, current state-of-the-art LiDAR perception~\cite{lion,dsvt} and camera-LiDAR fusion models~\cite{bevfusion,unitr} are primarily evaluated under matched sensing sparsity, since existing datasets do not provide controlled data to systematically investigate perception behavior under different LiDAR sensing sparsity. In general, it is still difficult to analyze how roadside perception performance varies across different LiDAR resolution, sensing geometry and distance, and under potential resolution mismatches between training and inference. 

To address this gap, we propose RESOLVE (Fig.~\ref{fig:overview}), the first real-world roadside cooperative perception dataset featuring synchronized multi-resolution LiDAR and camera data. Unlike prior datasets with fixed sensor configurations, RESOLVE allows controlled evaluation by providing multiple LiDAR resolution settings while keeping other sensing and environmental factors consistent.

Our contributions are as follows:
\begin{itemize}
    \item We introduce RESOLVE, a large-scale multi-resolution roadside cooperative perception dataset supporting infrastructure-based 3D object detection and tracking. It includes over 20,000 synchronized frames of images from traffic cameras covering four intersection directions and point clouds captured by six LiDARs with three resolution levels at two diagonal intersection corners. The dataset also provides 220k expert-annotated 3D bounding boxes across 10 categories of traffic participants, along with simulation assets and HD maps for reproducibility and future extensions.
    \item We present extensive experiments with benchmarks for evaluating unimodal LiDAR architectures and multimodal LiDAR–camera fusion models under varying LiDAR resolutions, object detecting distances, and resolution mismatches between training and inference, enabling controlled studies of unimodal performance and multimodal perception gain in roadside perception. 
     
    \item The benchmark results reveal several observations about multimodal cooperative perception under sparse LiDAR sensing, including how multimodal fusion may compensate for sensing sparsity and how performance varies with LiDAR resolution, sensing distance, and resolution mismatches between training and inference. This provides insights for future research in designing more robust and cost-efficient perception models under roadside sensing scenarios where sensing resolution, object heterogeneity, and distance-conditioned sparsity must be jointly considered.  
\end{itemize}

\section{Related Work}
Multiple ITS-relevant datasets have been made public by the research community. The first category is designed to facilitate the training of ego-vehicle autonomous driving or individual smart infrastructure for traffic monitoring. In contrast, an increasing number of cooperative perception datasets have emerged, aiming to enhance sensing coverage and mitigate occlusion by enabling data sharing among vehicles, between vehicles and infrastructure, and across distributed infrastructure components. Camera, LiDAR, and Radar have been deployed on ego vehicle for multimodal data collection~\cite{kitti,nuscenes,ominiHD,waymo,UnderstandingtheDomainGap}. To improve traffic monitoring performance, infrastructure-mounted sensors have also been used to collect data near intersections~\cite{baai,A9intersection,rope3d,ips300}, campuses~\cite{CORP}, and highways~\cite{zhu2024roscenes}.

To mitigate occlusions and achieve see-through capabilities for single vehicles, sensing data has also been shared among adjacent vehicles~\cite{xu2023v2v4real,cats} and between vehicles and infrastructure~\cite{dairv2x,tumtrafv2x,urbanv2x,holovic,v2xradar}. Similarly, research has focused on merging data from distributed infrastructure sensors~\cite{rcooper,zhu2024roscenes,v2xreal,urbanv2x} to improve sensing coverage and mitigate dead zones at the corners of individual sensors.

Among those involving infrastructure-to-infrastructure sensing data sharing, RESOLVE is the first real-world multimodal dataset with multi-resolution LiDAR and controlled deployment setup for both traffic camera and LiDAR sensors, enabling resolution-isolated cross-architecture comparison under point cloud distribution shifts. A summary of these datasets is given in \Cref{tab:liteture_review}.

\begin{table}[t]
\centering
\scriptsize
\setlength{\tabcolsep}{1.4pt}
\renewcommand{\arraystretch}{1.3}
\caption{A comprehensive comparison of real-world autonomous driving, V2X, and roadside datasets with our RESOLVE dataset. V: Vehicle-only, I: Infrastructure-only, V2V: Vehicle-to-Vehicle, V2I: Vehicle-to-Infrastructure, I2I: Infrastructure-to-Infrastructure, Mixed: Corridor \& Intersection, LiDAR Res.: LiDAR Resolution.}
\label{tab:liteture_review}
\begin{tabularx}{\textwidth}{@{} l c c c c c c c c c @{}}
\toprule
Dataset & Year & Type & Scene & \makecell{\# of\\LiDAR Res.} & \makecell{LiDAR Res.\\Impact} &
\makecell{Class} &
\makecell{Images} & \makecell{LiDAR\\Frames} & \makecell{3D\\Boxes} \\
\midrule
KITTI~\cite{kitti}        & 2012 & V  & Driving  & 1 & -- & 8  & 15k   & 15k   & 200k \\
nuScenes~\cite{nuscenes}     & 2019 & V & Driving & 1  &-- & 23 & 1.4M  & 400k  & 1.4M \\
Waymo Open~\cite{waymo}   & 2019 & V & Driving  & 2& -- & 4  & 1M    & 200k  & 12M \\
OmniHD-Scenes~\cite{ominiHD}   & 2024 & V & Driving  & 1 &--  & 4  & --  & 12k  & 514.6k \\
TAD-E2E~\cite{TADE2E} & 2025 & V & Driving  & 2 & -- & --   & -- & 1M  & -- \\
\midrule
V2V4Real~\cite{xu2023v2v4real}   & 2022 & V2V & Driving  & 1 & -- & 5  & 40k   & 20k   & 240k \\
DAIR-V2X~\cite{dairv2x}     & 2022 & V2I & Intersection & 2  & --  & 10 & 39k   & 39k   & 464k \\
V2X-Real~\cite{v2xreal}     & 2024 & V2I & Intersection  & 2 & --  & 10 & 171k  & 33k   & 1.2M \\
TUMTraf-V2X~\cite{tumtrafv2x}  & 2024 & V2I & Intersection  & 2 & --   & 8  & 5.0k  & 2k    & 29k \\
HoloVIC~\cite{holovic}  & 2024 & V2I & Intersection & 3 & --   & 3  &--  & 100k    & 11.47M \\
CATS-V2V~\cite{cats}  & 2025 & V2V & Driving  & 1 & --  & 10  & 1.26M  & 60k    & -- \\
UrbanIng-V2X~\cite{urbanv2x} & 2025 & V2I & Intersection & 3 & -- & 13 & 81.6k & 27.2k & 712k \\
V2X-Radar~\cite{v2xradar} & 2025 & V2I & Intersection  & 1 &-- & 5 & 40k & 20k & 350k \\
\midrule
IPS300+~\cite{ips300}      & 2022 & I & Intersection  & 1 &--  & 7  & 14.1k & 14.1k & 4.5M \\
Rope3D~\cite{rope3d}      & 2022 & I & Intersection  & 2 & -- & 13 & 50k   & --    & 1.5M \\
A9-Intersection~\cite{A9intersection}  & 2023 & I & Intersection  & 1&--   & 10 & 4.8k  & 4.8k  & 57.4k \\
CORP~\cite{CORP}         & 2024 & I & Campus  & 2 & -- & 5  & 205k  & 102k  & 215k \\
H-V2X~\cite{h-v2x}   & 2024 & I  & Highway & 0& --  & 4  &  1.9M  & --  & -- \\
RCooper~\cite{rcooper}     & 2024 & I2I &  Mixed  & 2 & --   & 10 & 50k   & 30k   & -- \\
RoScenes~\cite{zhu2024roscenes}         & 2024 & I2I & Highway & 0 & --& 4  & 1.21M  & --  & 21.39M \\
\midrule
RESOLVE (Ours)   & 2025 & I2I & Intersection  & 3 &  Yes & 10 & 100k  & 26k   & 220k \\
\bottomrule
\end{tabularx}
\end{table}

\section{RESOLVE Dataset}
RESOLVE is a multi-resolution, multi-modal roadside dataset for a complex urban intersection. The data covers high-speed and high-density traffic flow and includes diverse weather and lighting conditions. All annotations undergo expert processes and quality control to ensure consistency and accuracy, while low-latency synchronization across multiple sensors is achieved through a unified time base. Based on LiDAR resolutions, the dataset can be divided into three subsets: high, medium, and low, targeting the highest perception accuracy, the balance of performance and cost, and low-latency real-time applications, respectively.
\subsection{Sensor Setup}
\subsubsection{Sensor Specification.} 
Four high-resolution network cameras are installed on traffic light poles facing the east, south, west, and north directions to provide 360° coverage. Six LiDARs are deployed in two symmetric groups, each including 16-, 64-, and 128-beam units. The two groups are mounted on pedestrian signal poles at the northwest and southeast corners. To ensure fairness in subsequent comparative experiments and reduce the impact of pose factors on point cloud density differences, the positions and orientations of each LiDAR group are kept as consistent as possible within feasible limits, except for installation height. Edge computing servers and other modules used for data collection and synchronization are installed in a roadside cabinet next to the pedestrian pole near the southeast pole. Detailed parameters of each sensor are shown in \cref{tab:sensor_specs}.

\begin{table}[t]
\centering
\caption{Key Sensor Specifications in RESOLVE Dataset.}
\label{tab:sensor_specs}
\small
\setlength{\tabcolsep}{4pt}
\renewcommand{\arraystretch}{1.2}
\begin{tabularx}{\textwidth}{@{} l l X @{}}
\toprule
Sensor & Sensor Model &Details \\
\midrule
      & Ouster OS1-128 ($\times$2) &
128 beams, 10\,Hz, 360$^\circ$ horizontal FOV, $-22.5^\circ$ to $+22.5^\circ$ vertical FOV, $\le$ 90\,m range at $\ge$ 10\% reflectivity. \\
LiDAR & Ouster OS1-64 ($\times$2) &
64 beams, 10\,Hz, 360$^\circ$ horizontal FOV, $-22.5^\circ$ to $+22.5^\circ$ vertical FOV, $\le$ 90\,m range at $\ge$ 10\% reflectivity. \\
      & RoboSense Helios-16 ($\times$2) &
16 beams, 10\,Hz, 360$^\circ$ horizontal FOV, $-15.0^\circ$ to $+15.0^\circ$ vertical FOV, $\le$ 110\,m range at $\ge$ 10\% reflectivity. \\
\midrule
Camera & AXIS P1455-LE ($\times$4) &
RGB, 30\,Hz, 1920$\times$1080 resolution, 114$^\circ$ horizontal FOV, 58$^\circ$ vertical FOV. \\
\bottomrule
\end{tabularx}
\end{table}

\begin{figure}[t]
  \centering
  \begin{subfigure}{\linewidth}
    \centering
    \includegraphics[width=\linewidth]{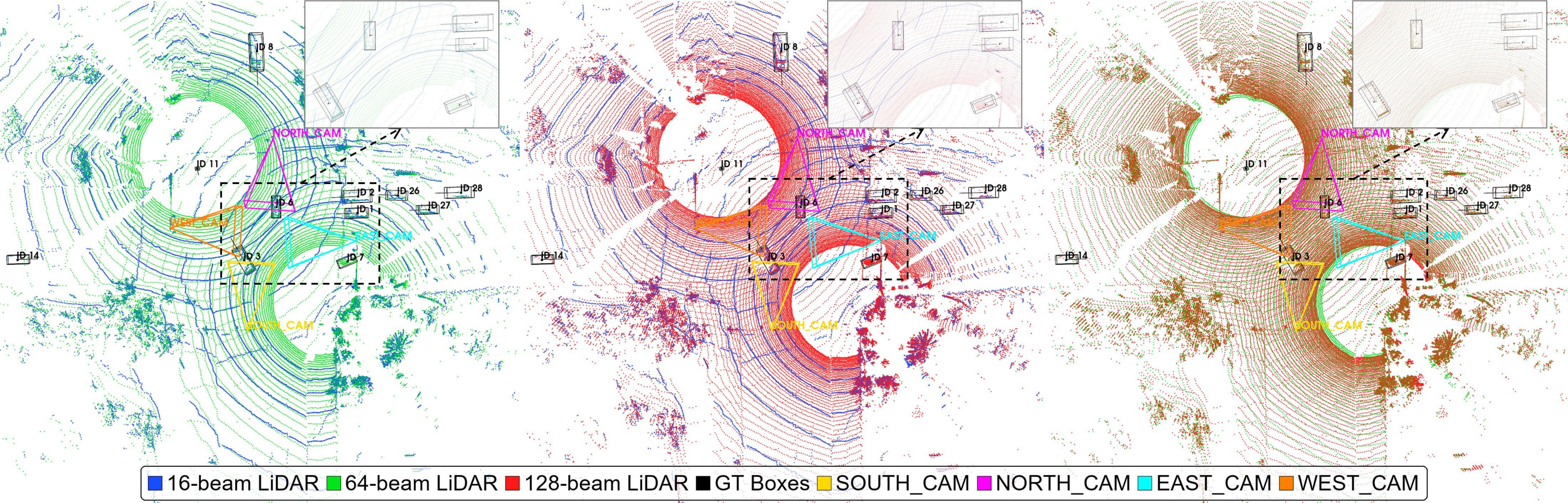}
    \caption{Calibration results between LiDAR and LiDAR}
    \label{fig:lidar_lidar_calib}
  \end{subfigure}

  \begin{subfigure}{\linewidth}
    \centering
    \includegraphics[width=\linewidth]{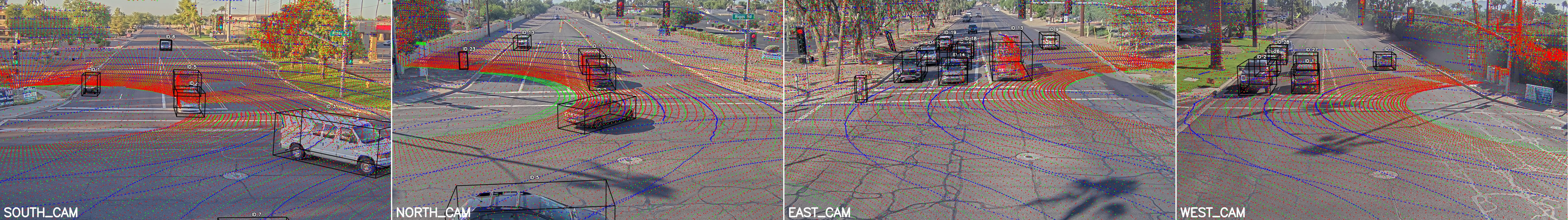}
    \caption{Calibration results between camera and LiDAR}
    \label{fig:lidar_cam_calib}
  \end{subfigure}
  \caption{The illustrations of temporal synchronization and spatial calibration across modalities. (a) shows calibration among LiDARs at different resolutions, including a zoomed-in view of target alignment, with camera frustums overlaid to verify camera–LiDAR extrinsics and coverage. (b) shows point clouds from different resolution LiDARs projected onto multiple camera views to assess cross-modal calibration.}
  \label{fig:calibration}
\end{figure}

\subsubsection{Sensor Synchronization.} Time synchronization of multimodal datasets is particularly critical. We use the UTC clock as the global time reference and configure the edge computer as the master clock of Precision Time Protocol (PTP)~\cite{PTP}. Each LiDAR is connected to the network through an industrial-level switch and receives synchronization signals as a slave node. Each camera uses Network Time Protocol (NTP) to synchronize with the time server on the edge computer. Based on time synchronization, we further coordinate and control the acquisition trigger of the data stream. The LiDAR uses a phase-locked loop mechanism to align the rotation cycles of each device, ensuring that the scanning cycle starts and ends within the same time window. Since Axis network cameras do not support externally triggered hardware-level synchronization, we perform cross-modal time registration in post-processing, matching each LiDAR frame with the closest camera frame in time, ensuring that the maximum time deviation does not exceed half of the camera frame interval, and discarding matching pairs that exceed the threshold to guarantee stable synchronization quality.
\subsubsection{Sensor Calibration.} After time synchronization, we perform spatial calibration for all sensors in Fig.~\ref{fig:calibration}. We estimate the intrinsics and distortion coefficients of each camera using a checkerboard~\cite{cam_intrinsic} under a standard pinhole model. For camera-LiDAR extrinsic calibration, we select multiple sets of 2D-3D corresponding points in the point cloud and image, and calculate the transformation matrix minimizing the reprojection error~\cite{cam_lidar_Ext}. Furthermore, we select several 3D corresponding points in different LiDAR point clouds, use the least squares criterion to minimize the point-to-point distance to obtain an initial coarse transformation, and then use the Iterative Closest Point (ICP) algorithm~\cite{lidar_lidar_ext} for fine registration, thus completing the spatial calibration from LiDAR to LiDAR.

\begin{figure}[t]
  \centering
  \begin{subfigure}[t]{0.32\linewidth}
    \centering
    \includegraphics[width=\linewidth,height=2.5cm]{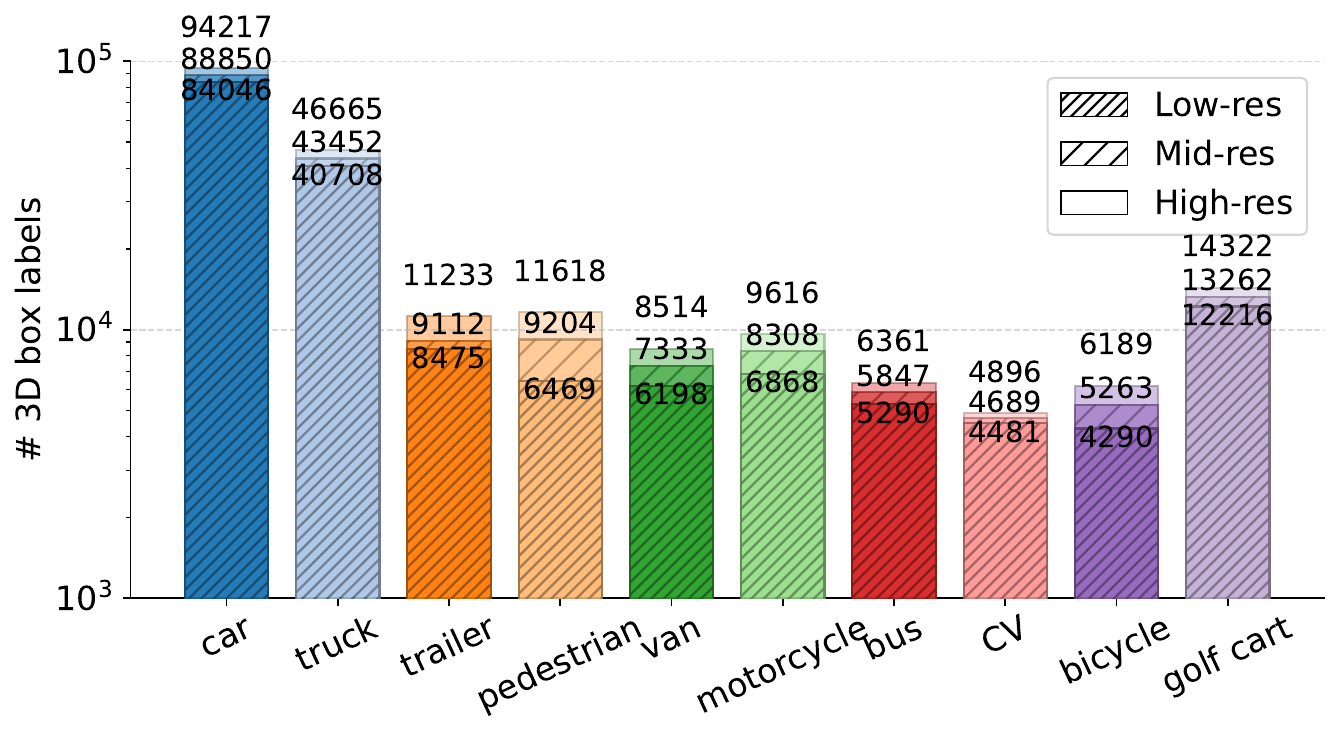}
    \caption{Valid 3D boxes per class}
    \label{fig:label_class}
  \end{subfigure}\hfill
  \begin{subfigure}[t]{0.32\linewidth}
    \centering
    \includegraphics[width=\linewidth,height=2.5cm]{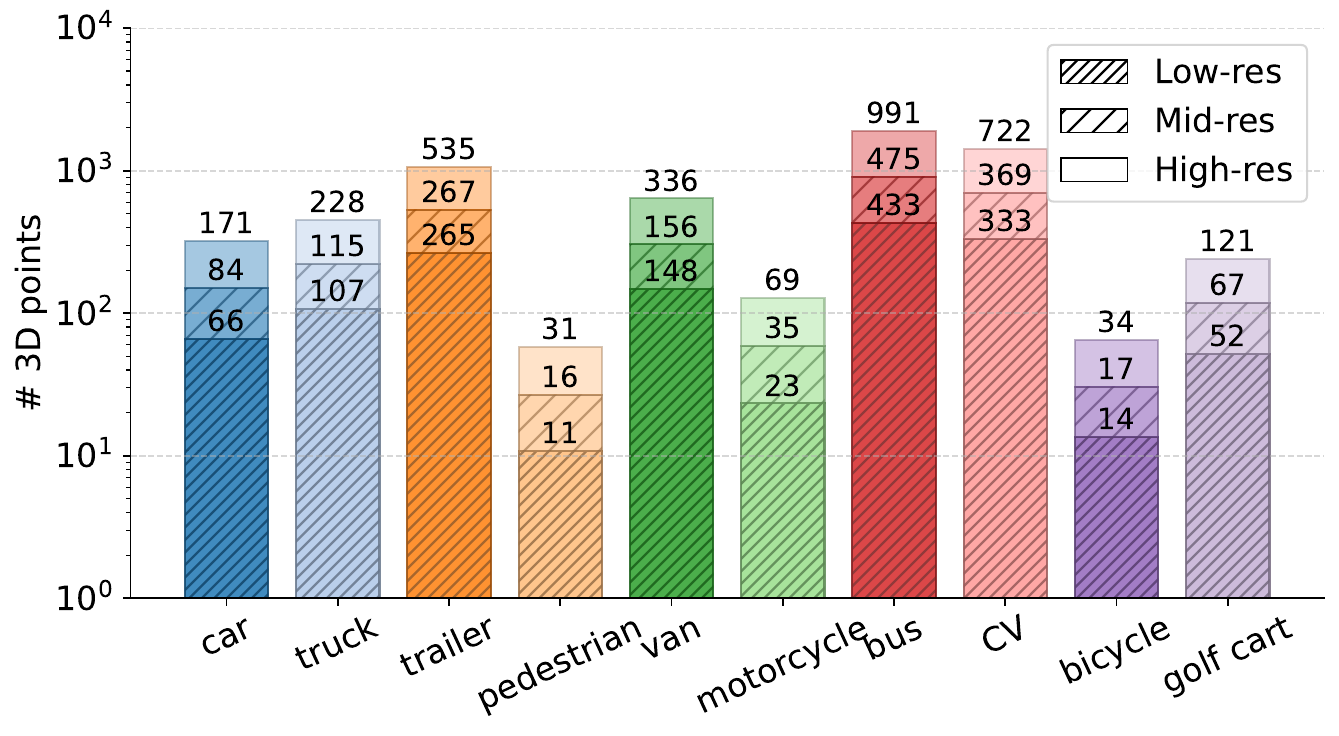}
    \caption{Avg. num. of LiDAR points}
    \label{fig:point_class}
  \end{subfigure}\hfill
  \begin{subfigure}[t]{0.32\linewidth}
    \centering
    \includegraphics[width=\linewidth,height=2.5cm]{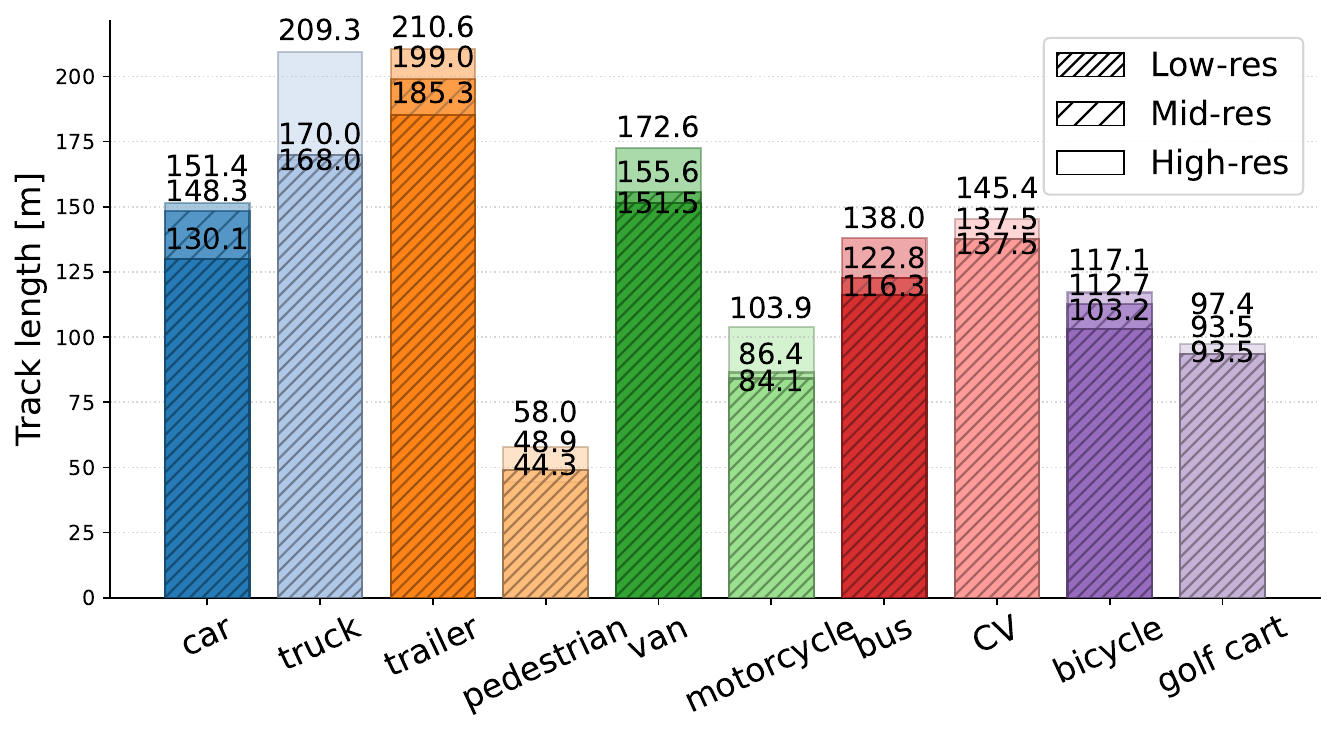}
    \caption{Max. track length per class}
    \label{fig:track_class}
  \end{subfigure}
  \caption{Class-wise statistics under three LiDAR resolutions. (a) Lower resolution reduces valid 3D boxes, especially for small targets. (b) As resolution increases, the average number of points increases, reflecting richer geometric observations. (c) Higher resolution LiDAR provides a slight gain for small targets track, but can extend the effective tracking length of large vehicles.}
  \label{fig:class_analysis}
\end{figure}

\begin{figure}[t]
  \centering
  \begin{subfigure}[t]{0.32\linewidth}
    \centering
    \includegraphics[width=\linewidth,height=3cm]{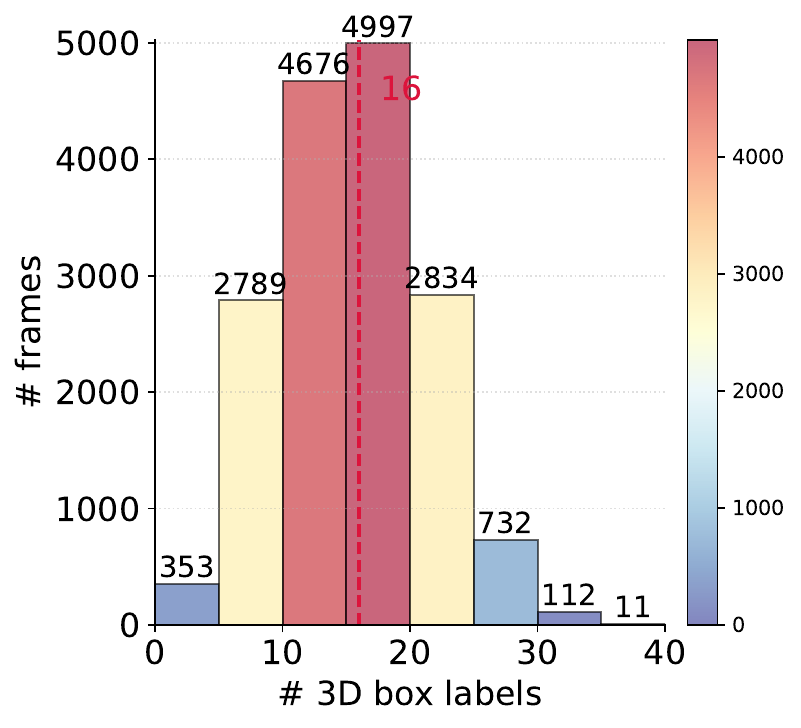}
    \caption{Valid 3D boxes per frame}
    \label{fig:label_frame}
  \end{subfigure}\hfill
  \begin{subfigure}[t]{0.32\linewidth}
    \centering
    \includegraphics[width=\linewidth,height=3cm]{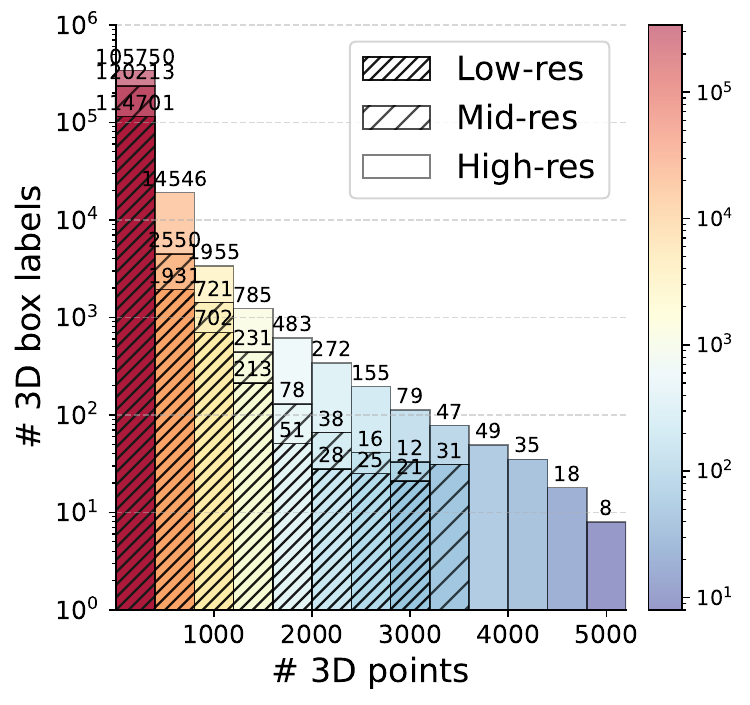}
    \caption{LiDAR points per 3D box}
    \label{fig:label_points}
  \end{subfigure}\hfill
  \begin{subfigure}[t]{0.32\linewidth}
    \centering
    \includegraphics[width=\linewidth,height=3cm]{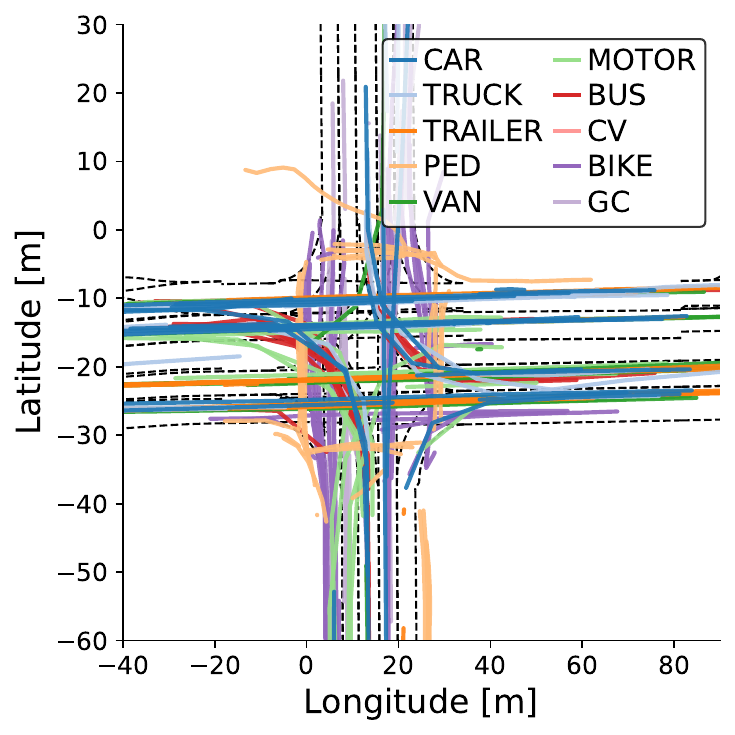}
    \caption{BEV Trajectories per class}
    \label{fig:bev_track}
  \end{subfigure}
  \caption{Statistics on valid boxes per frame and in-box points per box, along with visualization of motion trajectories. (a) On average, each frame contains 16 valid bounding boxes. Resolution variations have minimal impact on this metric. (b) Distribution of LiDAR points per valid 3D box under different resolutions. (c) BEV trajectory visualization on the intersection HD map, with trajectories color-coded by class.}
  \label{fig:label_analysis}
\end{figure}

\begin{figure}[!htb]
  \centering
   \includegraphics[width=\linewidth,height=4cm]{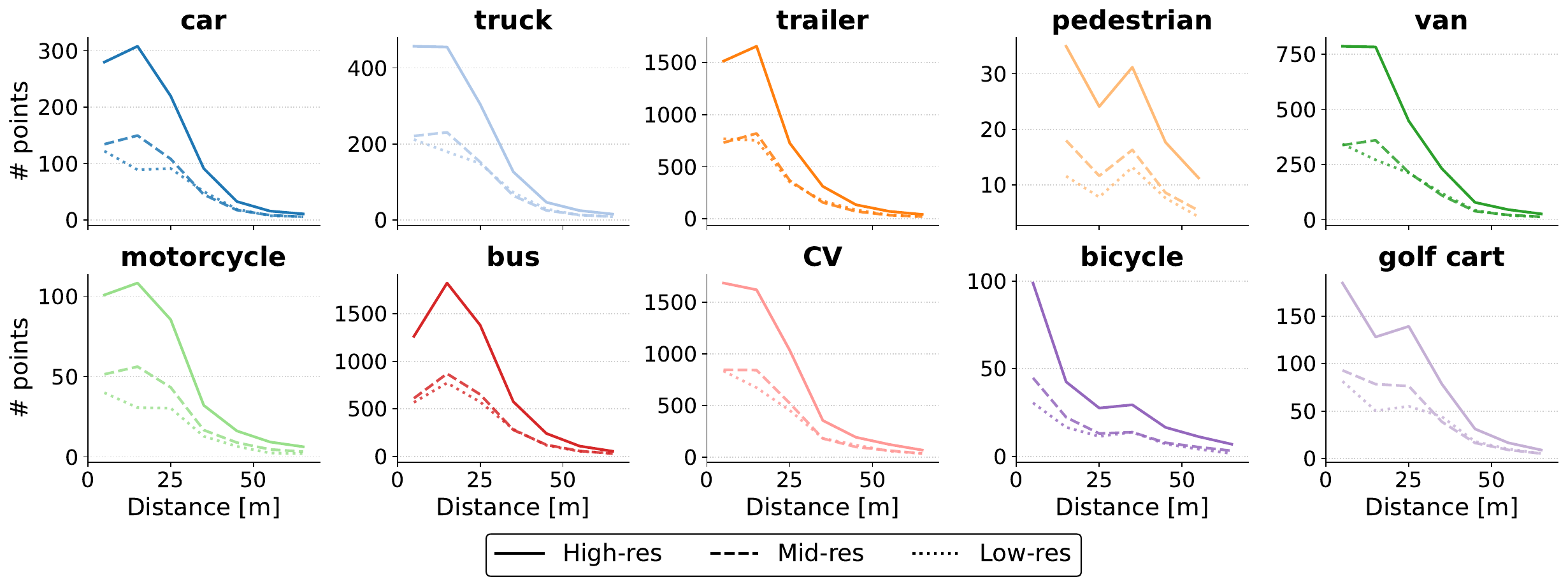}
  \caption{Statistics on the variation of average number of points with distance from object to the center of intersection at different resolutions. The resolution-caused difference is significant at close range but diminishes with increasing distance.}
  \label{fig:points_dist_analysis}
\end{figure}

\subsection{Data Collection and Annotation} 
RESOLVE dataset is designed and built for complex urban intersection scenarios, covering a variety of weather and lighting conditions, including sunny, cloudy, rainy, nighttime and strong backlight, to enhance scene diversity and evaluation challenges. In addition to common vehicle types found at urban intersections, the dataset also includes four categories of vulnerable road users, including golf carts, which are relatively rare in public benchmarks. All identifiable facial and license plate information is anonymized using Gaussian blur~\cite{blur}. The camera and LiDAR are sampled at frequencies of 30Hz and 10Hz respectively and parsed based on ROS2~\cite{ros2}. We use 10Hz as the annotation frequency to provide accurate 3D bounding boxes for each target, and assign a unique tracking ID to the target in each sequence to support temporal tasks. We also provide a format conversion tool that can convert this dataset into nuScenes~\cite{nuscenes} format, making it easy to be directly compatible and reused with mainstream detection framework OpenPCDet~\cite{openpcdet2020} and MMDetection3D~\cite{mmdet3d2020}.

\subsection{Dataset Analysis} 
In roadside sensing, low-resolution point clouds are often too sparse for small, distant, or occluded targets, weakening detection and annotation utility. We therefore deem a 3D box valid only if it contains at least five LiDAR points. However, this definition does not alter the ground truth set used for training or evaluation. Since we employ a consistent point cloud range and identical high-resolution annotations across three resolutions, 3D boxes containing fewer than five points are retained for training and evaluation even in low- or medium-resolution settings. \Cref{fig:label_class} shows that the number of valid 3D boxes decreases as the resolution decreases, indicating that the lower point density directly reduces the amount of usable supervision and the coverage of the evaluation targets. \Cref{fig:point_class} shows that 128-beam LiDAR yields much denser in-box points than 64-beam, while the gain from 16 to 64 is smaller, indicating non-linear density scaling in this intersection setup. The tracking results in \cref{fig:track_class} further show a limited track length for small targets due to the decay of long-range point density, with marginal benefit from higher resolution. In contrast, for large vehicles, higher resolution provides richer geometry and typically extends the track length by about 20\,m.

As shown in \cref{fig:label_frame}, a higher LiDAR resolution yields more valid boxes per frame, improving annotation usability and coverage. \Cref{fig:label_points} further shows that in-box point counts shift upwards at high resolution with a heavier upper tail, confirming that higher resolution can provide richer geometric information for some targets. bird's-eye-view (BEV) visualization in~\cref{fig:bev_track} presents the spatial layout of traffic flow, the spatial distribution of categories, and typical motion behaviors in the scene. As shown in~\cref{fig:points_dist_analysis}, the overall trend of the number of in-box points changing with the distance from the target to the intersection center is consistent across three resolutions. Different classes show slight differences due to variations in spatial distribution and motion patterns.

\section{Tasks}
\subsection{3D Object Detection}
We group representative 3D object detection architectures by backbone operator, covering unimodal LiDAR detectors based on sparse convolution~\cite{second}, transformer~\cite{transformer}, and mamba~\cite{mamba}, as well as multimodal detectors that fuse LiDAR with the camera branch. We benchmark SOTA methods on our RESOLVE dataset using the nuScenes protocol~\cite{nuscenes} over 10 classes, reporting mean Average Precision (mAP) and nuScenes Detection Score (NDS). Since attributes are unavailable, we exclude mAAE and recompute NDS accordingly as in~\cref{eq:nds_ours}.

\begin{equation}
\mathrm{NDS}_{\text{ours}}
=\frac{1}{9}\left[
5\,\mathrm{mAP}
+\sum_{m\in\{\mathrm{mATE},\mathrm{mASE},\mathrm{mAOE},\mathrm{mAVE}\}}
\max\!\bigl(1-m,\,0\bigr)
\right]
\label{eq:nds_ours}
\end{equation}

\subsection{3D Multi-object Tracking}
Multi-object tracking is commonly categorized into tracking-by-detection (TBD) and joint detection and tracking (JDT). TBD detects objects per frame and links them across time via data association (e.g., the Hungarian algorithm~\cite{Hungarian}) with track management and is therefore largely limited by detector quality. JDT jointly learns detection and tracking within a single network, offering tighter temporal modeling but higher complexity. In this work, we focus on TBD and evaluate several classic TBD trackers under different LiDAR resolutions using the nuScenes tracking protocol, reporting Average Multiple Object Tracking Accuracy (AMOTA) and Average Multiple Object Tracking Precision (AMOTP).

\subsection{Cooperative Perception}
Cooperative perception improves 3D detection by enabling multiple agents to share complementary observations. In this work, we treat two diagonally deployed infrastructure sensor suites as distinct agents and categorize cooperative perception methods by their fusion stage, including no fusion, early fusion, late fusion, and intermediate fusion~\cite{F-cooper,CoBEVT,V2X-ViT,OPV2V}, and report mAP over 10 categories following the nuScenes protocol~\cite{nuscenes}.

\section{Experiments}
\subsection{Implementation Details}
\subsubsection{3D Object Detection.} We split the dataset into training, validation, and test sets in a 7:2:1 ratio. The point cloud is restricted to the range of $x,y\in[-60m,88.8m]$ and $z\in[-8m,0m]$, with a voxel size of $[0.3m,0.3m]$. The image size of the fusion models is set to $256\times704$. All models are trained for 20 epochs with a batch size of 4 per GPU, using the adam optimizer and a onecycle scheduler with a learning rate of 0.001. The same training protocol is applied to three LiDAR resolution settings to ensure fair comparisons. The experiments are conducted on four NVIDIA RTX PRO 6000 BLACKWELL GPUs and evaluated on a single GPU. All experiments are implemented in OpenPCDet~\cite{openpcdet2020}. We have integrated and open-sourced all model implementations in a unified code repository and provided training weights to facilitate further research.
\subsubsection{3D Multi-object Tracking.} Since different TBD methods are tuned for specific benchmarks and even have special settings for different classes, we do not perform exhaustive optimization for each method. Therefore, the performance of some trackers may be lower than the results reported in their original benchmark tests. However, we adopt a uniform evaluation protocol to ensure fair comparisons. Specifically, we fix the distance threshold at 0.4, the threshold scan count at 40, the minimum recall rate at 0.1, and the ID switching time window as unrestricted. We have integrated and open-sourced all tracking models into a unified detection framework so that it can be used seamlessly with various detectors.

\subsubsection{Cooperative Perception.} All experiments are implemented in OpenCOOD~\cite{OPV2V}. To ensure a fair comparison across different cooperative perception paradigms, we employ a unified PointPillars-based backbone, where all collaborative detectors share the same anchor box definitions, focal classification loss, Smooth-L1 regression loss, and NMS threshold as the single-agent PointPillars baseline.

\begin{table}[t]
\centering
\caption{Evaluation results of 3D detection methods on our RESOLVE dataset under three LiDAR resolutions. L: LiDAR, C: Camera, SparseConv: Sparse Convolution.}
\label{tab:detection_result}
\setlength{\tabcolsep}{2.6pt}
\renewcommand{\arraystretch}{1.12}
\footnotesize
\begin{tabularx}{\linewidth}{l c c *{6}{>{\centering\arraybackslash}X}}
\toprule
\multirow{2}{*}{Model} & \multirow{2}{*}{Modality} & 
\multirow{2}{*}{\makecell{LiDAR\\[2pt]Backbone}} &
\multicolumn{3}{c}{mAP\,\mbox{$\uparrow$}} &
\multicolumn{3}{c}{NDS\,\mbox{$\uparrow$}} \\
\cmidrule(lr){4-6}\cmidrule(lr){7-9}
& & & Low & Mid & High & Low & Mid & High \\
\midrule
PointPillars~\cite{pointpillars}   & L   & SparseConv  & 75.1 & 79.7 & 80.6 & 71.2 & 72.6 & 73.6 \\
SECOND~\cite{second}               & L   & SparseConv  & 68.2 & 76.6 & 78.0 & 66.5 & 71.2 & 73.1 \\
CenterPoint~\cite{centerpoint}     & L   & SparseConv  & 79.9 & 86.4 & 87.4 & 73.9 & 79.9 & 80.9 \\
TransFusion-L~\cite{transfusion}   & L   & SparseConv  & 82.5 & 86.6 & 89.1 & 76.3 & 80.1 & 82.4 \\
VoxSeT~\cite{vst}                  & L   & Transformer & 87.4 & 89.1 & 90.1 & 82.9 & 81.6 & 83.1 \\
DSVT~\cite{dsvt}                   & L   & Transformer & 85.9 & 94.1 & 94.6 & 81.5 & 87.0 & 87.5 \\
Voxel Mamba~\cite{voxelmamba}      & L   & Mamba       & 85.7 & 94.9 & 95.4 & 81.7 & 88.5 & 89.1 \\
LION~\cite{lion}                   & L   & Mamba       & 88.1 & 95.1 & 95.9 & 84.2 & 89.8 & 91.1 \\
\midrule
BEVFusion~\cite{bevfusion}         & L$+$C  & SparseConv  & 86.3 & 92.6 & 93.1 & 79.8 & 84.8 & 86.4 \\
UniTR~\cite{unitr}                 & L$+$C  & Transformer & 89.7 & 94.4 & 94.9 & 83.7 & 87.3 & 87.4 \\
\bottomrule
\end{tabularx}
\end{table}

\subsection{Benchmark Results}
\subsubsection{3D Object Detection.} As shown in~\cref{tab:detection_result}, increasing LiDAR resolution from low to medium boosts the mean mAP across all models by about 7.4\%. The gains for SparseConv, Transformer, Mamba, and multimodal methods are about 8.0\%, 5.7\%, 9.3\%, and 6.3\%, respectively. Sparse convolution benefits more from higher intra-voxel point density~\cite{PDV}. Transformer-based models are relatively robust to sparsity due to global interactions, while Mamba models, due to serialization modeling, have higher requirements for token quality. Therefore, when medium resolution brings more continuous spatial structure and more reliable local statistics, their performance improvement is more significant. Multimodal methods benefit from improved cross-modal alignment and complementarity as LiDAR geometry becomes more reliable. When resolution increases further from medium to high, the overall mean mAP increases only by about 1.0\%. Mid-resolution already captures most discriminative geometry, and additional densification yields limited new information. Moreover, encoding stages such as voxelization and sampling can introduce information bottlenecks, preventing extra points from translating into proportional gains.

\begin{table}[t]
\centering
\setlength{\tabcolsep}{2.6pt} 
\renewcommand{\arraystretch}{1.08} 
\caption{Evaluation results of multi-object tracking methods on our RESOLVE dataset under three LiDAR resolutions. We use BEVFusion~\cite{bevfusion} as detector. The best performance is marked in \textcolor{red}{red}. CV, Tra, Motor, Bic, Ped, GC: Construction Vehicle, Trailer, Motorcycle, Bicycle, Pedestrian, Golf Cart. Res.: Resolution.}
\label{tab:tracking_result}
\resizebox{\linewidth}{!}{
\begin{tabular}{l c *{10}{c} *{10}{c}}
\toprule
\multirow{2}{*}{Model} & \multirow{2}{*}{Res.} &
\multicolumn{10}{c}{AMOTA $\uparrow$} &
\multicolumn{10}{c}{AMOTP $\downarrow$} \\
\cmidrule(lr){3-12}\cmidrule(lr){13-22}
& &
Car & Truck & CV & Bus & Tra & Van & Motor & Bic & Ped & GC &
Car & Truck & CV & Bus & Tra & Van & Motor & Bic & Ped & GC \\
\midrule

\multirow{3}{*}{AB3DMOT~\cite{AB3DMOT}} 
& Low   & 47.4 & 48.7 & 62.6 & 56.2 & 40.5 & 52.4 & 7.6 & 37.9 & 54.6 & 47.6 &  44.5 & 51.5 & 65.3 & 56.0 & 113.2 & 64.5 & 164.3 & 120.7 & 67.3 & 29.7 \\
& Mid  & 39.2 & 42.4 & 64.2 & 41.1 & 44.9 & 35.8 & 4.8 & 44.1 & 55.3 & 35.4 &  34.0 & 45.0 & 44.5 & 49.0 & 106.7 & 56.4 & 155.1 & 105.3 & 72.7 & 29.7 \\
& High & 35.0 & 39.4 & 69.1 & 57.1 & 46.8 & 33.0 & 8.0 & 45.4 & 58.7 & 37.7 &  30.9 & 36.5 & 31.3 & 34.4 & 100.6 & 42.3 & 140.4 & 97.1 & 70.6 & 18.7 \\

\multirow{3}{*}{CenterPoint~\cite{centerpoint}} 
& Low  & 88.9 & 82.8 & 87.9 & 92.8 & 74.6 & 69.1 & 76.9 & 93.0 & 78.1 & 92.9 & 32.9 & 41.3 & 50.5 & 49.1 & 60.6 & 29.6 & 69.1 & 32.2 & 51.7 & 26.9 \\
& Mid  & 85.1 & 83.8 & 93.1 & 93.5 & 83.3 & 70.5 & 82.1 & 93.3 & 71.9 & 94.8 & 31.0 & 37.6 & 39.1 & 36.4 & 44.8 & 33.0 & 46.2 & 23.3 & 70.7 & 22.1 \\
& High & \textcolor{red}{\uline{91.9}} & 87.1 & 94.0 & 95.8 & \textcolor{red}{\uline{90.0}} & 66.8 & \textcolor{red}{\uline{92.0}} & 99.3 & \textcolor{red}{\uline{92.7}} & 96.7 & \textcolor{red}{\uline{22.0}} & 32.6 & 25.2 & \textcolor{red}{\uline{25.7}} & 32.8 & \textcolor{red}{\uline{28.9}} & \textcolor{red}{\uline{33.9}} & 15.0 & \textcolor{red}{\uline{27.6}} & \textcolor{red}{\uline{15.3}} \\

\multirow{3}{*}{SimpleTrack~\cite{simpletrack}} 
& Low  & 88.5 & 85.6 & 92.6 & 94.1 & 84.2 & 71.9 & 52.8 & 95.3 & 73.4 & \textcolor{red}{\uline{97.8}} & 34.8 & 42.2 & 42.4 & 48.2 & 55.3 & 34.9 & 70.3 & 29.1 & 58.1 & 20.8 \\
& Mid  & 87.2 & 85.2 & 97.0 & \textcolor{red}{\uline{97.1}} & 89.5 & 74.6 & 53.1 & 98.4 & 77.8 & 96.0 & 34.2 & 40.4 & 31.2 & 35.7 & 39.3 & 38.9 & 50.5 & 18.0 & 68.5 & 21.8 \\
& High & 90.0 & \textcolor{red}{\uline{88.4}} & \textcolor{red}{\uline{97.1}} & 97.0 & 89.8 & \textcolor{red}{\uline{74.9}} & 56.7 & \textcolor{red}{\uline{99.7}} & 82.9 & 96.8 & 27.0 & \textcolor{red}{\uline{31.8}} & \textcolor{red}{\uline{20.6}} & 27.6 & \textcolor{red}{\uline{31.7}} & 30.5 & 39.2 & \textcolor{red}{\uline{13.4}} & 44.5 & 16.7 \\

\multirow{3}{*}{Poly-MOT~\cite{polymot}} 
& Low  & 79.6 & 72.6 & 65.9 & 77.6 & 70.0 & 74.2 & 28.2 & 70.8 & 90.8 & 80.4 & 33.1 & 43.6 & 72.9 & 68.2 & 48.4 & 48.3 & 81.1 & 36.9 & 31.9 & 23.3 \\
& Mid  & 69.6 & 63.5 & 61.3 & 64.6 & 68.8 & 65.3 & 39.3 & 81.2 & 74.7 & 75.7 & 45.2 & 53.8 & 89.1 & 79.3 & 49.1 & 62.3 & 66.6 & 36.7 & 77.8 & 26.5 \\
& High & 69.6 & 65.0 & 67.2 & 80.1 & 70.1 & 72.2 & 44.6 & 84.8 & 74.7 & 79.0 & 38.4 & 41.6 & 74.1 & 63.4 & 38.6 & 45.6 & 47.4 & 31.0 & 72.0 & 15.8 \\

\multirow{3}{*}{MCTrack~\cite{mctrack}} 
& Low  & 70.3 & 63.1 & 60.9 & 62.8 & 60.5 & 62.8 & 26.0 & 45.9 & 72.0 & 78.7 & 41.9 & 63.3 & 93.6 & 91.2 & 74.6 & 99.9 & 147.7 & 115.9 & 70.3 & 51.7 \\
& Mid  & 66.9 & 63.9 & 74.9 & 61.5 & 64.2 & 41.3 & 49.3 & 59.0 & 73.1 & 86.8 & 35.7 & 47.2 & 63.4 & 80.8 & 58.9 & 91.4 & 85.7 & 84.4 & 70.4 & 33.6 \\
& High & 68.1 & 64.4 & 85.0 & 85.9 & 72.2 & 42.9 & 56.8 & 79.7 & 82.7 & 89.9 & 26.9 & 32.6 & 32.7 & 30.5 & 39.0 & 68.5 & 49.5 & 39.6 & 41.7 & 16.7 \\
\bottomrule
\end{tabular}
}
\end{table}

\begin{figure}[t]
  \centering
  \begin{subfigure}[t]{0.24\linewidth}
    \centering
    \includegraphics[width=\linewidth]{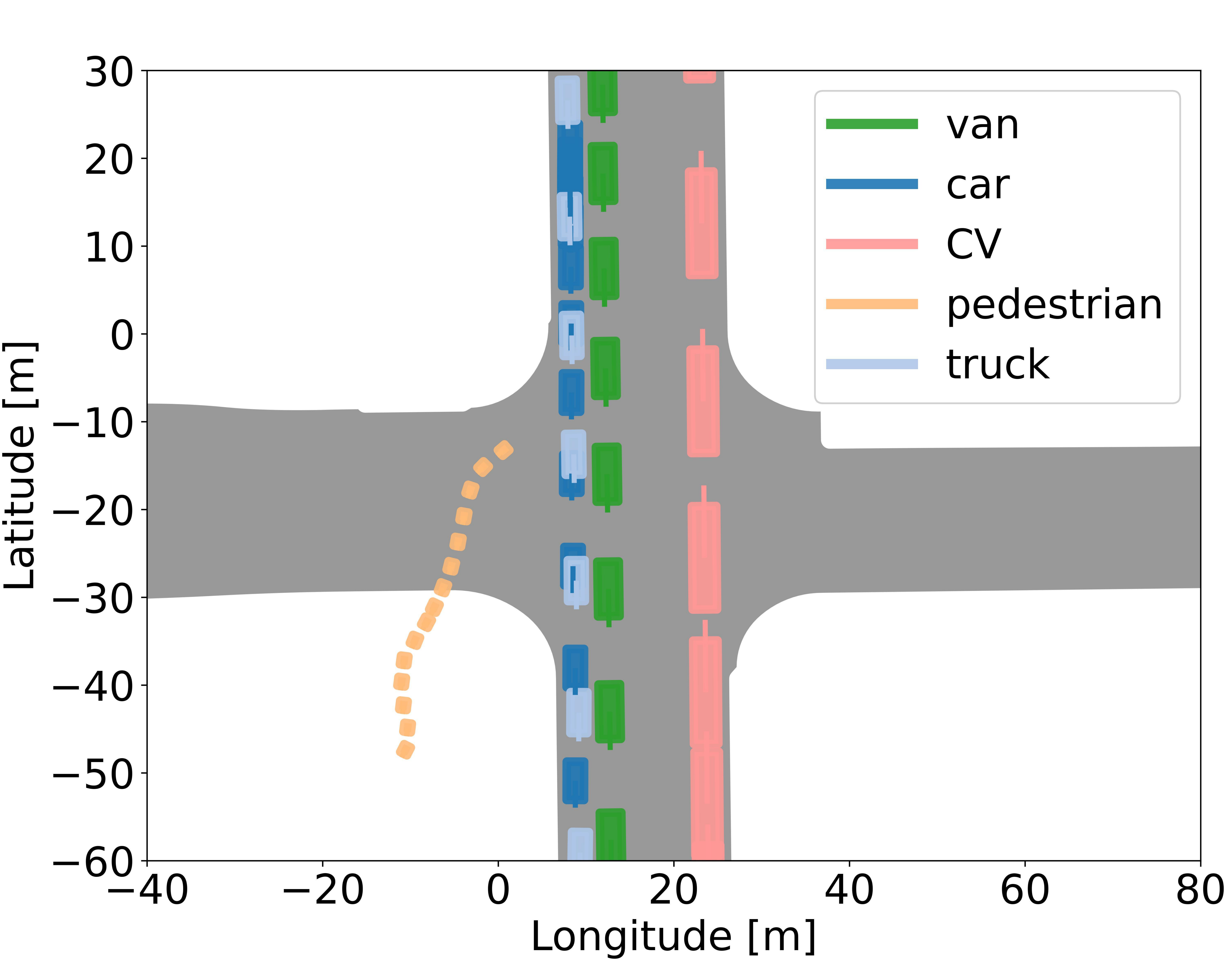}
    \caption{Ground Truth}
    \label{fig:gt_track}
  \end{subfigure}
  \begin{subfigure}[t]{0.24\linewidth}
    \centering
    \includegraphics[width=\linewidth]{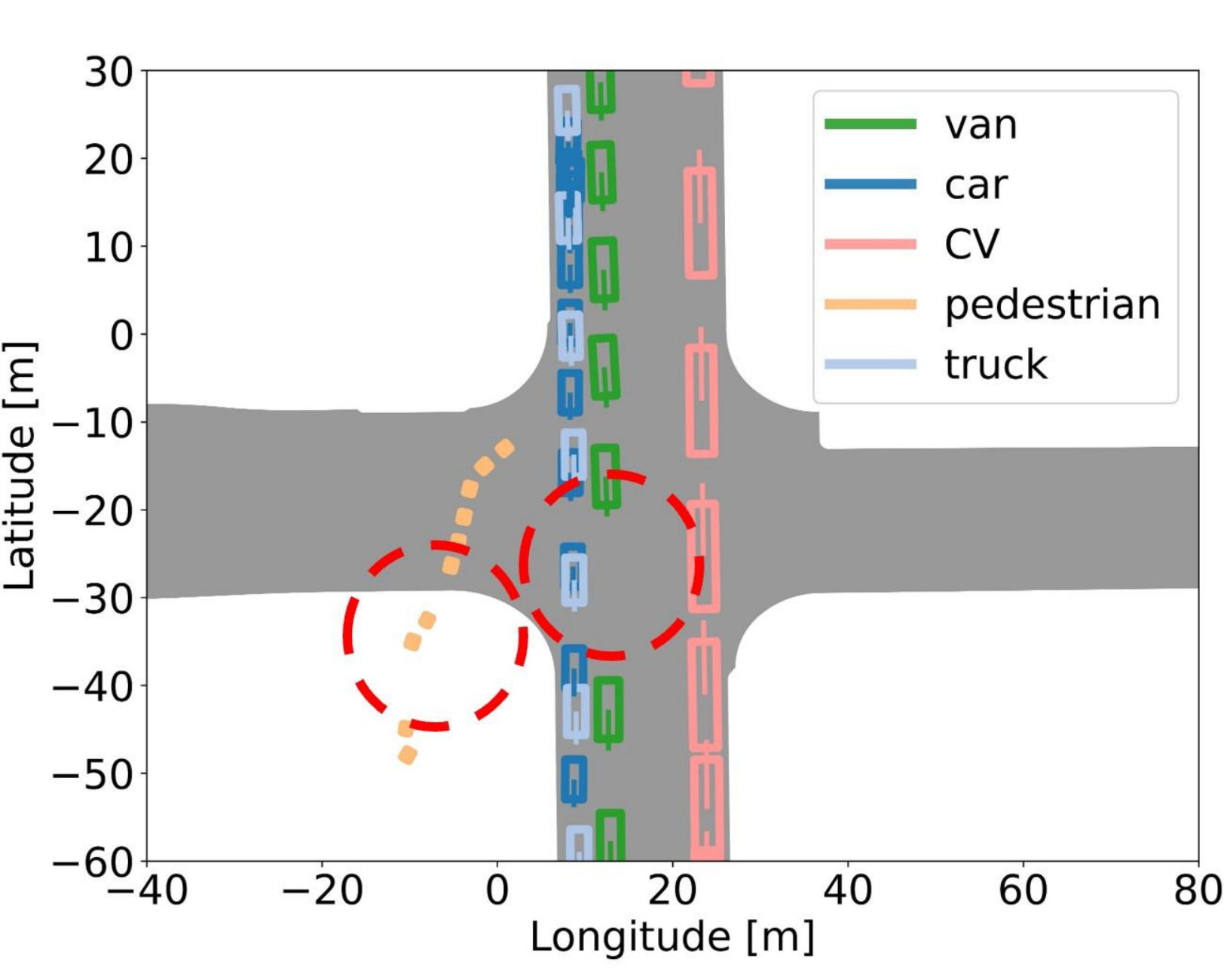}
    \caption{Low-res tracking}
    \label{fig:low_track}
  \end{subfigure}
  \begin{subfigure}[t]{0.24\linewidth}
    \centering
    \includegraphics[width=\linewidth]{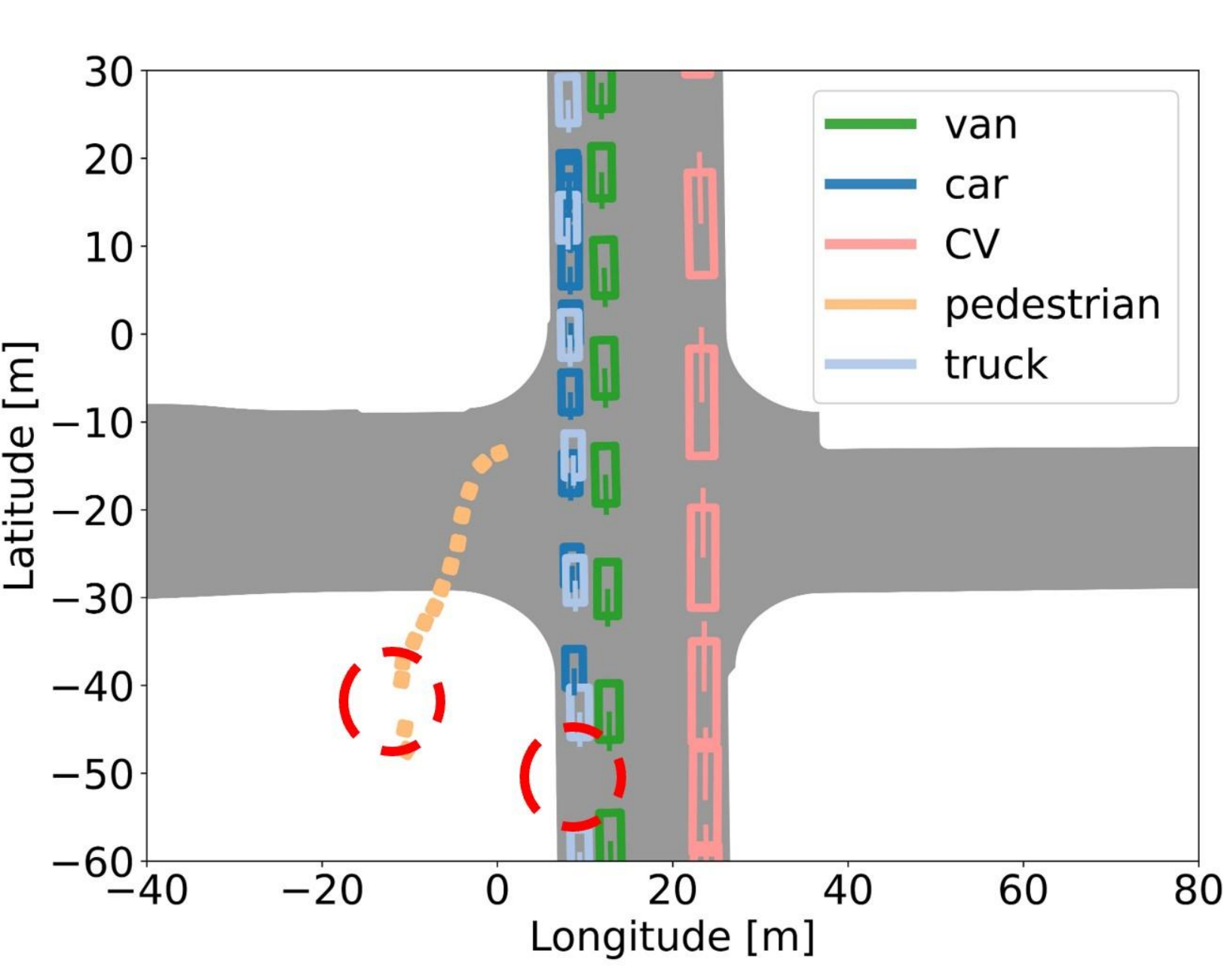}
    \caption{Mid-res tracking}
    \label{fig:mid_track}
  \end{subfigure}
  \begin{subfigure}[t]{0.24\linewidth}
    \centering
    \includegraphics[width=\linewidth]{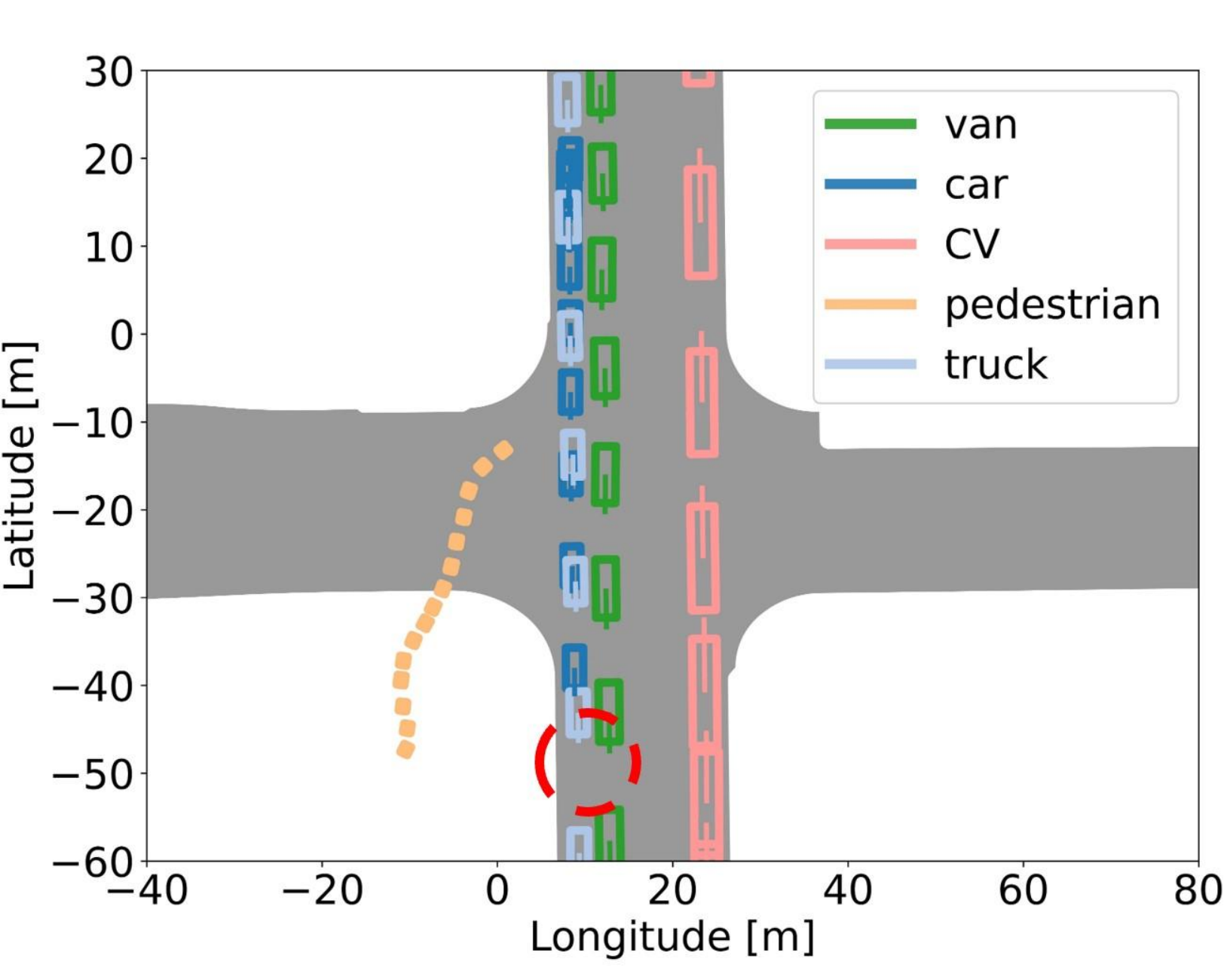}
    \caption{High-res tracking}
    \label{fig:high_track}
  \end{subfigure}
  \caption{Qualitative multi-object tracking results of SimpleTrack~\cite{simpletrack} on our RESOLVE across three resolution settings. Ground truths and predictions with a score threshold of 0.3 are overlaid and visualized at 15 Hz. Failure cases are marked in red circles.}
  \label{fig:tracking_results}
\end{figure}

\subsubsection{3D Multi-object Tracking.} According to~\cref{tab:tracking_result}, a higher resolution generally improves detection quality, thus benefiting downstream association. This is achieved by improving positioning accuracy. The average AMOTP decreases by 5.1\% from low to medium resolution and by 14.1\% from medium to high resolution. In contrast, AMOTA is not necessarily monotonically positively correlated with resolution due to factors such as association strategies and category characteristics. CenterPoint~\cite{centerpoint} provides more stable and reliable association and localization for small objects, such as fast-moving motorcycles and slow-moving pedestrians. SimpleTrack~\cite{simpletrack} achieves the highest AMOTA across most classes, demonstrating stronger matching robustness. As shown in~\cref{fig:tracking_results}, we visualize multi-frame tracking results. To avoid clutter, we display five representative trajectories of different categories in a single scene. It can be observed that higher resolution improves localization accuracy, especially for small targets.

\begin{figure}[t]
    \centering
    \includegraphics[width=0.8\linewidth]{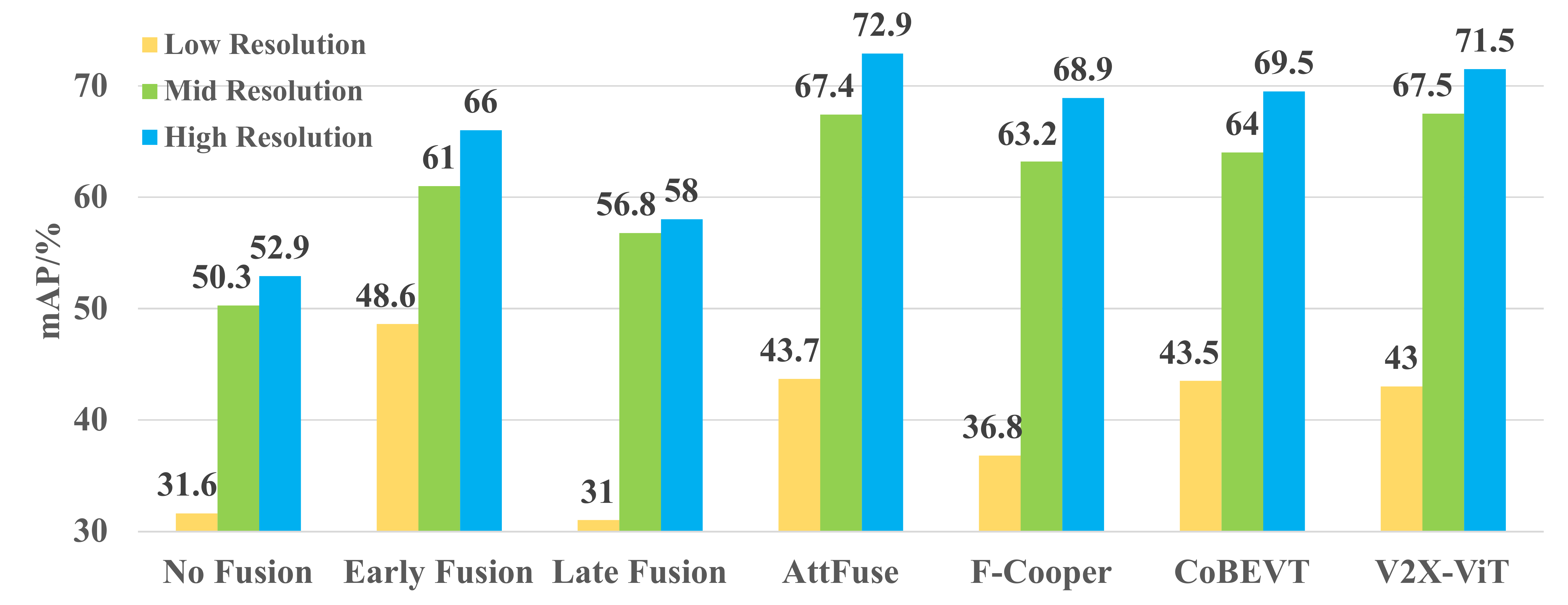}
    \caption{Agent-level cooperative perception performance under three LiDAR resolutions.}
    \label{fig:agent_perception}
\end{figure}

\subsubsection{Cooperative Perception.} As shown in Fig.~\ref{fig:agent_perception}, cooperative fusion consistently improves over the no-fusion baseline, but different fusion paradigms show different sensitivity to resolution. Early fusion performs well at higher resolutions due to raw-level aggregation, but becomes less robust with sparse LiDAR inputs. Late fusion brings limited gains, as decision-level aggregation cannot recover geometric details. By exchanging learned BEV features, intermediate fusion methods (e.g., AttFuse~\cite{OPV2V}, F-Cooper~\cite{F-cooper}, CoBEVT~\cite{CoBEVT}, and V2X-ViT~\cite{V2X-ViT}), which account for agent heterogeneity and inter-agent synchronization, outperform late and early fusion in most settings. The advantage of feature-level fusion  is most evident at medium resolution, where feature-level collaboration compensates for reduced point density using complementary viewpoints. However, the clear drop at low resolution suggests that cooperation alone cannot fully recover severely missing geometry, highlighting the need to explicitly account for resolution-induced point cloud sparsity.

\begin{table}[b]
\centering
\caption{Comparison of unimodal and multi-modal models with same LiDAR modules. Results in the same color form a paired comparison.}
\label{tab:map_comp}
\setlength{\tabcolsep}{3pt}
\renewcommand{\arraystretch}{1}
\footnotesize
\begin{tabular}{l c c c c c c}
\toprule
\multirow{2}{*}{Model} & \multirow{2}{*}{Modality} & 
\multirow{2}{*}{\makecell{LiDAR\\[2pt]Backbone}} &
\multicolumn{3}{c}{mAP\,\mbox{$\uparrow$}}\\
\cmidrule(lr){4-6}
& & & Low & Mid & High \\
\midrule
TransFusion-L~\cite{transfusion}   & L   & SparseConv  & 82.5 & \textcolor{blue}{\uline{86.6}} & \textcolor{red}{\uline{89.1}}\\
BEVFusion~\cite{bevfusion}         & L$+$C  & SparseConv  & \textcolor{blue}{\uline{86.3(-0.3)}} & \textcolor{red}{\uline{92.6(+3.5)}} & 93.1 \\
\bottomrule
\end{tabular}
\end{table}

\subsection{Analysis}
\subsubsection{Multimodal Perception Gain under point sparsity.}
\begin{figure}[t]
  \centering
  \begin{subfigure}[t]{0.32\linewidth}
    \centering
    \includegraphics[width=\linewidth]{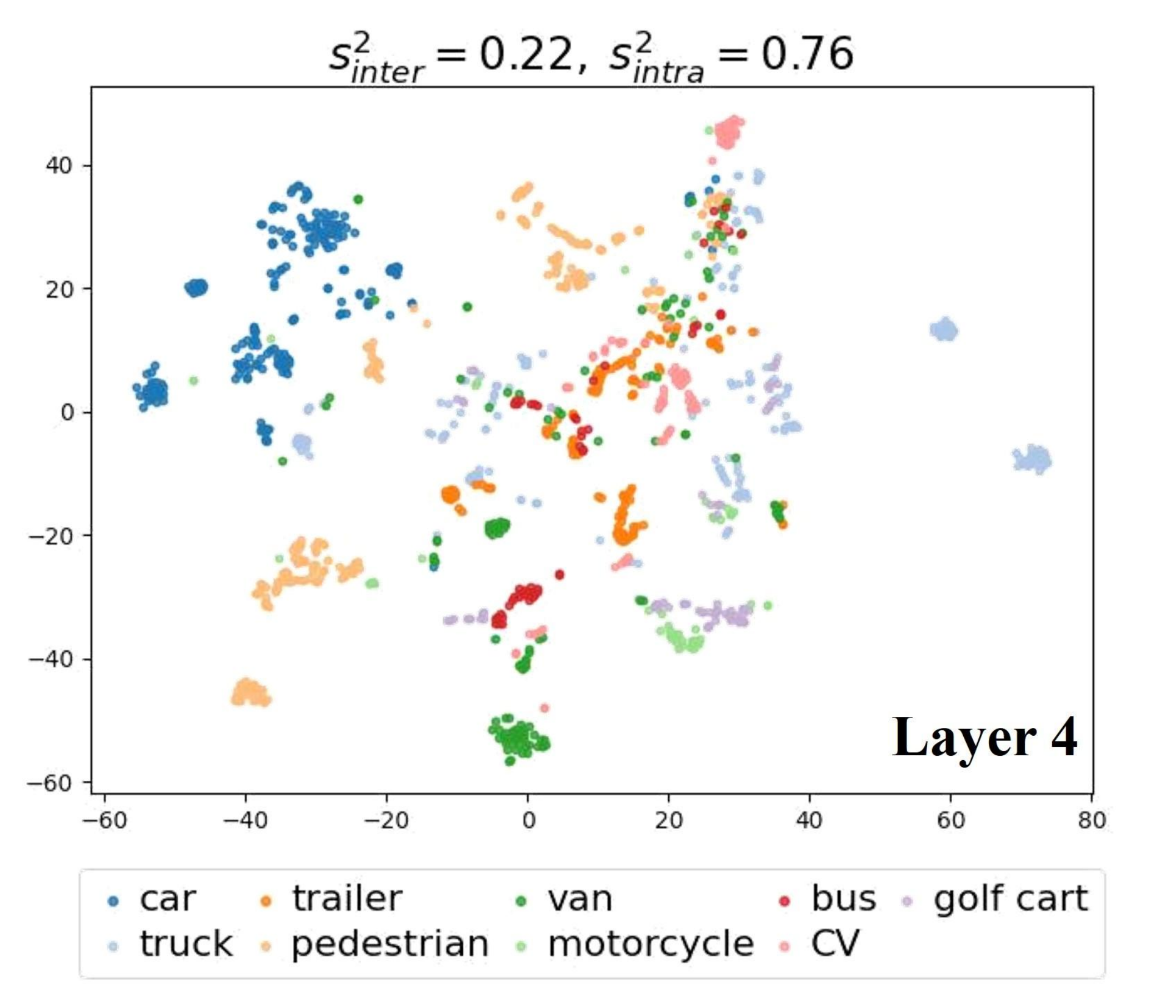}
    \caption{Low-res, mAP=82.5}
    \label{fig:high_lidar_only_tsne}
  \end{subfigure}
  \begin{subfigure}[t]{0.32\linewidth}
    \centering
    \includegraphics[width=\linewidth]{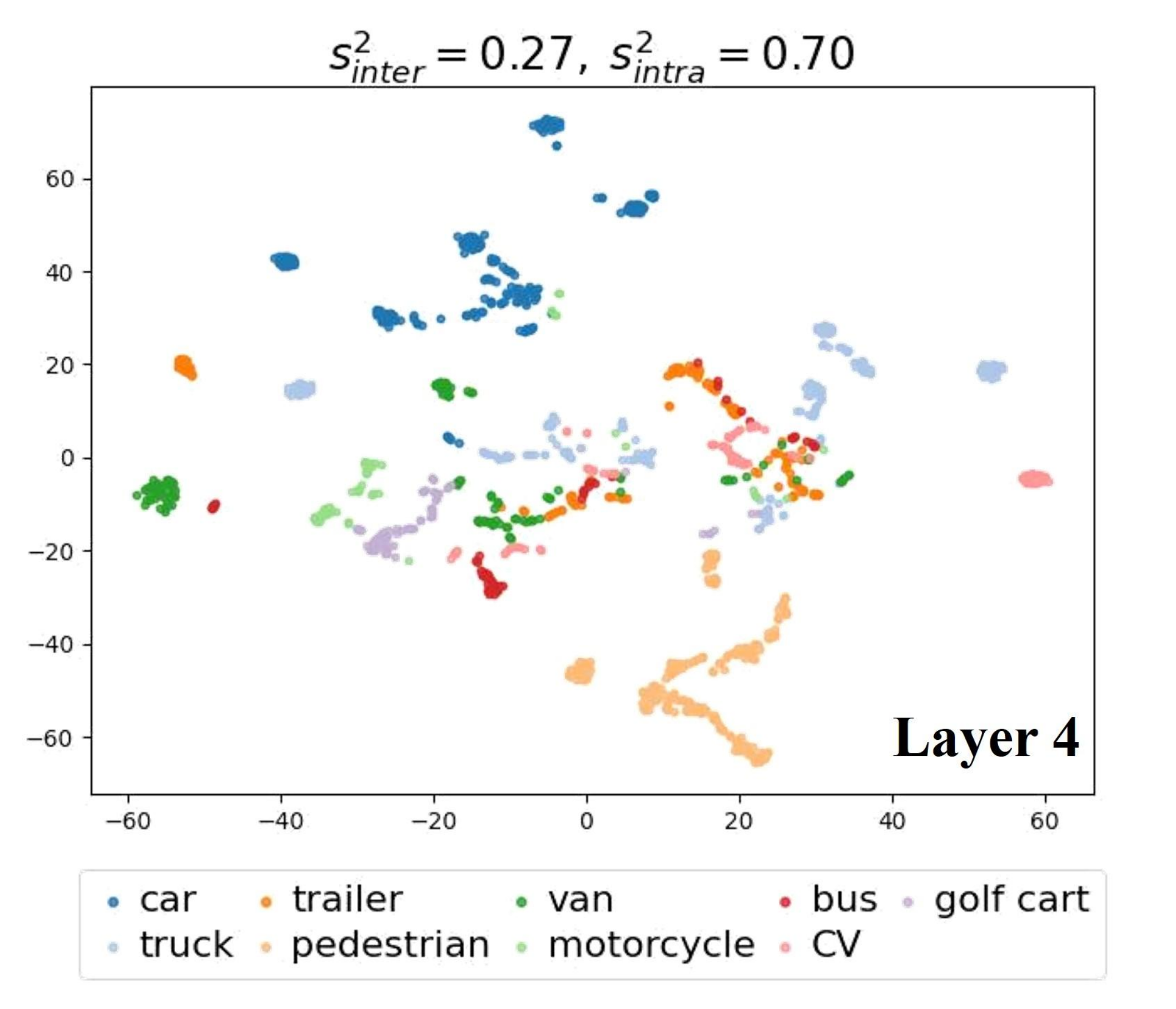}
    \caption{Mid-res, mAP=86.6}
    \label{fig:mid_lidar_only_tsne}
  \end{subfigure}
  \begin{subfigure}[t]{0.32\linewidth}
    \centering
    \includegraphics[width=\linewidth]{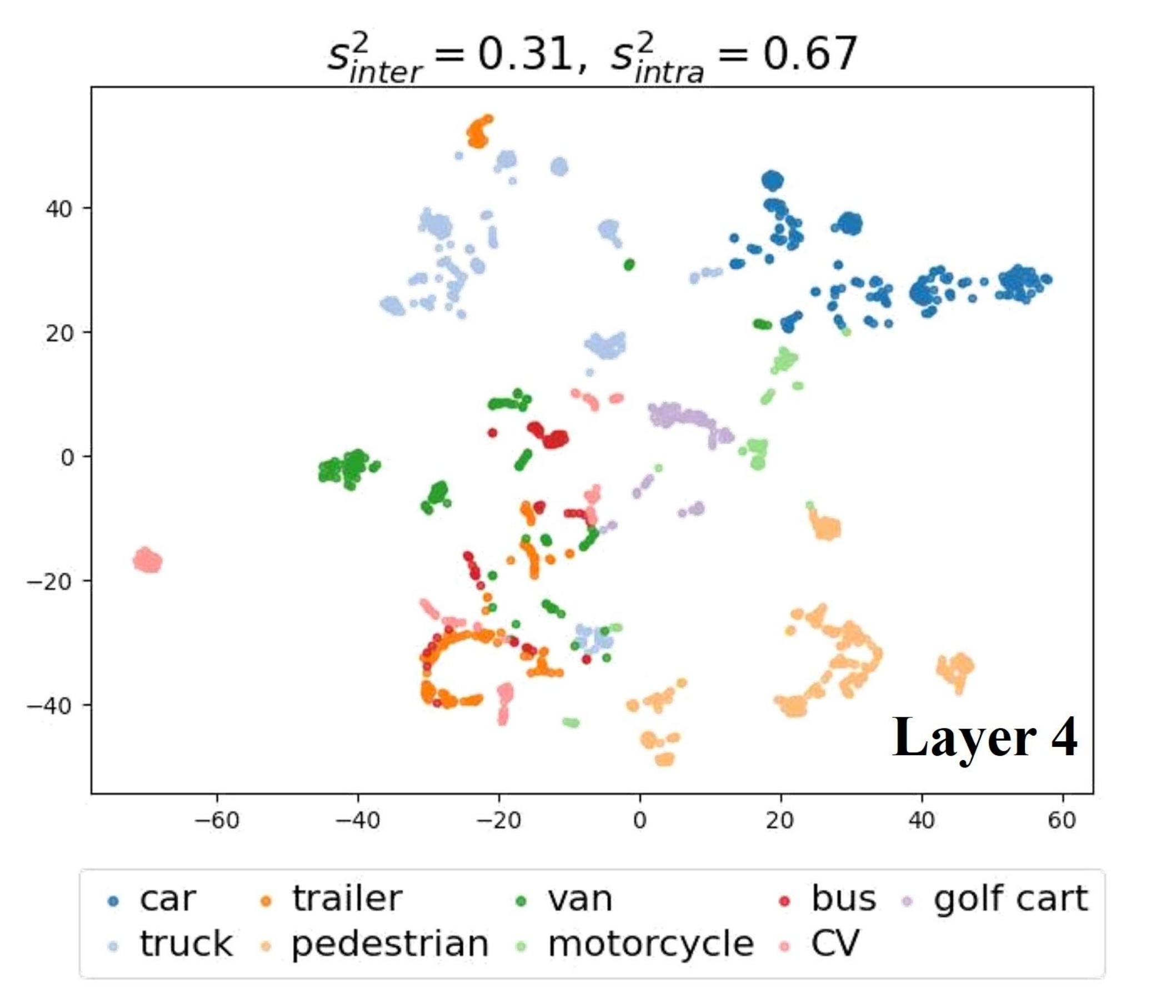}
    \caption{High-res, mAP=89.1}
    \label{fig:low_lidar_only_tsne}
  \end{subfigure}
  \caption{t-SNE plots of the impact of LiDAR resolutions on feature representation capability of unimodal models under Transfusion-L~\cite{transfusion} architecture.}
  \label{fig:lidar_tsne_results}
\end{figure}

Higher-resolution LiDAR improves unimodal detection performance primarily through enhanced feature discriminability. To examine this effect, we visualize the features of the last backbone layer using t-SNE~\cite{t-SNE} and quantify class separability using inter-class and intra-class variance~\cite{fisher}, as shown in~\cref{fig:lidar_tsne_results}. The results show that higher resolution leads to larger inter-class variance and smaller intra-class variance, indicating improved separability across classes and stronger clustering within the same class. We further observe that multimodal fusion models using lower-resolution data match or even outperform unimodal models using higher-resolution data, as shown in~\cref{tab:map_comp}, suggesting that camera features may partially compensate for LiDAR sensing sparsity. 

\begin{figure}[t]
  \centering
  \includegraphics[width=\linewidth]{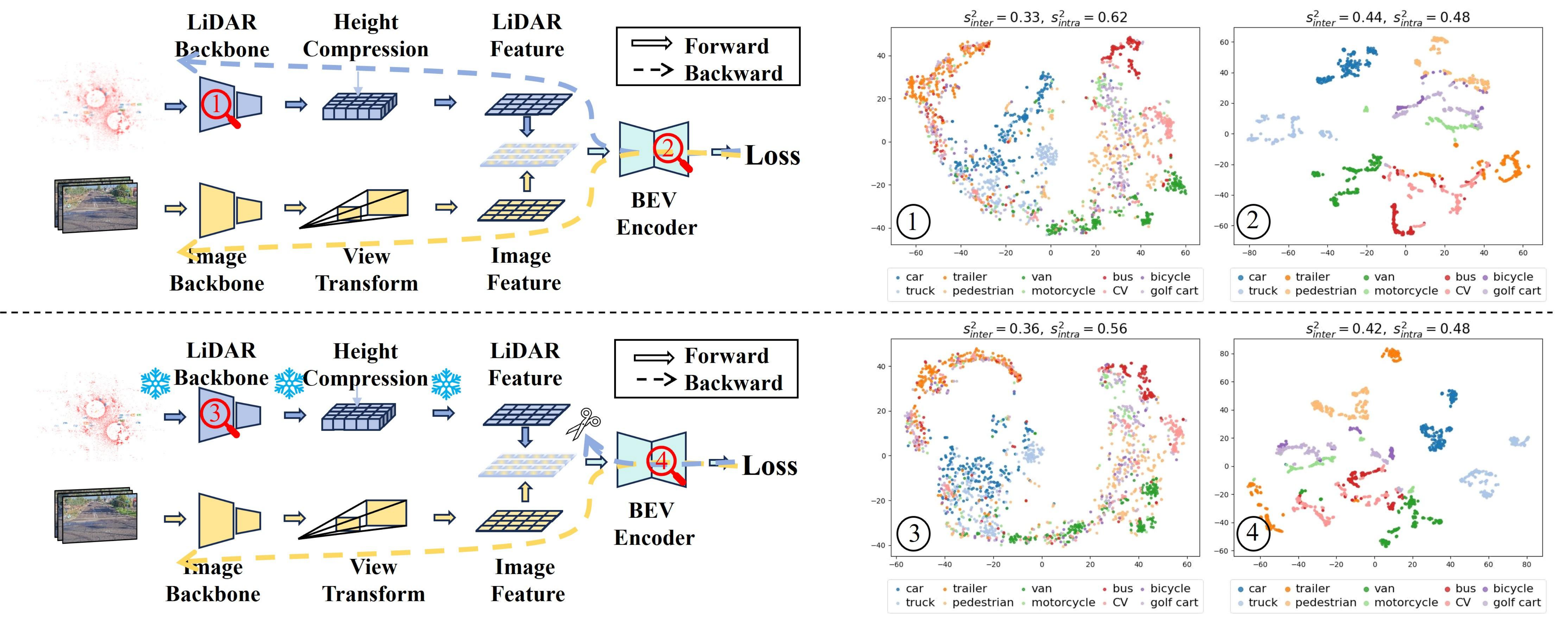}
  \caption{Controlled experimental design under a general multimodal architecture. Top: standard backpropagation, mAP=93.1. Bottom: stop gradients in the LiDAR branch and freeze LiDAR-module parameters, mAP=89.4. \Circled{1} and \Circled{3} visualize the features of the first layer of LiDAR backbone, while \Circled{2} and \Circled{4} visualize the fused features.}
  \label{fig:arch_comp}
\end{figure}

To further investigate how multimodal learning influences LiDAR feature representations, we follow the protocol in~\cite{MultimodalBoosting} and conduct two training settings in the BEVFusion~\cite{bevfusion} architecture, as shown on the left of~\cref{fig:arch_comp}: (i) allow standard training with gradient backpropagation to both LiDAR and camera branches, and (ii) truncate the gradient from the multimodal loss propagated to the LiDAR backbone, and substitute it the unimodal LiDAR loss. We analyze feature distributions from early LiDAR backbone layers and late fusion layers using t-SNE visualizations and variance metrics (right of~\cref{fig:arch_comp}), allowing both qualitative and quantitative investigation of how camera information influences LiDAR feature encoding.


\begin{figure}[t]
  \centering
  \begin{subfigure}[t]{0.32\linewidth}
    \centering
    \includegraphics[width=\linewidth]{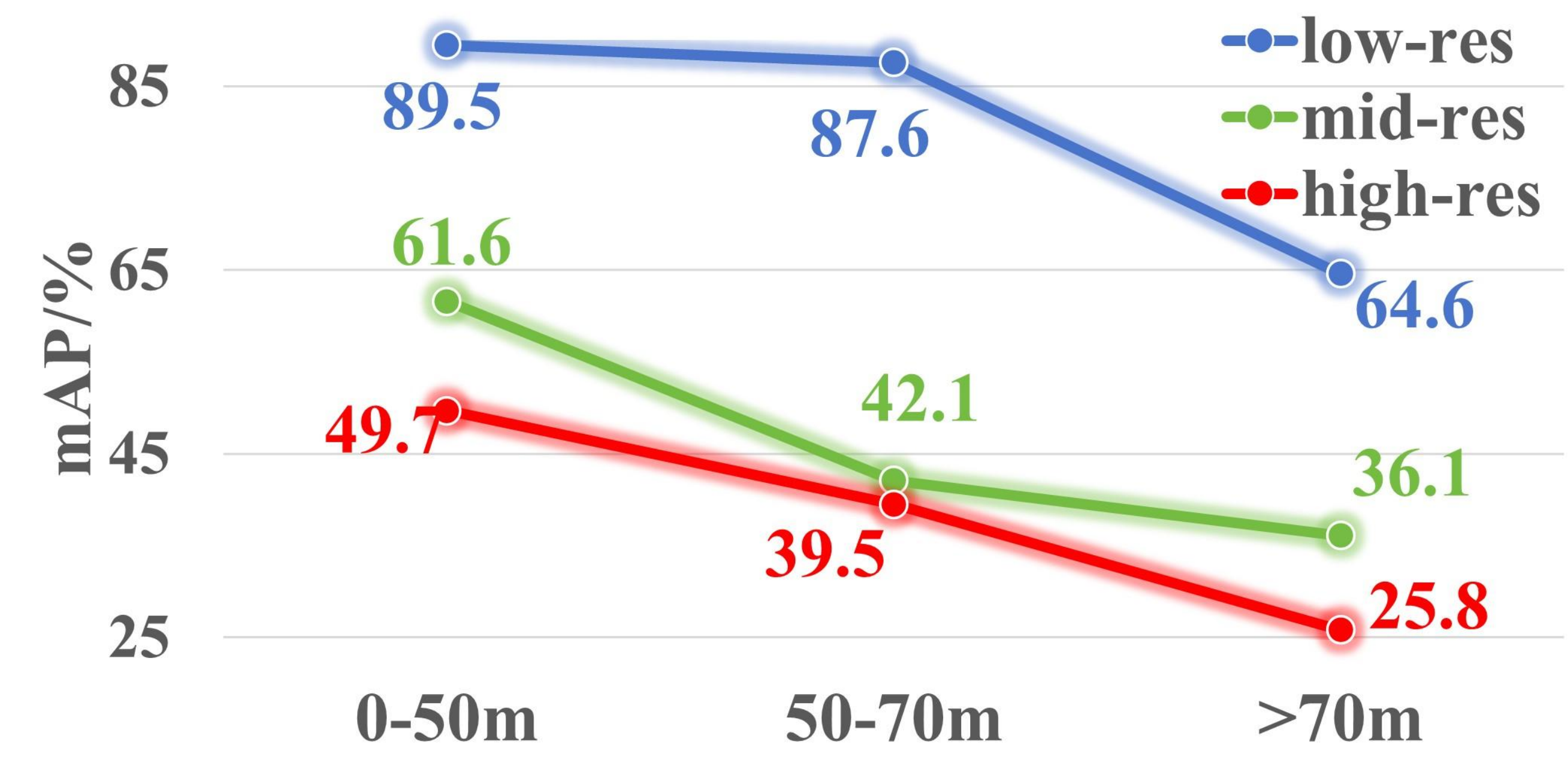}
    \caption{Under low-res training}
    \label{fig:low_test_transfer}
  \end{subfigure}
  \begin{subfigure}[t]{0.32\linewidth}
    \centering
    \includegraphics[width=\linewidth]{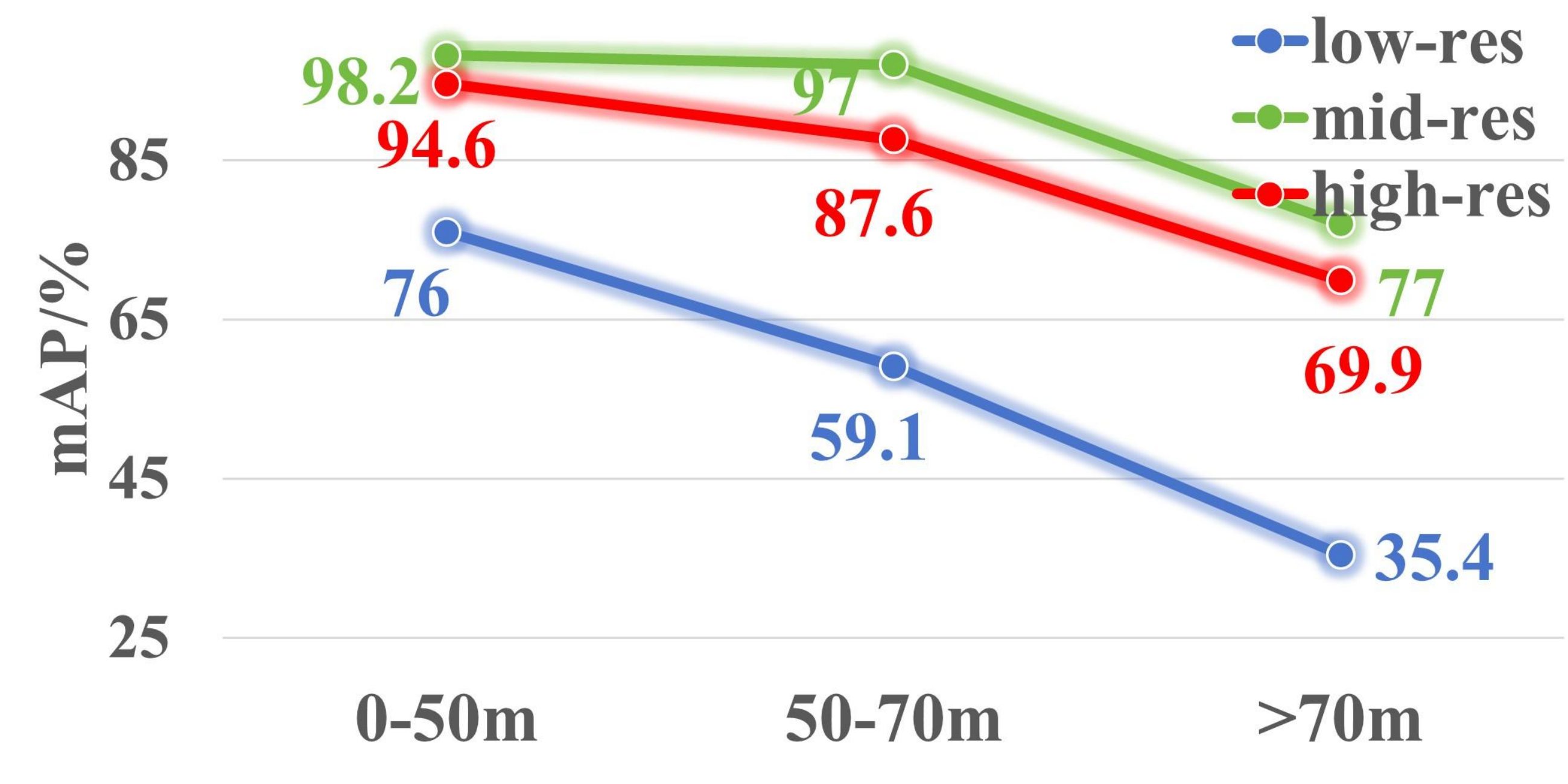}
    \caption{Under mid-res training}
    \label{fig:mid_test_transfer}
  \end{subfigure}
  \begin{subfigure}[t]{0.32\linewidth}
    \centering
    \includegraphics[width=\linewidth]{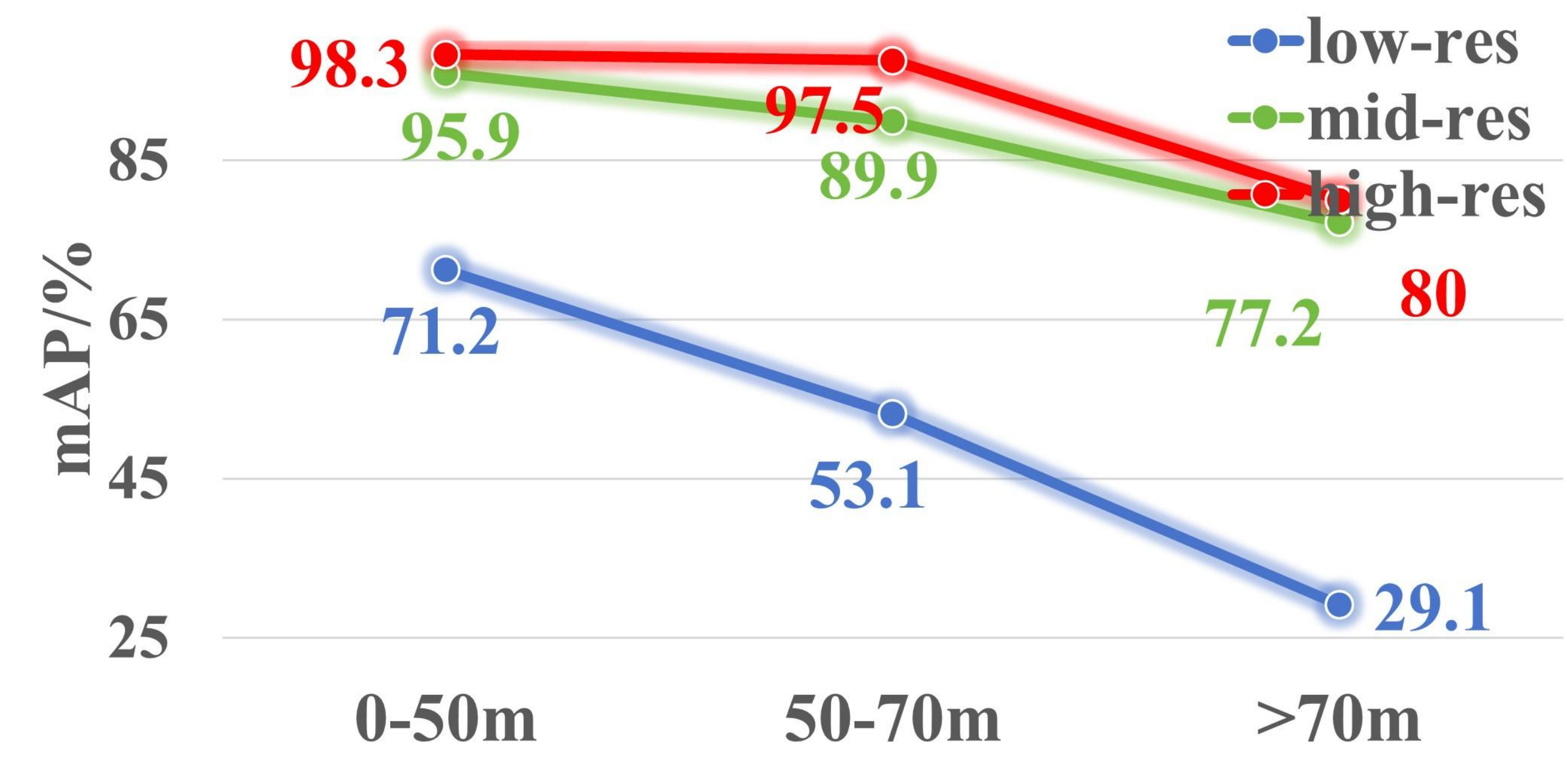}
    \caption{Under high-res training}
    \label{fig:high_test_transfer}
  \end{subfigure}
  \caption{Unimodal performance degradation caused by different LiDAR resolutions used during training and inference. Voxel Mamba~\cite{voxelmamba} is used as unimodal detector.}
  \label{fig:transfer_learning}
\end{figure}

Compared to the truncated-gradient setting, standard multimodal training (marked number 1 in~\cref{fig:arch_comp}) leads to reduced class separability in early LiDAR backbone layers, indicating that multimodal loss alters feature learning both in late fusion stages and in the LiDAR branch during early representation stages. Despite this reduced early-layer discriminability due to gradient from multi-modal loss~\cite{MultimodalBoosting}, the fusion model achieves higher detection performance and stronger feature separability at the fusion layers (marked number 2 in~\cref{fig:arch_comp}). This implies that multimodal learning encourages the LiDAR backbone to rely less on point-density cues and instead align more effectively between camera and LiDAR features~\cite{Supervise}. Overall, these results suggest that multimodal fusion can mitigate LiDAR sensing sparsity by reshaping LiDAR feature learning, enabling competitive perception performance even with lower-resolution sensors and offering clues for designing more cost-effective multimodal perception systems in roadside cooperative perception scenarios.

\subsubsection{Impact of LiDAR Resolution Mismatch.} As shown in~\cref{fig:transfer_learning}, benchmark results also imply that LiDAR resolution shifts between training and inference introduce a clear domain gap that degrades unimodal detection performance. While previous work~\cite{DomainGeneralization,lidarDistillation,DensityInsensitive} on domain adaptation uses downsampling or cross-dataset settings to investigate this issue, RESOLVE provides a controlled protocol that isolates resolution shifts. The model trained at low resolution drops by 40\% when transferred to higher resolution, while a high-resolution model drops only 4.6\% on mid resolution but still drops 40\% on low resolution with severe long-range degradation. This gap is even more detrimental for multimodal detection (see supplementary materials), indicating that resolution mismatch disrupts both unimodal feature learning and cross-modal alignment.

\section{Conclusion}
We propose RESOLVE, the first real-world multi-resolution, multi-modal dataset for roadside cooperative perception. It benchmarks how resolution-induced point-cloud distribution shifts affect model performance on downstream roadside perception tasks, while isolating other sensing factors. We also conduct comprehensive experiments on unimodal and multimodal models. The results show how multimodal fusion can compensate for degraded point-cloud branch feature learning under reduced LiDAR resolution, providing insights into the design of more robust and cost-effective roadside perception models that explicitly account for resolution mismatch, target distance, and object diversity. One limitation of RESOLVE is that it is restricted to single-intersection scenarios. We will explore multi-intersection and corridor-level scenarios in future work. 

\section*{Acknowledgements}
The author would like to thank the Maricopa County Department of Transportation, Arizona, USA, for its technical support. This work was partially supported by an NVIDIA Academic Grant Program Award.

%
%
\bibliographystyle{splncs04}
\bibliography{main}

\clearpage

\appendix
\section*{Appendix}

\begin{figure}[b]
  \centering
  \begin{subfigure}[t]{0.24\linewidth}
    \centering
    \includegraphics[width=\linewidth,height=2.5cm]{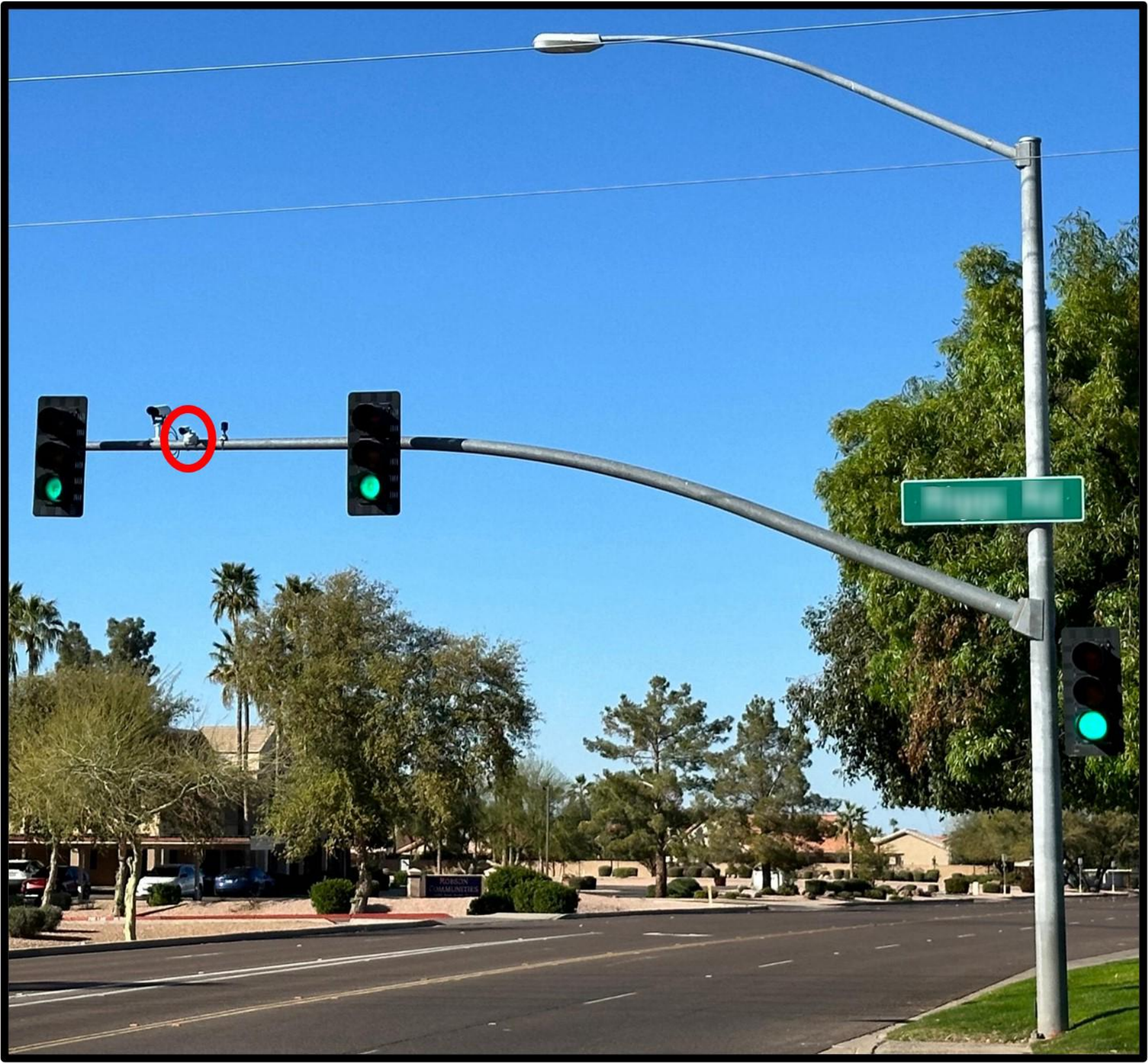}
     \caption{NORTH CAM}
  \end{subfigure}
  \begin{subfigure}[t]{0.24\linewidth}
    \centering
    \includegraphics[width=\linewidth,height=2.5cm]{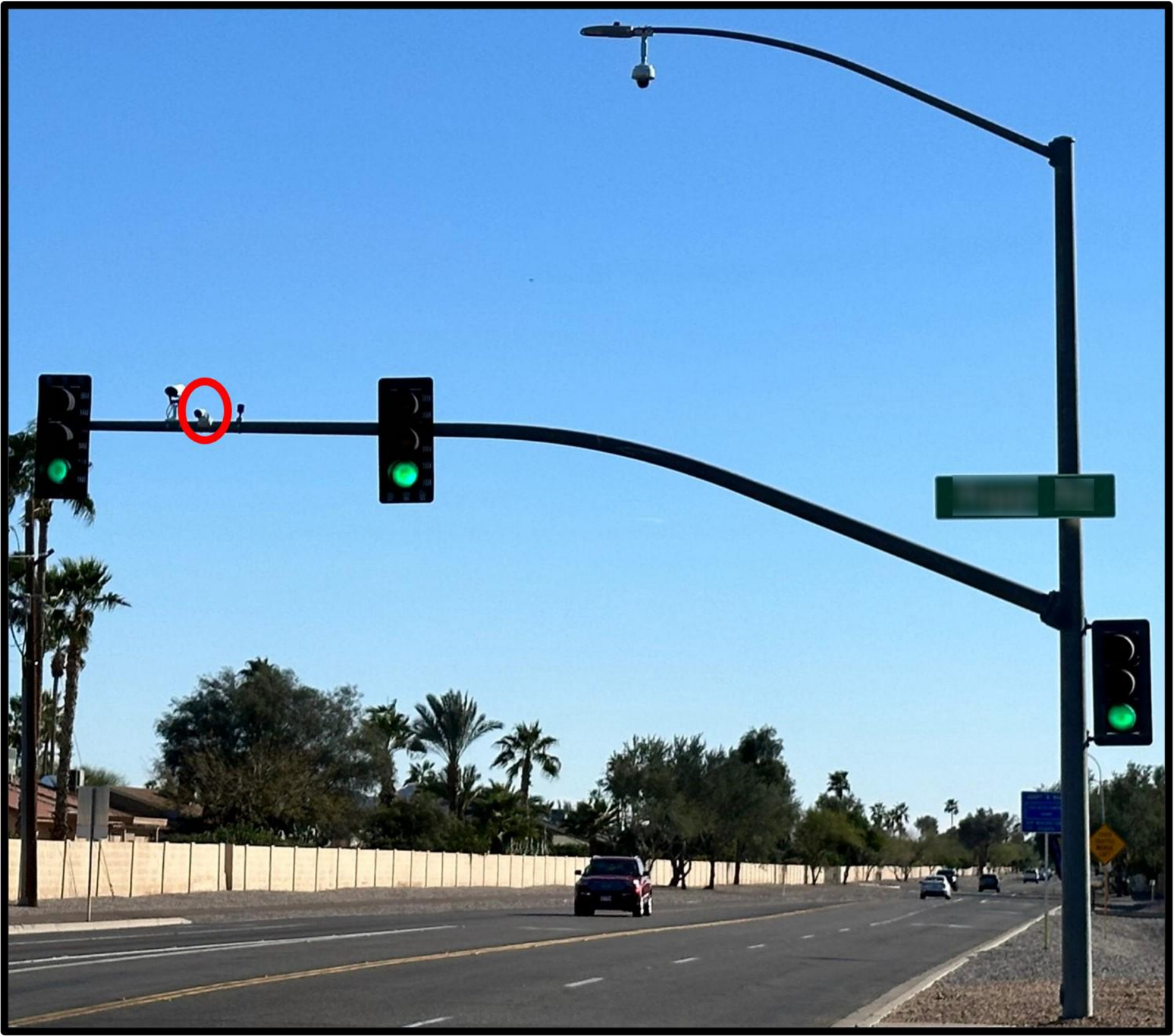}
    \caption{SOUTH CAM}
  \end{subfigure}
  \begin{subfigure}[t]{0.24\linewidth}
    \centering
    \includegraphics[width=\linewidth,height=2.5cm]{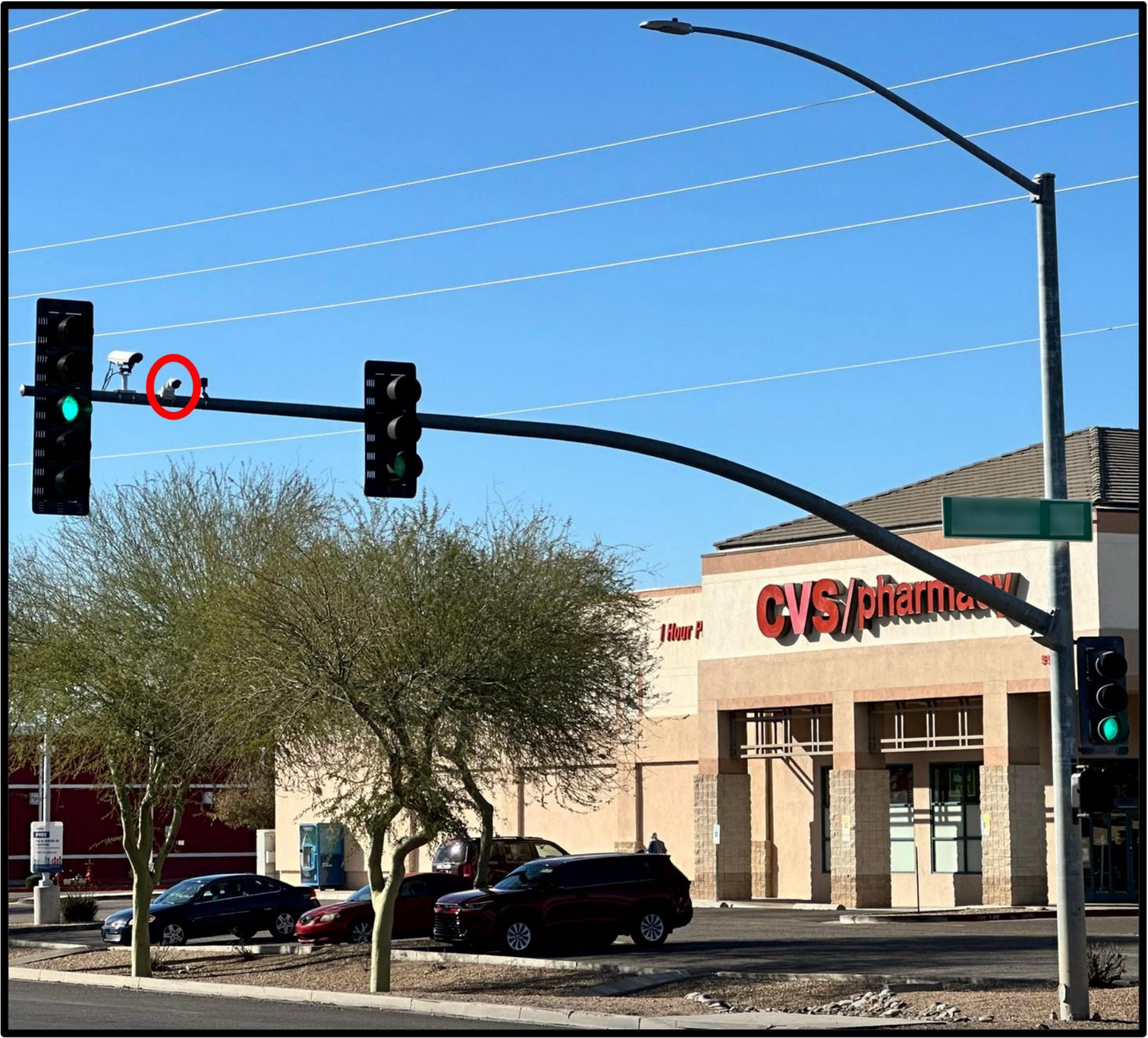}
    \caption{WEST CAM}
  \end{subfigure}
  \begin{subfigure}[t]{0.24\linewidth}
    \centering
    \includegraphics[width=\linewidth,height=2.5cm]{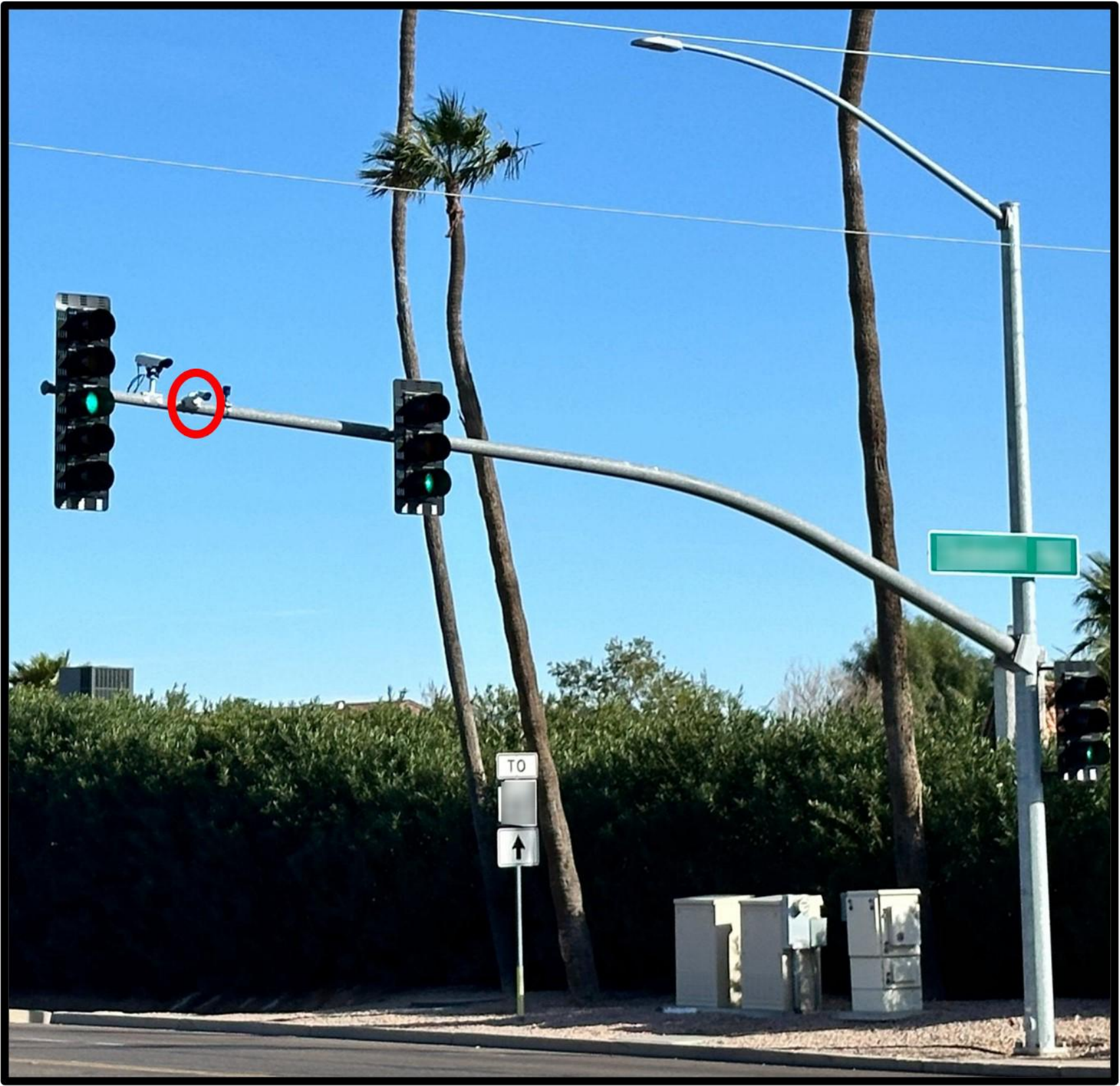}
    \caption{EAST CAM}
  \end{subfigure}
  \caption{Installation location diagram of infrastructure cameras, marked in red circles.}
  \label{fig:cam_locations}
\end{figure}

\begin{figure}[t]
  \centering
  \begin{subfigure}[t]{0.4\linewidth}
    \centering
    \includegraphics[width=\linewidth,height=6cm]{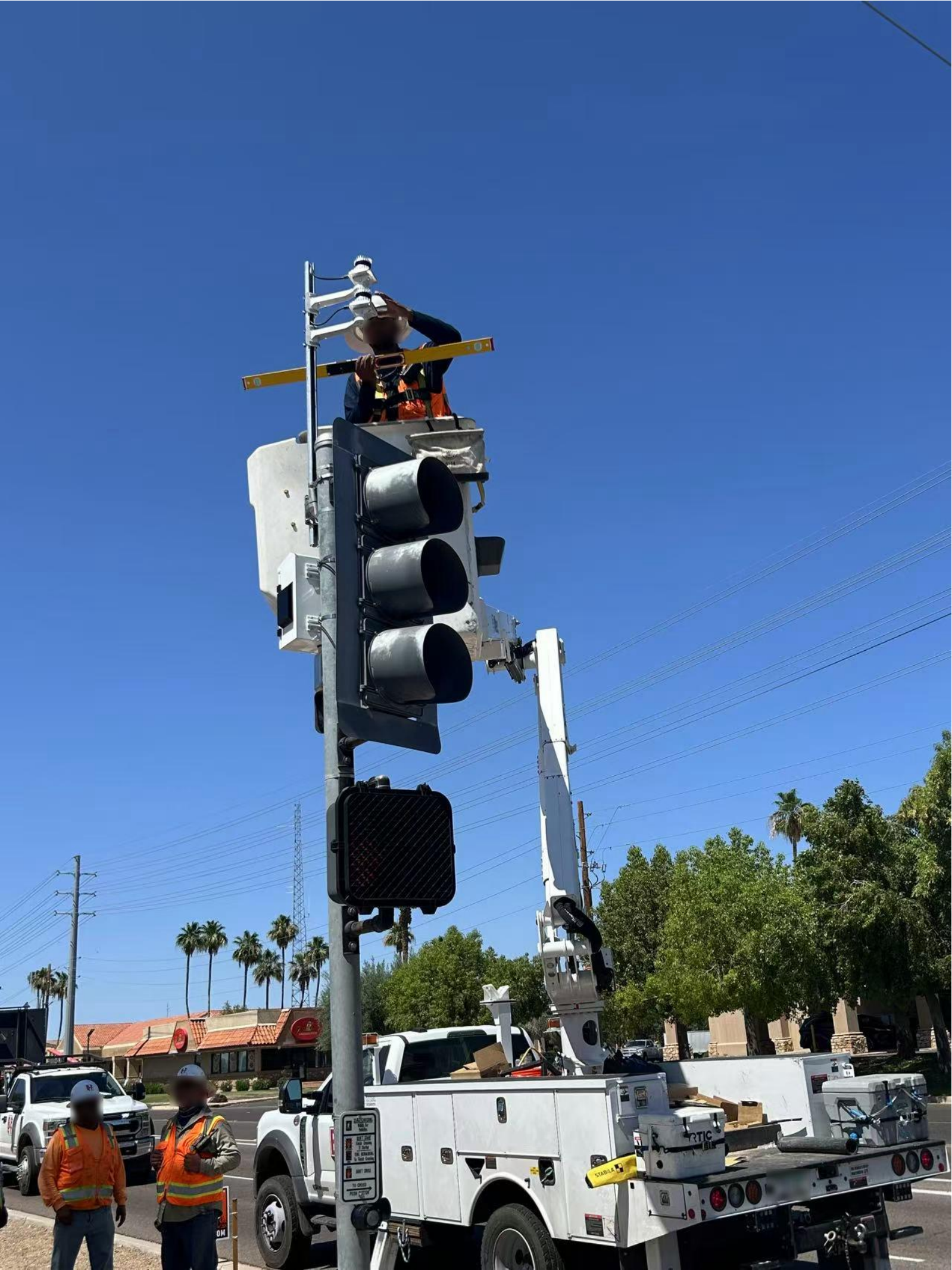}
     \caption{Ensure precise alignment of different LiDARs}
    \label{fig:lidar_place}
  \end{subfigure}
  \begin{subfigure}[t]{0.5\linewidth}
    \centering
    \includegraphics[width=\linewidth,height=6cm]{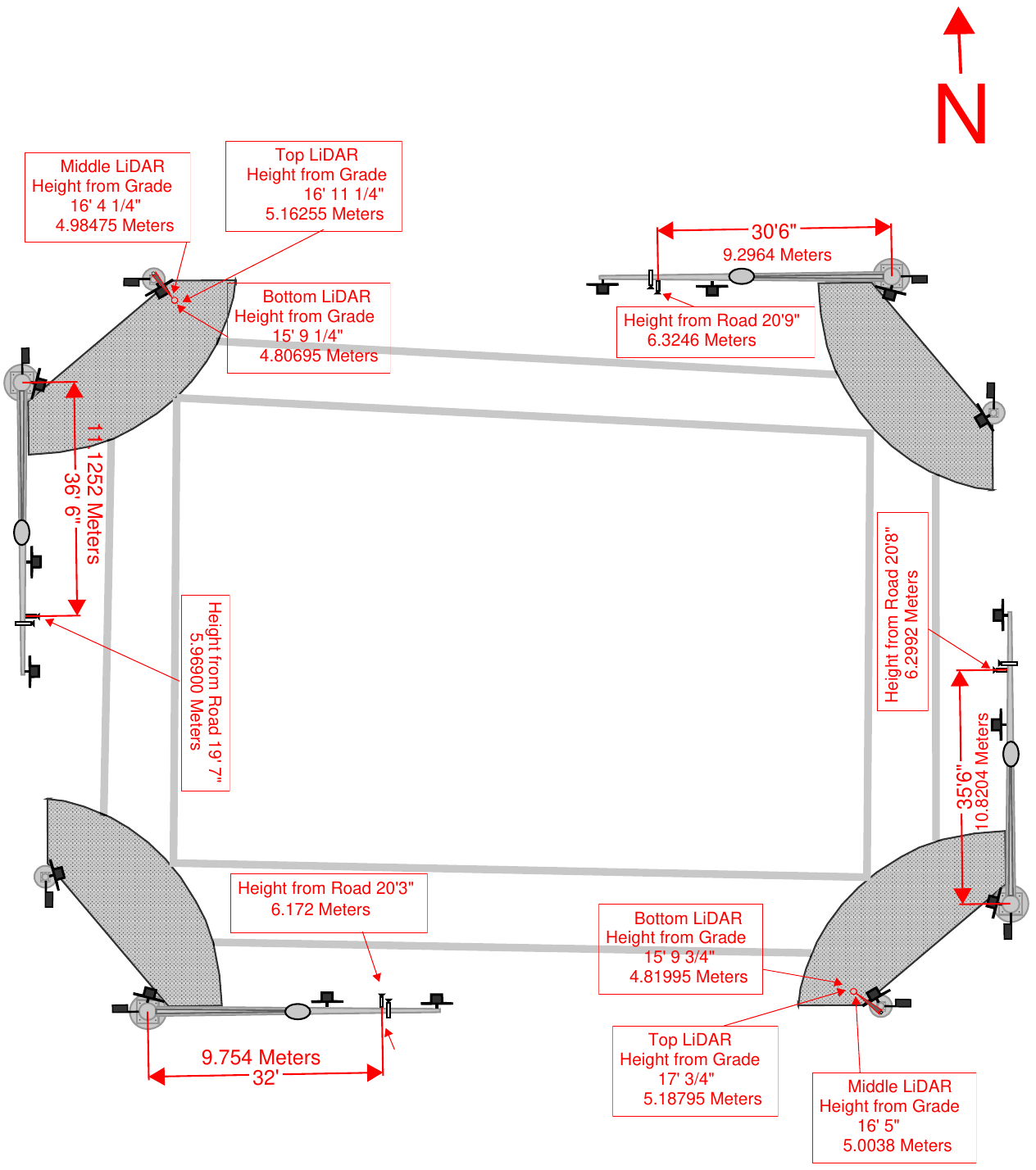}
    \caption{Precise position measurement of all sensors}
    \label{fig:position_all_sensors}
  \end{subfigure}
  \caption{A professional team ensures that the LiDAR installations within each group differ only in height and provides specific measurement data for all sensors.}
  \label{fig:sensor_position_overrall}
\end{figure}

\begin{figure}[t]
    \centering
    \includegraphics[width=0.47\linewidth]{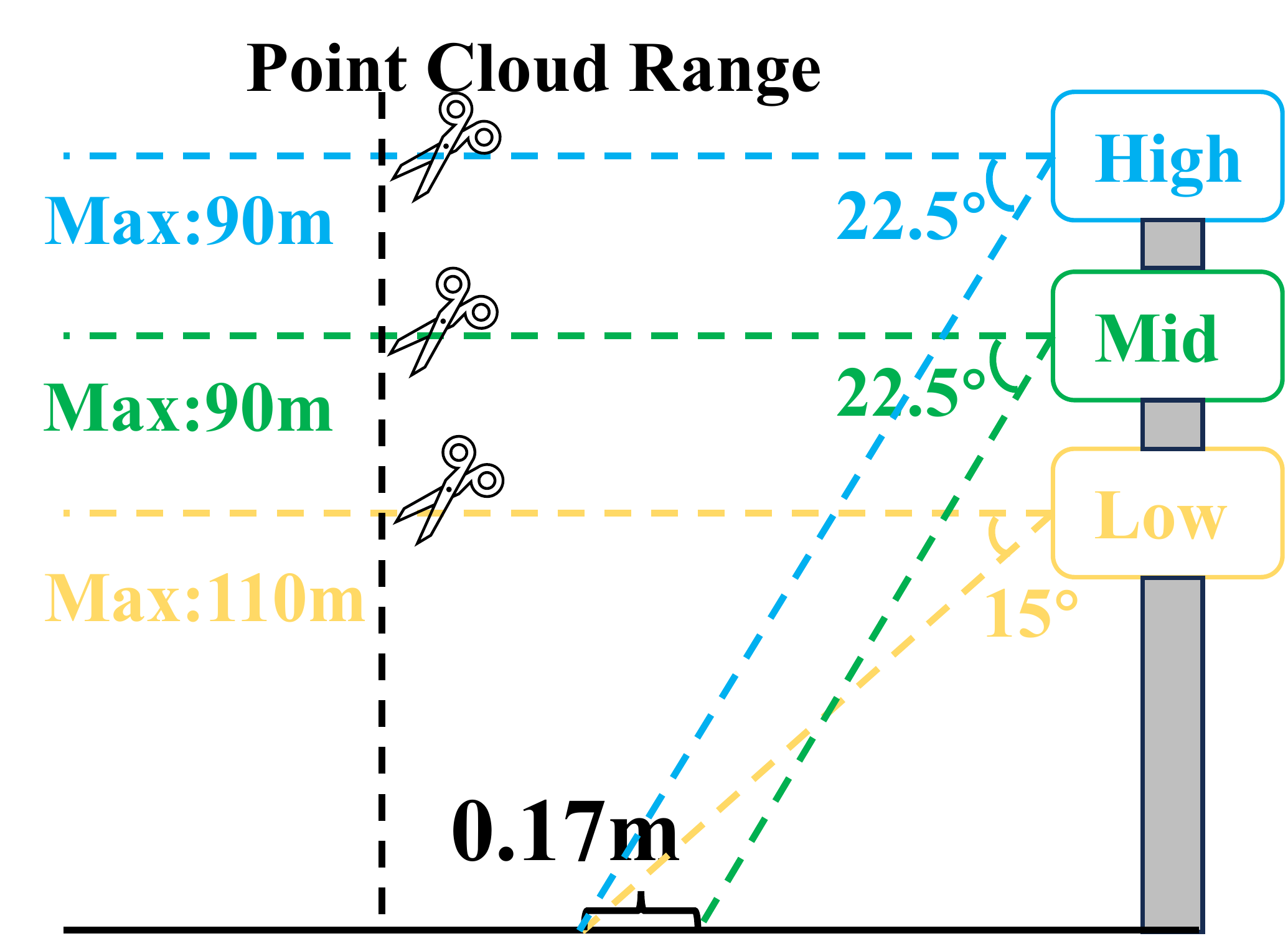}
     \caption{Influence by sensing range, VFOV, and installation height}
    \label{fig:influence}
\end{figure}

\section{Sensor Setup}
This section provides more details of sensor setup during real-world deployment. To align with common traffic camera deployment practices used by local transportation agencies, the research team installed four cameras on traffic signal poles at the four approaches of the intersection, as shown in~\cref{fig:cam_locations}. After fine adjustments to their orientation and field of view, the setup achieved full 360° coverage with minimal blind spots. 

Additionally, with the goal of balancing sensing coverage and maintaining relatively uniform point cloud distribution near the intersection center for roadside cooperative perception~\cite{lidar_place_infra}, two sets of LiDAR sensors with three levels of resolution (six units in total) were deployed on pedestrian signal poles at the northwest and southeast corners of the intersection, as shown in~\cref{fig:lidar_place}. The LiDAR systems were configured to differ only in installation height while keeping all other parameters identical. For each set of LiDARs with the same resolution placed at the two diagonal corners, the installation heights were selected to minimize sensing gaps near the intersection center while expanding coverage along the four approaching road segments. Also, the horizontal position coordinates and zero-phase orientation for all LiDARs are kept consistent for the ease of downstream data calibration. We also provide a coordinate diagram of all sensors obtained through precise measurements to facilitate initial coarse alignment and estimation during sensor spatial calibration in~\cref{fig:position_all_sensors}.

To reduce confounding factors and isolate the effect of LiDAR resolution, we keep installation-related factors as consistent as possible as shown in Fig.~\ref{fig:influence}. The high- and medium-resolution LiDARs have the same detection range and vertical field of view (VFOV). Their mounting heights differ by only 0.17 m, resulting in 99\% overlap in ground coverage. Although the low-resolution LiDAR has a narrower VFOV ($\pm15^\circ$) and a longer detection range (110 m), we lower its mounting height and apply a unified point cloud range constraint ($x,y \in [-60, 88.8]$ m), achieving 95\% effective ground-coverage overlap with the high-resolution LiDAR. Notably, mounting all LiDARs on high-level traffic signal poles would substantially reduce this overlap.

\section{Data Annoatation}
We complete 2D and 3D annotations using the MolarData platform. The annotations undergo multiple rounds of revision and review by professionals to ensure quality. The dataset contains 10 categories of annotated objects: pedestrians, golf carts, motorcycles, bicycles, cars, trucks, vans, construction vehicles, trailers, and buses. For each object, we annotate its bounding box center coordinates, dimensions, and the yaw angle, and assign a unique tracking ID. Since the annotation tool cannot directly provide velocity information, we calculate the decomposed velocities in the x and y directions using the displacement of adjacent frames during post-processing.

Furthermore, we fuse the point clouds from two synchronized high-resolution LiDARs into one coordinate and perform 3D annotation only from the perspective of this fused point cloud to effectively reduce the annotation workload. After completing the 3D annotation in the fused point cloud, and combining the spatial calibration results, we project these 3D bounding boxes onto the coordinate system of LiDAR at other resolutions and count the number of points within each box. When the number of points is greater than a threshold of five, the box is considered a valid box and included in the dataset statistics. When the number of points is less than five, the box is still retained but not counted as a valid box. For 2D annotation, the platform supports 3D-to-2D projection transformation. Given the extrinsic transform matrix of the LiDAR to each camera, the labeled 3D bounding boxes can be automatically projected onto each image. Subsequently, the 2D bounding boxes obtained by manual adjustment are performed to obtain more accurate 2D annotation results.

\begin{figure}[t]
  \centering
   \includegraphics[width=\linewidth]{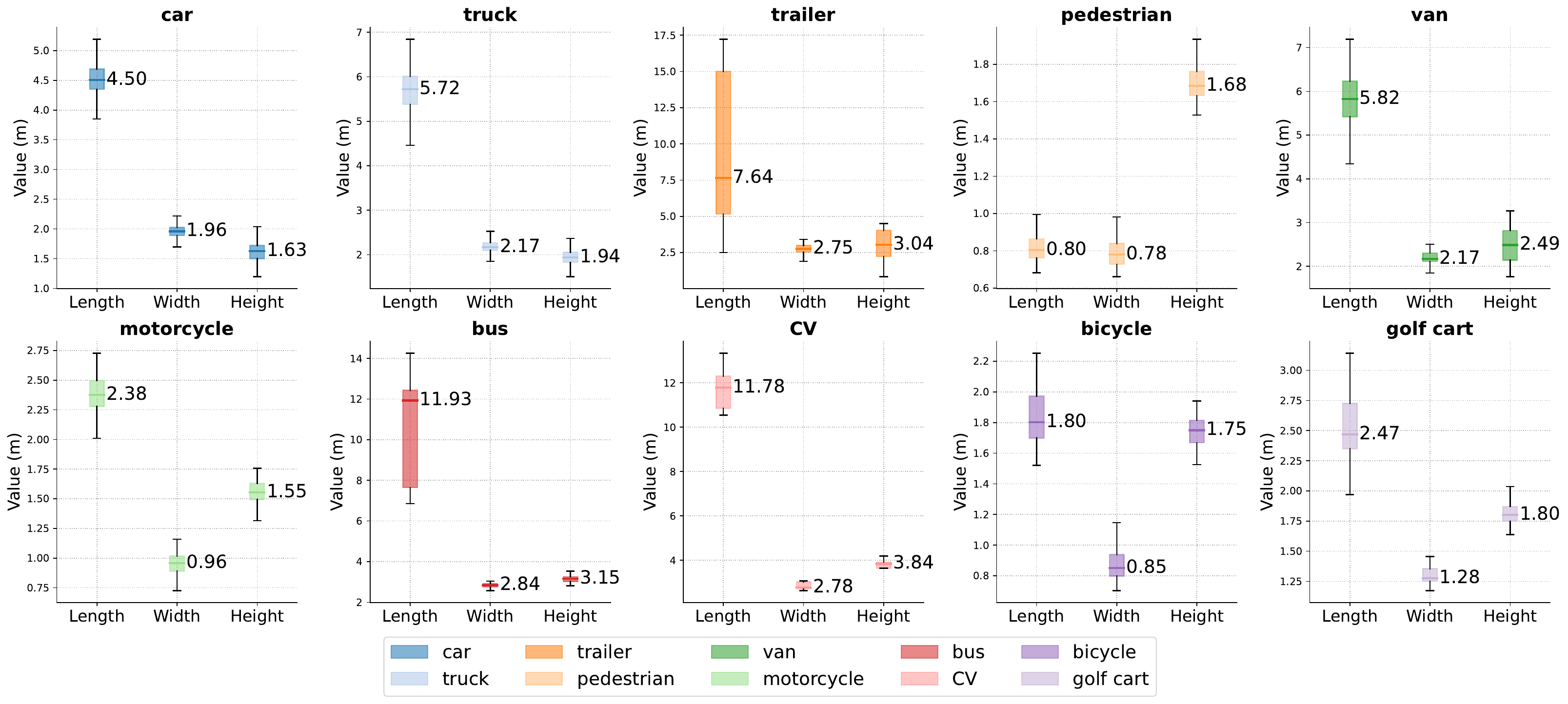}
  \caption{Distribution of object dimensions (length, width, height) for each class in RESOLVE dataset. The box plots depict the intra-class statistical spread and emphasize inter-class scale variations.}
  \label{fig:dims_classes}
\end{figure}

\section{Dataset Analysis}
\subsection{Class Sizes}
As shown in~\cref{fig:dims_classes}, we perform a statistical analysis of the 3D dimensions of 10 categories in our dataset. Based on the scale differences exhibited by different categories, we can further classify them into superclasses: pedestrians, small vehicles (motorcycles, bicycles, golf carts), medium-sized vehicles (cars, trucks, vans), and large vehicles (buses, trailers, construction vehicles). Meanwhile, the size distribution of each category shows a certain degree of intra-class dispersion, with trailers exhibiting the most significant length fluctuations. This is also reflected in the LiDAR feature visualization, where trailers show larger intra-class variance and a more dispersed distribution, reflecting the diversity of trailer types in terms of vehicle structure and shape.

\begin{figure}[t]
  \centering
   \includegraphics[width=\linewidth]{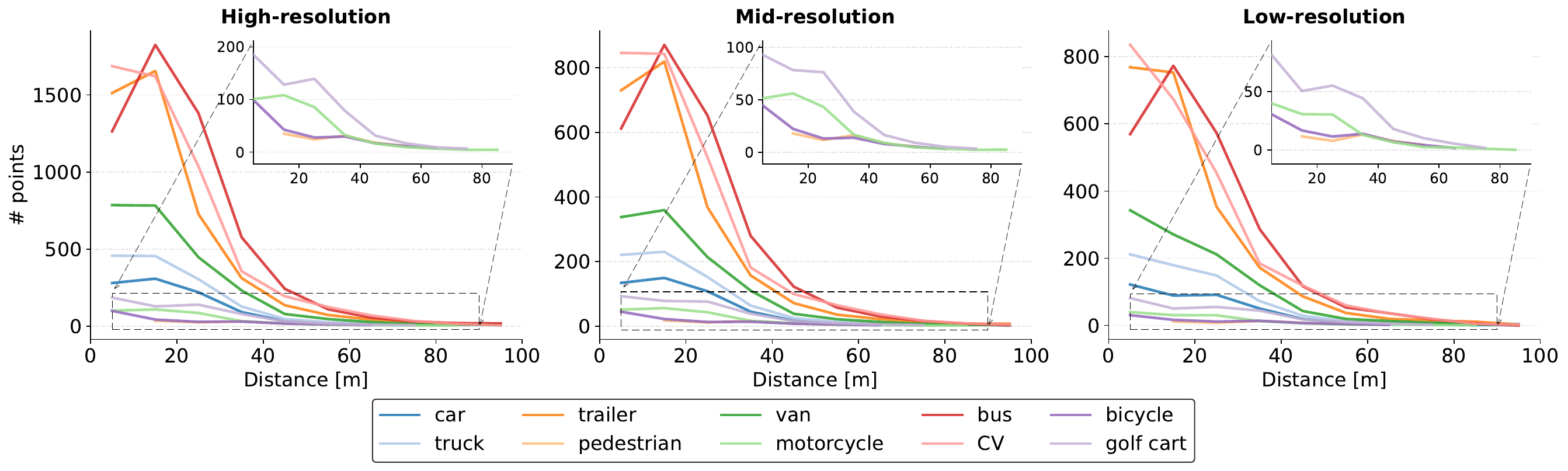}
  \caption{Relationship between the number of points within the bounding box and the distance from the object to the center of the intersection at three resolutions. }
  \label{fig:points_distance}
\end{figure}

\begin{figure}[t]
  \centering
   \includegraphics[width=\linewidth]{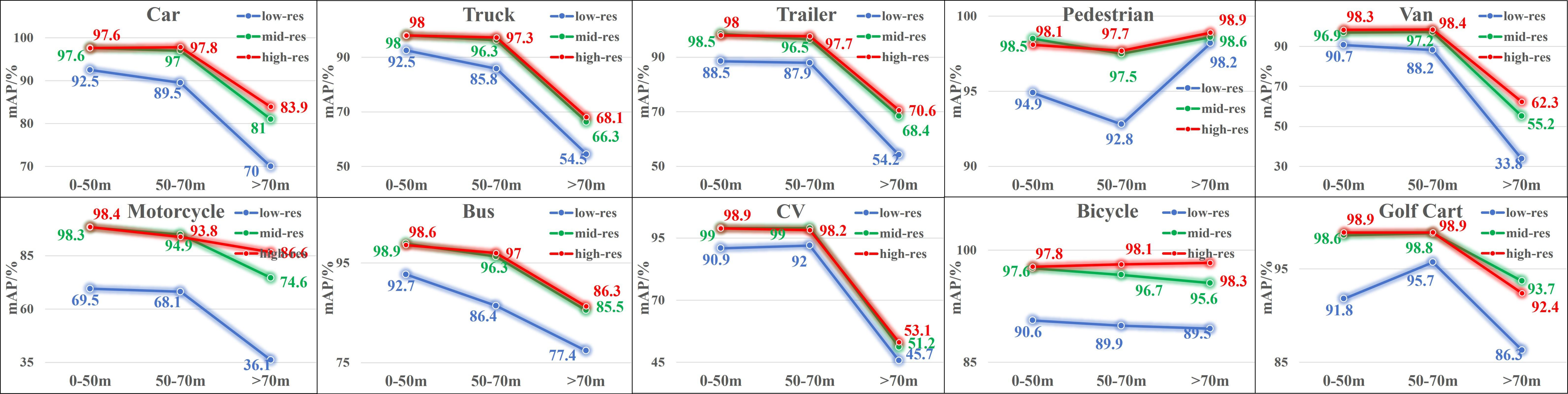}
  \caption{Distance-wise mAP of different classes under three resolutions. We use Voxel Mamba~\cite{voxelmamba} as the detection model. Increasing LiDAR resolution can improve mAP, and this advantage is more pronounced at long distances where the number of in-box points is close to the lower limit. }
  \label{fig:dist_class_map}
\end{figure}

\subsection{Class Distances}
As shown in~\cref{fig:dist_class_map}, we evaluate the performance of each class across different distance ranges. The results show that as the distance from the target to the center of the intersection increases, the overall detection accuracy decreases, with only slight fluctuations in some categories. It is consistent with the decrease trend of the number of in-box points in~\cref{fig:points_distance}, indicating a decline in the geometry and observability of distant targets. For medium to large vehicles, resolution has a limited impact within 50 meters, but beyond 70 meters, the impact becomes crucial as the number of point clouds decreases. For small targets, especially bicycles and motorcycles, the mAP drops more significantly at low resolution due to the thresholding effect of sparse point clouds, while high resolution partially mitigates this problem by increasing the number of point clouds.

\begin{figure}[!ht]
  \centering
   \includegraphics[width=\linewidth]{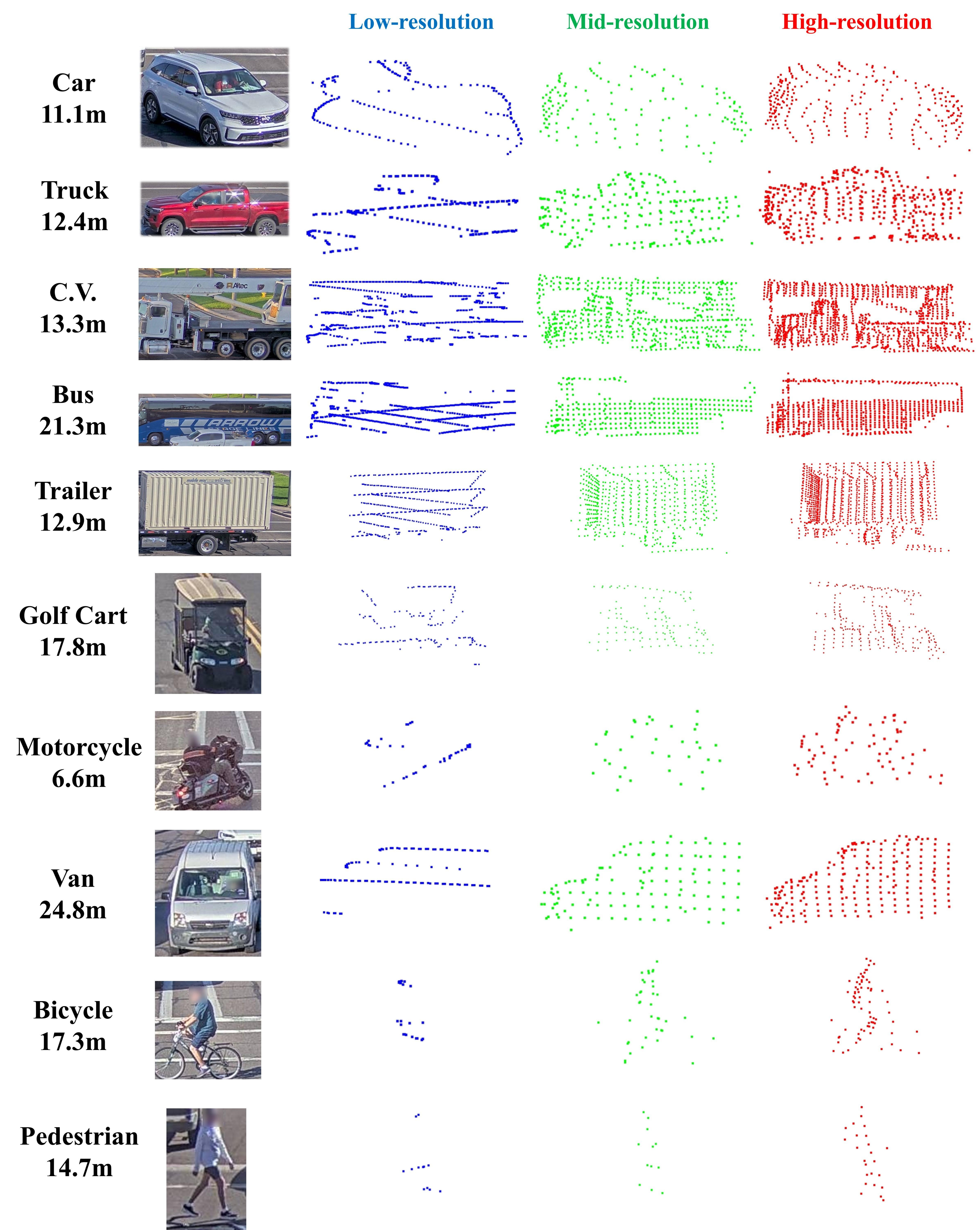}
  \caption{Comparison of point cloud shapes for 10 traffic participants in our RESOLVE dataset at three LiDAR resolutions, including the notes of the distance from the intersection center.}
  \label{fig:class_gt_pc}
\end{figure}

\subsection{Class Point Clouds}
As shown in~\cref{fig:class_gt_pc}, we visualize and compare the point cloud shapes of 10 typical traffic participants in the RESOLVE dataset at three LiDAR resolutions, and label the distance from the intersection center next to each sample image. The results show that the point cloud density significantly increases with increasing resolution, and the target geometry gradually evolves from sparse outlines to more complete three-dimensional information, enhancing the ability to represent details. Meanwhile, different categories are significantly affected by object sizes. Large vehicles (such as buses, vans, and trailers) maintain relatively stable structural representations at all resolutions, while small targets and vulnerable traffic participants (such as pedestrians, bicycles, motorcycles, and golf carts) are more prone to point cloud discrepancies and structural gaps under low resolution or longer distance conditions, increasing observational uncertainty.

\section{Experiments}
\subsection{Metrics}
\subsubsection{3D Object Detection \& Cooperative Perception.} We evaluate 3D object detection using mean Average Precision (mAP). Following the nuScenes~\cite{nuscenes} official protocol, predictions are matched to ground truth by the 2D center distance in the BEV plane. For each class, detections are ranked by confidence and greedily matched one-to-one to unmatched ground-truth boxes within a distance threshold to form TPs. Unmatched detections are FPs. AP is computed from the resulting precision-recall curve, and mAP is obtained by averaging AP over classes and over the set of distance thresholds as shown in~\cref{eq:map}.

\begin{equation}
\mathrm{mAP}
=
\frac{1}{|\mathcal{C}|}\sum_{c\in\mathcal{C}}
\left(
\frac{1}{|\mathcal{D}|}\sum_{\tau\in\mathcal{D}} \mathrm{AP}_c(d)
\right)
\label{eq:map}
\end{equation}

where:
\begin{itemize}
    \item $\mathcal{C}$ denotes the set of classes.
    \item $\mathcal{D}$ denotes the set of 2D center-distance thresholds used to compute AP.
    \item $\mathrm{AP}_c(d)$ is the Average Precision for class $c$ computed under threshold $d$ from the corresponding PR curve.
\end{itemize}
\subsubsection{Multi-object Tracking.} We evaluate multi-object tracking using AMOTA and AMOTP. These metrics are robust to threshold choices because they average the classic scores MOTA and MOTP over a predefined set of operating points such as confidence thresholds or recall levels. At each operating point, MOTA measures tracking accuracy by accounting for false negatives, false positives, and identity switches, and normalizing the total error by the number of ground-truth objects over the whole sequence. MOTP measures tracking precision by averaging the localization error over all correctly matched prediction and ground-truth pairs, using the BEV center distance as the error measure. AMOTA and AMOTP are then obtained by averaging MOTA and MOTP across all operating points as shown in~\cref{eq:amota,eq:amotp},respectively.

\begin{equation}
\mathrm{AMOTA}
=
\frac{1}{|\mathcal{K}|}\sum_{k\in\mathcal{K}}
\left(
1-\frac{\sum_{t}\big(\mathrm{FN}_{t,k}+\mathrm{FP}_{t,k}+\mathrm{IDS}_{t,k}\big)}
{\sum_{t}\mathrm{GT}_{t}}
\right)
\label{eq:amota}
\end{equation}

where:
\begin{itemize}
    \item $\mathcal{K}$ denotes the set of operating points.
    \item $\mathrm{FN}_{t,k}$ is the number of false negatives at time $t$ under operating point $k$.
    \item $\mathrm{FP}_{t,k}$ is the number of false positives at time $t$ under operating point $k$.
    \item $\mathrm{IDS}_{t,k}$ is the number of identity switches at time $t$ under operating point $k$.
    \item $\mathrm{GT}_{t}$ is the number of ground-truth objects at time $t$.
\end{itemize}

\begin{equation}
\mathrm{AMOTP}
=
\frac{1}{|\mathcal{K}|}\sum_{k\in\mathcal{K}}
\left(
\frac{\sum_{t}\sum_{(i,j)\in\mathcal{C}_{t,k}} d\!\left(\hat{x}_{t,i},x_{t,j}\right)}
{\sum_{t}\left|\mathcal{C}_{t,k}\right|}
\right)
\label{eq:amotp}
\end{equation}

where:
\begin{itemize}
    \item $\mathcal{K}$ denotes the set of operating points.
    \item $\mathcal{C}_{t,k}$ denotes the set of correct associations at time $t$ under operating point $k$.
    \item $d(\hat{x}_{t,i},x_{t,j})$ denotes the localization error between prediction and ground truth at time $t$, such as the Euclidean distance.
\end{itemize}

\subsection{Implementation Details}
\subsubsection{3D Object Detection.} All detection models are implemented in the OpenPCDet framework~\cite{openpcdet2020}. All models are trained for 20 epochs with a batch size of 4. We use the Adam optimizer with a learning rate of $1\times10^{-3}$ and weight decay $10^{-2}$. A OneCycle learning rate schedule is applied, where the momentum is cycled from $0.95$ to $0.85$. For BEVFusion~\cite{bevfusion}, since the above learning strategy yields suboptimal performance, we follow the original training setting using the Adam optimizer with a learning rate of $1\times10^{-4}$ and a cosine-annealing learning rate schedule.
To improve late-stage training stability and reduce the distribution shift introduced by strong augmentations, we disable ground-truth sampling augmentation during the last 4 epochs. 

All training is performed on four NVIDIA RTX PRO 6000 BLACKWELL GPUs, while inference and computation are performed on a single GPU with a batch size of 1. All reported results are obtained from the checkpoint with the best validation performance. It should be noted that we do not perform sufficient hyperparameter tuning on all models to achieve optimal performance. Therefore, it is understandable that some models underperformed on the test set. We have publicly released the configuration files and trained weights to facilitate replication and further improvement in subsequent research.

\subsubsection{Multi-object Tracking.} For different TBD methods, we largely follow the default settings in the original implementations. To adapt to the 10 categories in our RESOLVE dataset, we additionally complement the missing class-wise thresholds and track lifecycle parameters to enable direct execution and fair comparisons.

\begin{itemize}
   \item AB3DMOT~\cite{AB3DMOT}: Use $dist\_3d$ or $giou\_3d$ to set up class-wise associations. Most categories use the threshold $thres=-0.2$ and the tracking management parameters $min\_hits=1$ and $max\_age=3$.
  \item CenterPoint~\cite{centerpoint}: Velocity-extrapolated center matching with class-wise gating thresholds, and the maximum missing frames is set to $max\_age=3$.
  \item MCTrack~\cite{mctrack}: Hungarian matching using $giou\_3d$ cost, with global filtering $score=0.3$ and $nms\_score=0.08$, plus class-wise cost thresholds and Kalman filter noise parameters.
  \item Poly-MOT~\cite{polymot}: Two-stage Hungarian matching, class-wise score filtering $score=0.2$ and $nms\_score=0.08$, class-wise motion models and lifecycle parameters $max\_age=10$ and $min\_hits=1$.
  \item SimpleTrack~\cite{simpletrack}: Bipartite matching using $giou\_3d$, with $score=0.01$ and $max\_age=2$, enabling NMS with $nms\_score=0.1$ and applying redundancy suppression.
\end{itemize}

\subsubsection{Cooperative Perception.} In early fusion, raw point clouds from cooperative agents are first transformed into a unified coordinate system, followed by joint voxelization. In late fusion, each agent independently performs object detection and transforms its predicted bounding boxes into the ego-agent coordinate system for final aggregation. For intermediate fusion, each agent first extracts BEV features using the same PointPillars encoder~\cite{pointpillars}, and the resulting feature maps are fused before being passed to the detection head. 

The cooperative PointPillars encoder adopts a single-layer PFN with 64 output channels. The BEV backbone consists of blocks with depths of $[3, 5, 8]$, strides of $[2, 2, 2]$, and channel dimensions of $[64, 128, 256]$. Multi-scale upsampling is then applied with output channels of $[128, 128, 128]$. F-Cooper~\cite{F-cooper} performs spatial maxout fusion on BEV feature maps, CoBEVT~\cite{CoBEVT} applies window-based Transformer fusion after feature downsampling, and V2X-ViT~\cite{V2X-ViT} further incorporates agent-level and spatial-level Transformer attention, together with relative temporal encoding and spatial correction matrices.

\subsection{t-SNE Visualization}
\begin{figure}[t]
  \centering
  \begin{subfigure}[t]{0.32\linewidth}
    \centering
    \includegraphics[width=\linewidth]{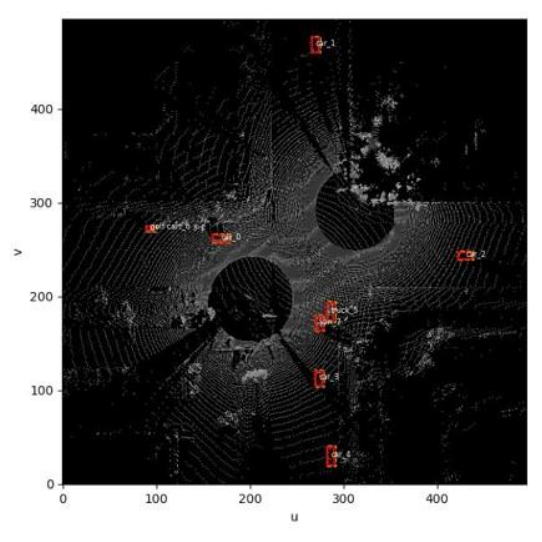}
    \caption{LiDAR BEV feature}
    \label{fig:lidar_bev_f}
  \end{subfigure}
  \begin{subfigure}[t]{0.67\linewidth}
    \centering
    \includegraphics[width=\linewidth]{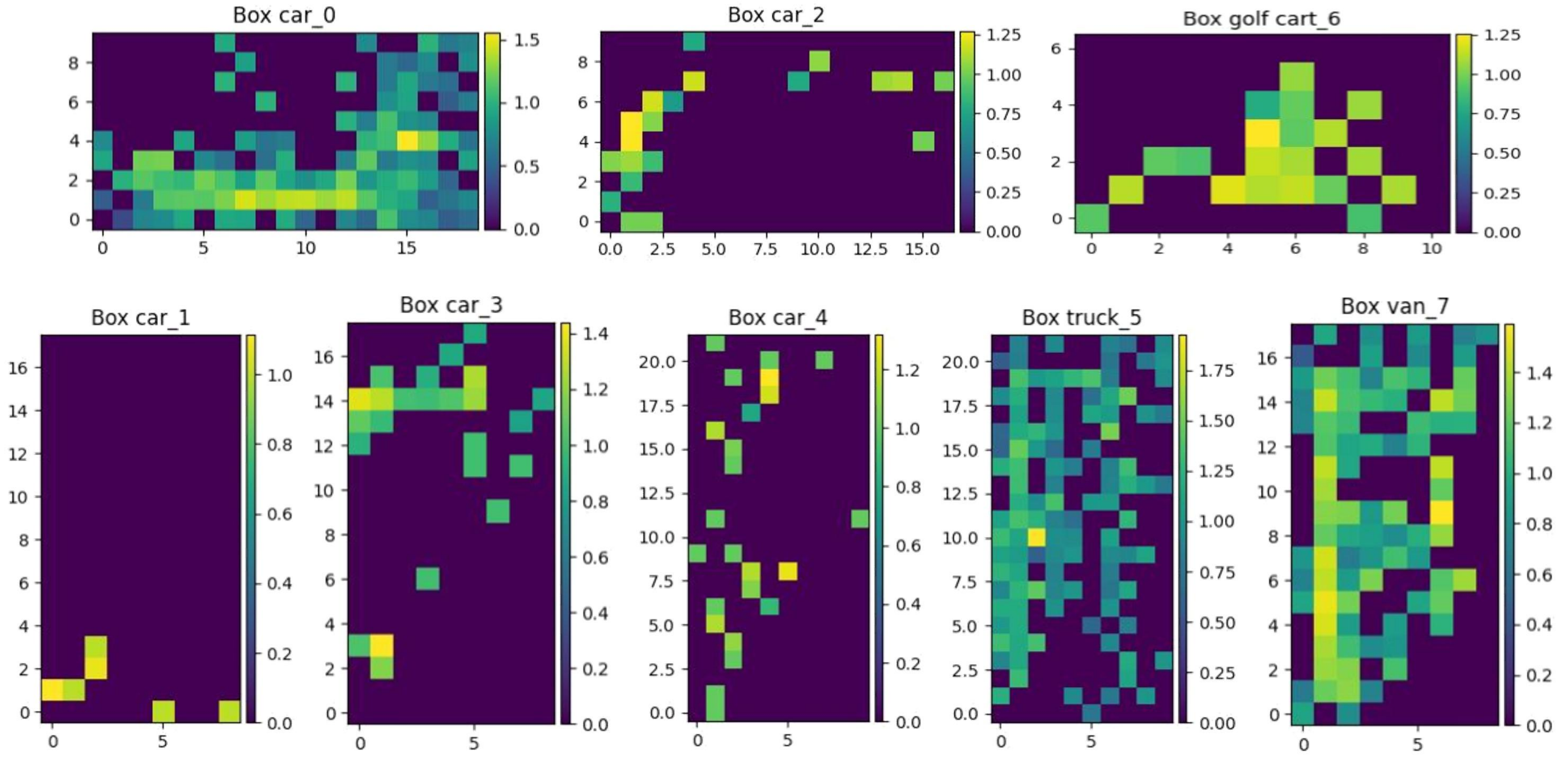}
    \caption{Class feature clipping}
    \label{fig:box_clip}
  \end{subfigure}
  \caption{The illustrations show how we visualize LiDAR backbone features using t-SNE. (a) Sparse convolutional features project onto BEV view. (b) Clipping region features of each object and performing average pooling.}
  \label{fig:t-sne_clip}
\end{figure}

t-distributed stochastic neighbor embedding (t-SNE)~\cite{t-SNE} is a non-linear dimensionality reduction visualization method that maps high-dimensional features to two-dimensional or three-dimensional space while preserving the relative structure of samples in their local neighborhoods as much as possible, thus intuitively demonstrating the clustering and separability of features. \Cref{fig:t-sne_clip} shows how we use t-SNE to visualize the high-dimensional features of the LiDAR branch. First, the features are extracted through sparse convolution, then the height dimension is compressed and projected onto the BEV view. The red box in~\cref{fig:lidar_bev_f} marks the patch where the ground truth is located. To ensure the feature integrity of each object, we further extract the slightly larger yellow box area as the feature of that object.

Assume that there are $n$ ground-truth objects in a frame. After patch cropping, we obtain for each object a feature block of shape $(C,h,w)$ as shown in~\cref{fig:box_clip}, where $h$ and $w$ denote the height and width of the feature, respectively. We then apply global average pooling over the spatial dimensions to aggregate the feature block into a $C$-dimensional vector in $\mathbb{R}^{C}$. Next, we collect all feature vectors at the object-level across frames to form a set of samples $\mathbf{X}\in\mathbb{R}^{N_{\text{total}}\times C}$. After standardization, we apply Principal Component Analysis (PCA) to reduce the feature dimension to $50$ for improved stability, and then feed the reduced features into t-SNE to obtain a 2D embedding $\mathbf{Y}\in\mathbb{R}^{N_{\text{total}}\times 2}$. Finally, we visualize $\mathbf{Y}$ and color-code the samples by their object categories.

\subsection{Further Benchmark Results} 
\subsubsection{3D Object Detection.} As shown in~\cref{tab:class_low_det_result,tab:class_mid_det_result,tab:class_high_det_result}, we present the specific detection results for each category at three LiDAR resolutions. Overall performance generally improves with increasing resolution, with more significant improvements for small, weak-echo objects such as pedestrians, motorcycles, and bicycles. Regarding models, multi-modal methods outperform at low resolutions, with UniTR~\cite{unitr} performing best in most categories. At medium to high resolutions, unimodal detectors further demonstrate their performance, with the Mamba~\cite{voxelmamba,lion} architecture achieving leading or near-optimal results across multiple categories. Mainstream vehicle categories such as cars show stable and high performance, while trailers and vans remain relatively low due to differences in shape and size, which warrants further improvement.

\begin{table}[t]
\centering
\caption{Detection results for each class at low LiDAR resolution. Best in \textcolor{red}{red}.}
\label{tab:class_low_det_result}
\setlength{\tabcolsep}{2.5pt}
\renewcommand{\arraystretch}{1.1}
\scriptsize
\begin{tabular}{c c *{10}{c}}
\toprule
\multirow{2}{*}{Model} & \multirow{2}{*}{Modality} & \multicolumn{10}{c}{mAP $\uparrow$}\\
\cmidrule(lr){3-12}
& &
car & truck & cv & bus & trailer & van & motor & bicycle & ped & gc  \\
\midrule
PointPillars~\cite{pointpillars}    & L & 87.0 & 81.7 & 84.7 & 81.3 & 73.6 & 61.7 & 56.8 & 80.6 & 58.7 & 84.5 \\
SECOND~\cite{second}                & L & 85.7 & 80.0 & 82.1 & 77.5 & 67.8 & 44.3 & 61.7 & 87.0 & 25.4 & 70.1 \\
CenterPoint~\cite{centerpoint}      & L & 85.9 & 79.6 & 81.8 & 83.1 & 74.9 & 70.9 & 66.1 & 84.5 & 84.0 & 88.5 \\
TransFusion-L~\cite{transfusion}    & L & 85.1 & 80.1 & 83.6 & 84.1 & 76.7 & 77.7 & 65.9 & 86.5 & 93.1 & 92.1 \\
VoxSeT~\cite{vst}                   & L & 90.0 & 86.0 & 87.2 & 87.0 & 81.8 & 82.1 & 73.6 & 94.1 & 97.1 & 94.9 \\
DSVT~\cite{dsvt}                    & L & 88.7 & 83.9 & 88.0 & 85.4 & 77.5 & 79.1 & 72.3 & 93.1 & 96.8 & 94.2 \\
Voxel Mamba~\cite{voxelmamba}       & L & 90.0 & 85.4 & \textcolor{red}{\uline{97.1}} & 88.5 & 79.0 & 82.4 & 66.0 & 90.7 & 95.1 & 93.2 \\
LION~\cite{lion}                    & L & 90.2 & 86.4 & 88.6 & 87.7 & 81.4 & 83.5 & 75.8 & \textcolor{red}{\uline{94.8}} & \textcolor{red}{\uline{97.5}} & 95.2 \\
\midrule
BEVFusion~\cite{bevfusion}         & LC & 89.1 & 84.8 & 84.6 & 88.0 & 81.5 & 78.3 & 73.7 & 93.7 & 95.2 & 94.2 \\
UniTR~\cite{unitr}                 & LC & \textcolor{red}{\uline{91.0}} & \textcolor{red}{\uline{87.1}} & 86.8 & \textcolor{red}{\uline{90.8}} & \textcolor{red}{\uline{85.1}} & \textcolor{red}{\uline{84.9}} & \textcolor{red}{\uline{80.4}} & 94.0 & 96.8 & \textcolor{red}{\uline{96.7}}\\
\bottomrule
\end{tabular}
\end{table}

\begin{table}[t]
\centering
\caption{Detection results for each class at mid LiDAR resolution. Best in \textcolor{red}{red}.}
\label{tab:class_mid_det_result}
\setlength{\tabcolsep}{2.5pt}
\renewcommand{\arraystretch}{1.1}
\scriptsize
\begin{tabular}{c c *{10}{c}}
\toprule
\multirow{2}{*}{Model} & \multirow{2}{*}{Modality} & \multicolumn{10}{c}{mAP $\uparrow$}\\
\cmidrule(lr){3-12}
& &
car & truck & cv & bus & trailer & van & motor & bicycle & ped & gc  \\
\midrule
PointPillars~\cite{pointpillars}    & L & 94.4 & 87.8 & 92.0 & 84.4 & 73.6 & 63.1 & 79.9 & 79.6 & 54.1 & 88.0 \\
SECOND~\cite{second}                & L & 93.5 & 89.0 & 90.7 & 85.8 & 74.9 & 49.4 & 91.5 & 92.6 & 24.2 & 74.4 \\
CenterPoint~\cite{centerpoint}      & L & 79.3 & 84.7 & 86.2 & 90.3 & 81.4 & 81.0 & 86.9 & 89.0 & 82.0 & 92.3 \\
TransFusion-L~\cite{transfusion}    & L & 88.6 & 84.1 & 86.1 & 89.0 & 79.9 & 83.8 & 83.5 & 85.7 & 89.3 & 95.0 \\
VoxSeT~\cite{vst}                   & L & 93.7 & 87.7 & 89.0 & 94.3 & 85.4 & 77.5 & 86.1 & 94.8 & 85.7 & 96.9 \\
DSVT~\cite{dsvt}                    & L & 95.2 & 91.4 & 92.2 & 95.9 & 88.7 & 88.5 & 93.9 & 98.0 & 98.1 & 98.7 \\
Voxel Mamba~\cite{voxelmamba}       & L & 96.1 & \textcolor{red}{\uline{93.1}} & \textcolor{red}{\uline{93.0}} & \textcolor{red}{\uline{96.4}} & 90.0 & 90.9 & 94.7 & 97.7 & 98.2 & 98.5 \\
LION~\cite{lion}                    & L & \textcolor{red}{\uline{96.3}} & 93.0 & 92.7 & 96.3 & \textcolor{red}{\uline{90.1}} & \textcolor{red}{\uline{91.5}} & \textcolor{red}{\uline{96.2}} & 98.0 & 98.3 & \textcolor{red}{\uline{98.8}} \\
\midrule
BEVFusion~\cite{bevfusion}          & LC & 95.8 & 92.4 & 76.0 & 95.6 & 89.5 & 88.8 & 94.4 & 98.7 & \textcolor{red}{\uline{98.6}} & 98.5 \\
UniTR~\cite{unitr}                  & LC & 95.7 & 91.8 & 91.5 & 95.5 & 89.8 & 89.6 & 94.4 & \textcolor{red}{\uline{98.7}} & 98.2 & 98.5 \\
\bottomrule
\end{tabular}
\end{table}

\begin{table}[!t]
\centering
\caption{Detection results for each class at high LiDAR resolution. Best in \textcolor{red}{red}.}
\label{tab:class_high_det_result}
\setlength{\tabcolsep}{2.5pt}
\renewcommand{\arraystretch}{1.1}
\scriptsize
\begin{tabular}{c c *{10}{c}}
\toprule
\multirow{2}{*}{Model} & \multirow{2}{*}{Modality} & \multicolumn{10}{c}{mAP $\uparrow$}\\
\cmidrule(lr){3-12}
& &
car & truck & cv & bus & trailer & van & motor & bicycle & ped & gc  \\
\midrule
PointPillars~\cite{pointpillars}   & L & 93.1 & 87.3 & 92.0 & 89.4 & 75.0 & 61.0 & 82.0 & 80.5 & 56.7 & 88.8 \\
SECOND~\cite{second}               & L & 95.3 & 90.6 & 91.7 & 85.9 & 78.3 & 50.3 & 94.0 & 94.5 & 24.0 & 75.0 \\
CenterPoint~\cite{centerpoint}     & L & 90.2 & 85.4 & 87.9 & 90.2 & 83.2 & 82.5 & 89.0 & 91.3 & 82.1 & 92.1 \\
TransFusion-L~\cite{transfusion}   & L & 90.0 & 85.9 & 88.7 & 91.7 & 84.4 & 87.8 & 87.2 & 87.3 & 92.7 & 95.2 \\
VoxSeT~\cite{vst}                  & L & 94.2 & 90.2 & 92.3 & 93.7 & 87.8 & 85.9 & 87.2 & 85.7 & 85.5 & 97.9 \\
DSVT~\cite{dsvt}                   & L & 96.0 & 92.4 & 92.4 & 96.1 & 89.6 & 89.8 & 95.6 & 97.6 & 97.6 & 98.8 \\
Voxel Mamba~\cite{voxelmamba}      & L & 96.8 & \textcolor{red}{\uline{93.8}} & 93.1 & 96.6 & 91.0 & 92.0 & 95.6 & 98.6 & 98.2 & 98.3 \\
LION~\cite{lion}                   & L & \textcolor{red}{\uline{96.9}} & 93.7 & \textcolor{red}{\uline{93.3}} & \textcolor{red}{\uline{96.6}} & \textcolor{red}{\uline{91.3}} & \textcolor{red}{\uline{92.6}} & \textcolor{red}{\uline{98.0}} & \textcolor{red}{\uline{98.8}} & 98.4 & \textcolor{red}{\uline{98.9}} \\
\midrule
BEVFusion~\cite{bevfusion}        & LC & 94.8 & 91.0 & 91.5 & 93.8 & 87.5 & 91.8 & 90.8 & 93.3 & 96.6 & 98.8 \\
UniTR~\cite{unitr}                & LC & 96.7 & 93.0 & 91.7 & 95.7 & 90.7 & 89.7 & 96.0 & 98.3 & \textcolor{red}{\uline{98.8}} & 98.7 \\
\bottomrule
\end{tabular}
\end{table}

In addition to accuracy, we also compare the inference latency and GPU memory usage of various methods at three LiDAR resolutions, as shown in~\cref{tab:latency_memeory}. Increasing resolution leads to higher computational overhead. When the resolution increases from low to medium, the average latency of each method increases by 14.6\%, and GPU memory usage increases by 2\%. When the resolution further increases from medium to high, the average latency increases by 20.4\%, and GPU memory usage increases by 7.5\%. For unimodal models, PointPillars ~\cite{pointpillars} exhibits the fastest inference speed across all resolutions due to its simpler encoding structure, while CenterPoint ~\cite{centerpoint} has the lowest GPU memory usage. The Mamba-based model~\cite{voxelmamba,lion}, compared to the Transformer-based model~\cite{dsvt,vst}, shows approximately a 1.5x increase in latency and a 15\%  increase in GPU memory usage. Compared to unimodal methods, multi-modal methods show an approximately 1.5x increase in average latency, along with a significant increase in GPU memory usage. Among them, BEVFusion~\cite{bevfusion}, due to the introduction of the TransFusion head~\cite{transfusion}, further increases its memory requirements. UniTR, due to its more complex fusion structure, exhibits the highest latency and memory consumption among all methods. We clearly demonstrate the trade-off between accuracy and efficiency. Lightweight models are more suitable for real-time applications. While given sufficient computing resources and memory budget, stronger fusion models typically deliver more robust performance improvements.

\begin{table}[t]
\centering
\caption{Inference latency and memory consumption under different LiDAR resolutions. Increasing resolution leads to higher computational overhead. Best in \textcolor{red}{red}.}
\label{tab:latency_memeory}
\setlength{\tabcolsep}{5pt}
\renewcommand{\arraystretch}{1.1}
\footnotesize
\begin{tabular}{l c c c c c c c}
\toprule
\multirow{2}{*}{Model} & \multirow{2}{*}{Modality} & \multicolumn{3}{c}{Latency (ms) $\downarrow$} & \multicolumn{3}{c}{Memory (MB) $\downarrow$}\\
\cmidrule(lr){3-5} \cmidrule(lr){6-8} 
& &
low & mid & high & low & mid & high  \\
\midrule
PointPillars~\cite{pointpillars}    & L  & \textcolor{red}{\uline{9.7}} & \textcolor{red}{\uline{13.7}} & \textcolor{red}{\uline{16.3}} & 149.1 & 152.0 & 156.7\\
SECOND~\cite{second}                & L  & 14.3 & 19.3 & 25.2 & 53.6 & 72.8 & 90.3\\
CenterPoint~\cite{centerpoint}      & L  & 20.6 & 27.8 & 36.9 & \textcolor{red}{\uline{47.3}} & \textcolor{red}{\uline{66.0}} & \textcolor{red}{\uline{84.4}}\\
TransFusion-L~\cite{transfusion}    & L  & 18.7 & 22.6 & 30.7 & 1198.1 & 1218.6 & 1239.0\\
VoxSeT~\cite{vst}                   & L  & 27.7 & 33.2 & 44.2 & 1341.4 & 1341.4 & 2048.0\\
DSVT~\cite{dsvt}                    & L  & 34.8 & 43.7 & 48.1 & 1423.4 & 1433.6 & 1433.6\\
Voxel Mamba~\cite{voxelmamba}       & L  & 77.7 & 87.0 & 103.8 & 1822.7 & 1833.0 & 1843.2\\
LION~\cite{lion}                    & L  & 83.7 & 97.1 & 120.4 & 1587.2 & 1648.6 & 1699.8\\
\midrule
BEVFusion~\cite{bevfusion}         & LC  & 63.9 & 73.1 & 92.6 & 1433.6 & 1454.1 & 1474.6\\
UniTR~\cite{unitr}                 & LC  & 128.2 & 131.9 & 143.1 & 3594.2 & 3686.4 & 3809.3 \\
\bottomrule
\end{tabular}
\end{table}

\subsubsection{Multi-object Tracking.} Tracking-by-detection (TBD) methods are highly sensitive to detector performance. As shown in~\cref{tab:better_tracking_result}, to evaluate the impact of detector quality on tracking, we replace the detector with the more powerful Voxel Mamba~\cite{voxelmamba} and compare various trackers under three LiDAR resolution settings. The results show that after using Voxel Mamba, each tracker achieved stable gains at different resolutions. The most significant improvements are observed at low resolutions, with AMOTA improving by approximately 9.6\% to 26.22\% and AMOTP decreasing by approximately 68.8\% to 84.59\%. Significant improvements are also maintained at medium resolutions, with AMOTA improving by approximately 8.7\% to 20.4\% and AMOTP decreasing by approximately 51.5\% to 83.0\%. At high resolutions, the gains converge relatively, but still show consistent improvement, with AMOTA improving by approximately 5.5\% to 10.3\% and AMOTP decreasing by approximately 38.3\% to 70.7\%. MCTrack~\cite{mctrack} show the largest improvements across all three resolutions.

From the resolution perspective, lower LiDAR resolution yields sparser point clouds and weaker geometric cues, which significantly increases detection noise (higher localization variance and more FN/FP). Since TBD trackers rely on per-frame detections, such errors are amplified into frequent track interruptions and association failures, making tracking performance strongly detector-bounded at low resolution. Therefore, upgrading the detector leads to the largest gains under the low resolution. As the resolution increases, detections become more stable and the tracking bottleneck gradually shifts from detection quality to data association and occlusion handling. Consequently, the marginal benefit of a stronger detector diminishes and the gains converge at high resolution, while still remaining consistent.

\begin{table}[t]
\centering
\setlength{\tabcolsep}{2.6pt} 
\renewcommand{\arraystretch}{1.08} 
\caption{Evaluation results of multi-object tracking methods using Voxel Mamba~\cite{voxelmamba} as detector. The best performance is marked in \textcolor{red}{red}. CV, Tra, Motor, Bic, Ped, GC: Construction Vehicle, Trailer, Motorcycle, Bicycle, Pedestrian, Golf Cart. Res.: Resolution.}
\label{tab:better_tracking_result}
\resizebox{\linewidth}{!}{
\begin{tabular}{l c *{10}{c} *{10}{c}}
\toprule
\multirow{2}{*}{Model} & \multirow{2}{*}{Res.} &
\multicolumn{10}{c}{AMOTA $\uparrow$} &
\multicolumn{10}{c}{AMOTP $\downarrow$} \\
\cmidrule(lr){3-12}\cmidrule(lr){13-22}
& &
Car & Truck & CV & Bus & Tra & Van & Motor & Bic & Ped & GC &
Car & Truck & CV & Bus & Tra & Van & Motor & Bic & Ped & GC \\
\midrule

\multirow{3}{*}{AB3DMOT~\cite{AB3DMOT}} 
& Low   & 65.4 & 67.8 & 75.7 & 55.4 & 54.5 & 67.1 & 9.6 & 36.4 & 67.3 & 50.3 & 15.5 & 20.4 & 28.1 & 22.7 & 91.3 & 26.0 & 138.1 & 104.2 & 16.5 &  12.9\\
& Mid   & 59.1 & 63.5 & 72.1 & 45.9 & 50.4 & 61.2 & 10.6 & 42.6 & 74.2 & 57.4 & 19.1 & 19.1 &26.6  & 20.4 & 94.7 & 24.2 & 134.8 &95.6  & 17.6&14.8 \\
& High  & 59.0 & 63.4 & 72.1 & 45.9 & 50.4 & 61.0 & 10.6 & 42.6 & 74.2 & 57.5 & 19.1 & 19.1 & 26.6 & 20.4 & 94.7 & 24.3 & 134.8 & 95.6 & 17.1 & 14.9 \\

\multirow{3}{*}{CenterPoint~\cite{centerpoint}} 
& Low  & 91.7 & 91.2 & 94.8 & 94.6 & 88.2 & 90.5 & 76.7 & 91.5 & 93.1 & 95.4 & 10.7 & 15.5 & \textcolor{red}{\uline{7.8}} & 11.5 & 19.8 & \textcolor{red}{\uline{14.3}} & 10.6 & 9.1 & 4.2 & \textcolor{red}{\uline{3.1}} \\
& Mid  & 93.3 & 92.5 & 94.6 & 96.5 & 90.2 & 90.5 & \textcolor{red}{\uline{90.9}} & 95.6 & 96.7 & 96.5 & \textcolor{red}{\uline{9.1}} & \textcolor{red}{\uline{13.8}} & 12.3 & \textcolor{red}{\uline{7.8}} & \textcolor{red}{\uline{13.5}} & 18.5 & 8.0 & 2.8 & 3.1 & 7.1 \\
& High & 93.3 & 92.5 & 94.6 & 96.5 & 90.2 & 90.5 & 90.8 & 95.6 & 96.7 & 96.5 & 9.1 & 13.8 & 12.3 & 7.8 & 13.5 & 18.5 & 8.0 & 2.8 & 3.1 & 7.1 \\

\multirow{3}{*}{SimpleTrack~\cite{simpletrack}} 
& Low  & 94.0 & 93.0 & 96.2 & \textcolor{red}{\uline{98.2}} & 90.4 & \textcolor{red}{\uline{92.3}} & 62.7 & 96.0 & 98.2 & 95.4 & 15.1 & 14.7 & 13.5 & 14.1 & 22.6 & 19.3 & 20.3 & 6.0 & 4.1 & 6.2 \\
& Mid  & \textcolor{red}{\uline{94.7}} & \textcolor{red}{\uline{93.1}} & \textcolor{red}{\uline{97.3}} & 96.9 & \textcolor{red}{\uline{93.9}} & 90.3 & 67.2 & \textcolor{red}{\uline{99.3}} & \textcolor{red}{\uline{99.0}} & \textcolor{red}{\uline{98.9}} & 15.1 & 19.0 & 12.8 & 16.9 & 16.8 & 23.3 & 12.4 & 5.4 & \textcolor{red}{\uline{2.9}} & 5.2 \\
& High & 94.6 & 93.1 & 97.3 & 96.9 & 93.9 & 90.3 & 67.2 & 99.3 & 96.2 & 96.4 & 15.1 & 19.0 & 12.8 & 16.9 & 16.8 & 23.3 & 12.3 & 5.4 & 8.1 & 10.4 \\

\multirow{3}{*}{Poly-MOT~\cite{polymot}} 
& Low  & 83.3 & 79.1 & 78.1 & 77.5 & 76.2 & 85.8 & 48.4 & 78.7 & 96.1 & 82.7 & 24.7 & 28.9 & 56.3 & 45.3 & 25.0 & 19.4 & 22.8 & 27.5 & 27.6 & 8.7 \\
& Mid  & 80.8 & 77.5 & 74.1 & 75.3 & 77.4 & 80.6 & 51.9 & 80.6 & 96.5 & 84.2 & 25.2 & 28.3 & 59.6 & 49.6 & 18.7 & 18.4 & 22.2 & 28.3 & 26.8 & 7.0 \\
& High & 80.7 & 77.5 & 74.1 & 75.4 & 77.4 & 84.9 & 51.9 & 80.6 & 93.5 & 84.2 & 25.2 & 28.3 & 59.6 & 49.6 & 18.7 & 18.4 & 22.3 & 28.3 & 31.4 & 7.0 \\

\multirow{3}{*}{MCTrack~\cite{mctrack}} 
& Low  & 80.5 & 79.7 & 83.8 & 77.0 & 73.4 & 76.9 & 45.6 & 71.4 & 83.9 & 88.9 & 15.8 & 15.4 & 12.9 & 10.4 & 19.7 & 19.3 & 15.7 & 3.8 & 9.2 & 8.8 \\
& Mid  & 76.6 & 77.4 & 87.2 & 82.0 & 73.6 & 74.2 & 49.9 & 79.7 & 77.6 & 93.3 & 14.3 & 13.8 & 12.5 & 7.9 & 18.7 & 18.3 & \textcolor{red}{\uline{7.9}} & \textcolor{red}{\uline{2.3}} & 7.8 & 7.0 \\
& High & 76.5 & 75.7 & 87.2 & 82.0 & 73.6 & 74.1 & 49.7 & 79.7 & 76.2 & 93.3 & 14.3 & 13.8 & 12.5 & 7.9 & 18.7 & 18.3 & 7.9 & 2.3 & 7.8 & 7.0 \\
\bottomrule
\end{tabular}
}
\end{table}

\begin{figure}[t]
  \centering
  \begin{subfigure}[t]{0.32\linewidth}
    \centering
    \includegraphics[width=\linewidth]{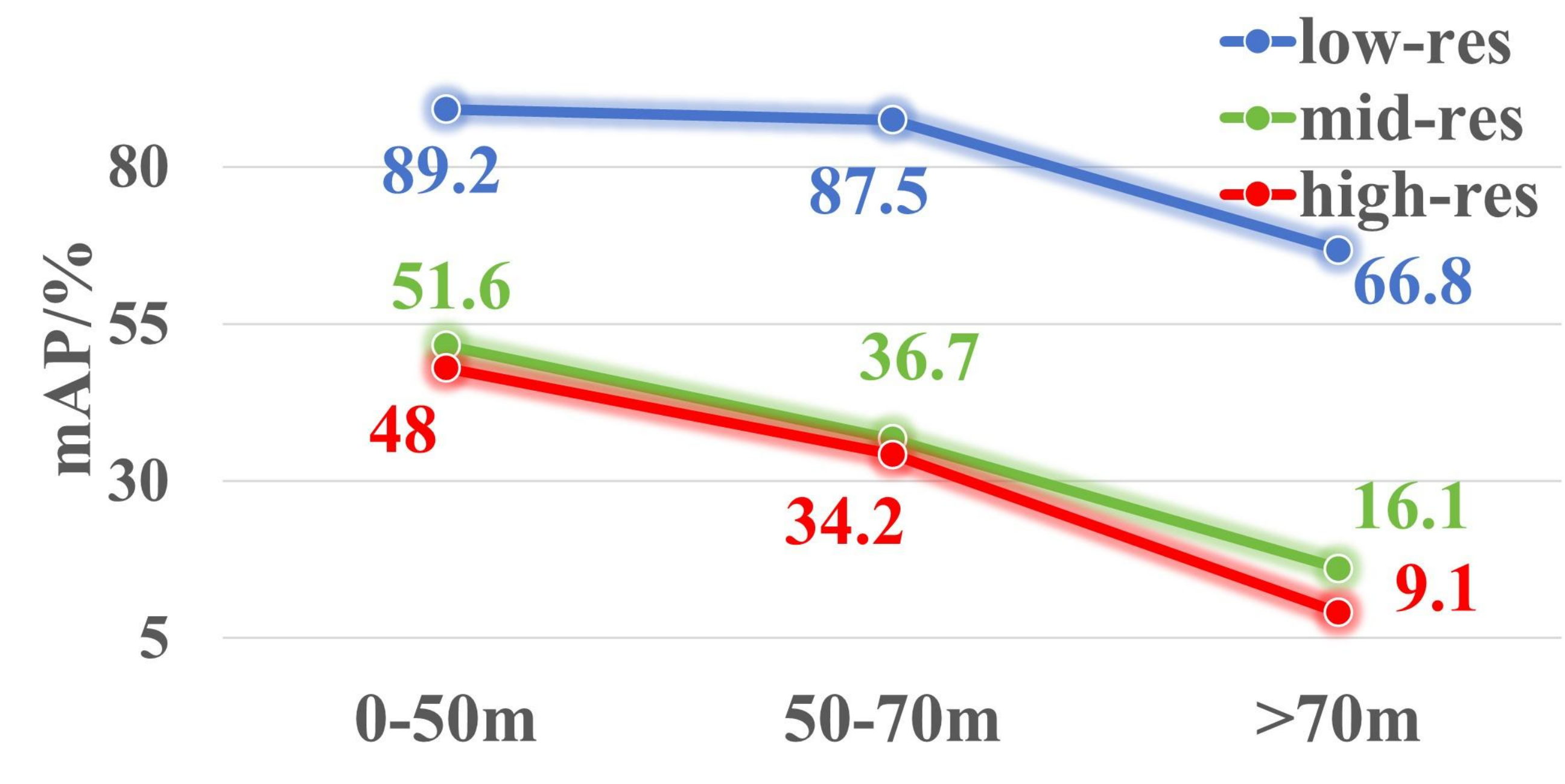}
  \end{subfigure}
  \begin{subfigure}[t]{0.32\linewidth}
    \centering
    \includegraphics[width=\linewidth]{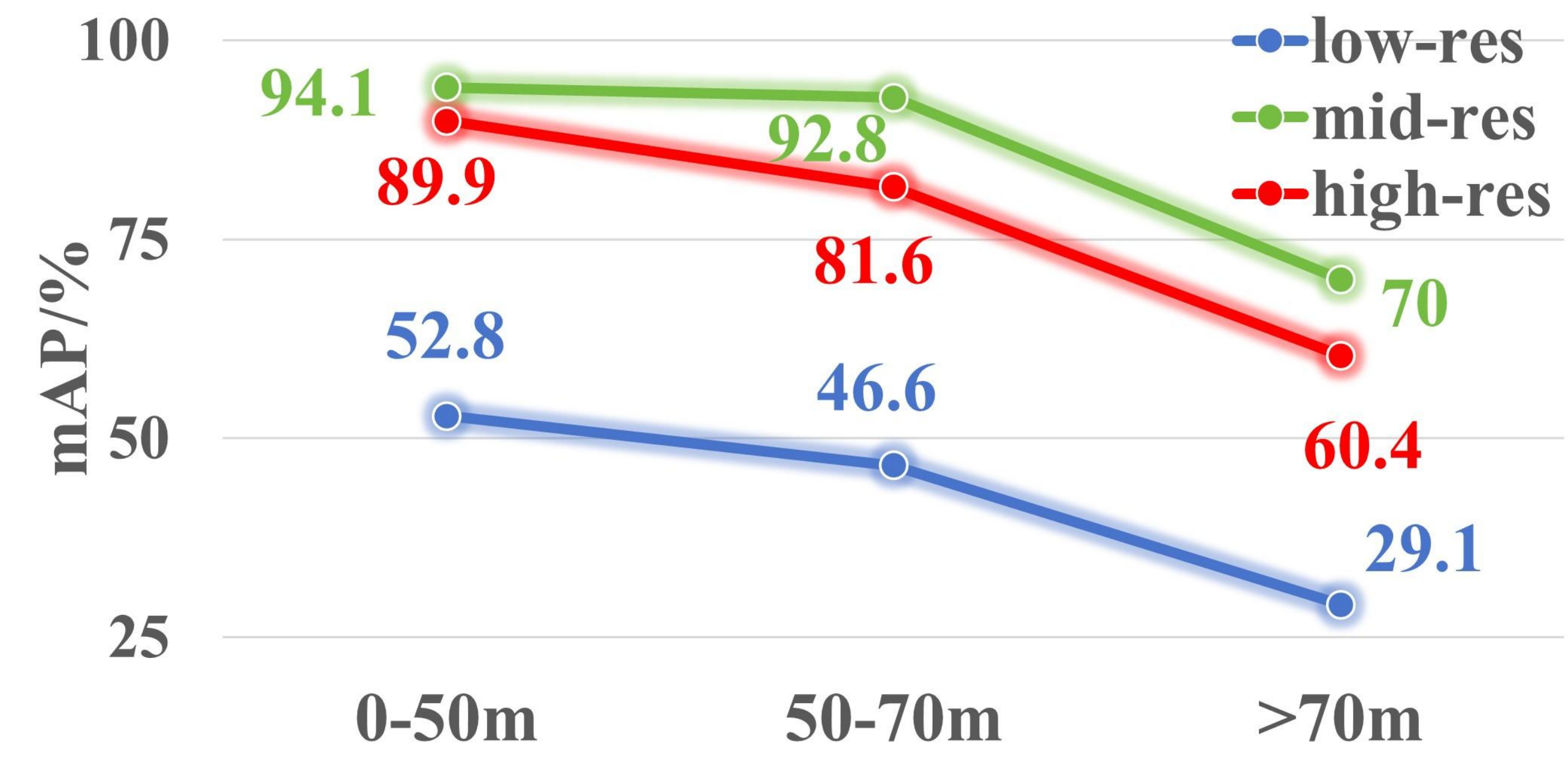}
  \end{subfigure}
  \begin{subfigure}[t]{0.32\linewidth}
    \centering
    \includegraphics[width=\linewidth]{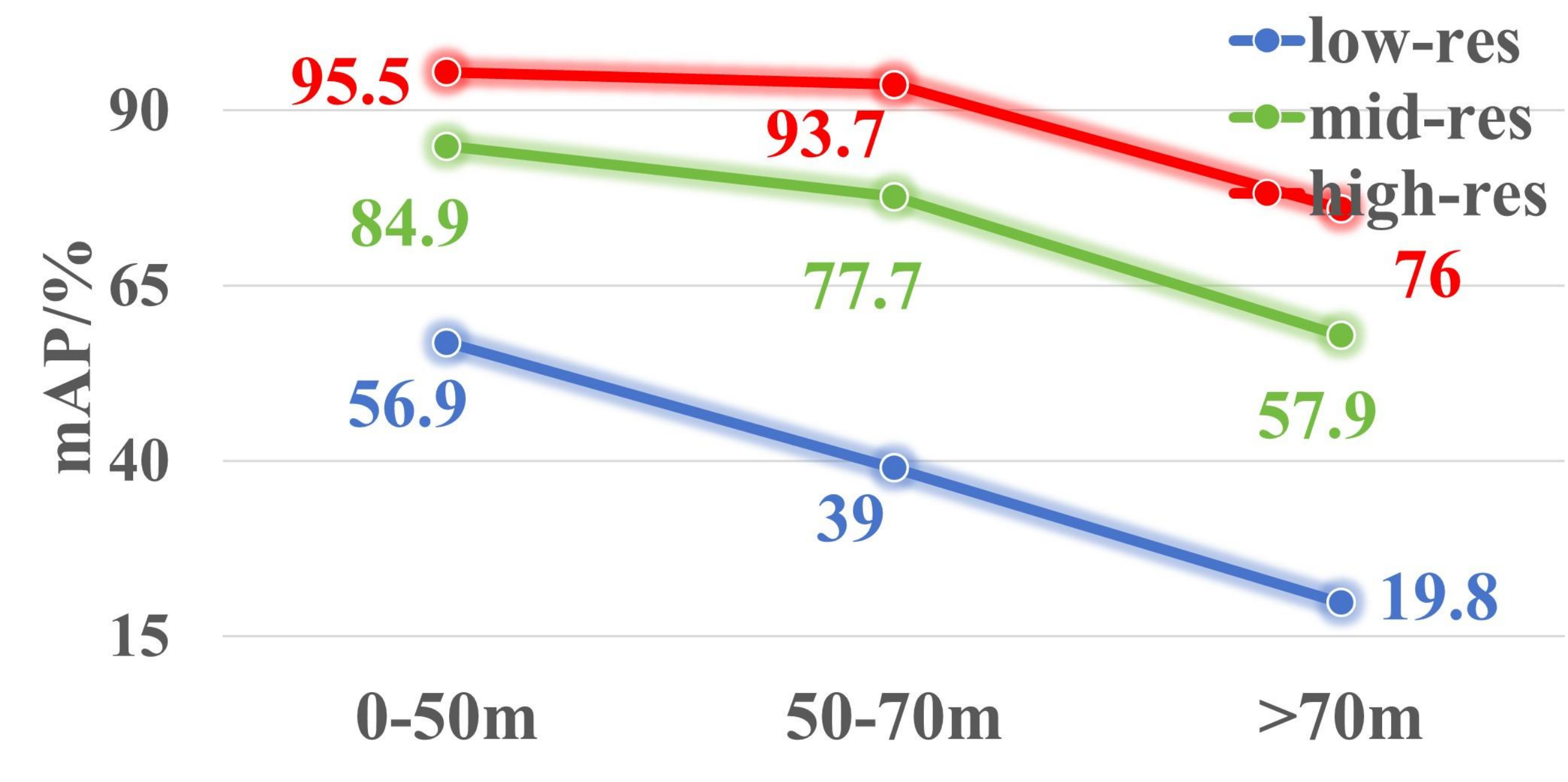}
  \end{subfigure}

   \begin{subfigure}[t]{0.32\linewidth}
    \centering
    \includegraphics[width=\linewidth]{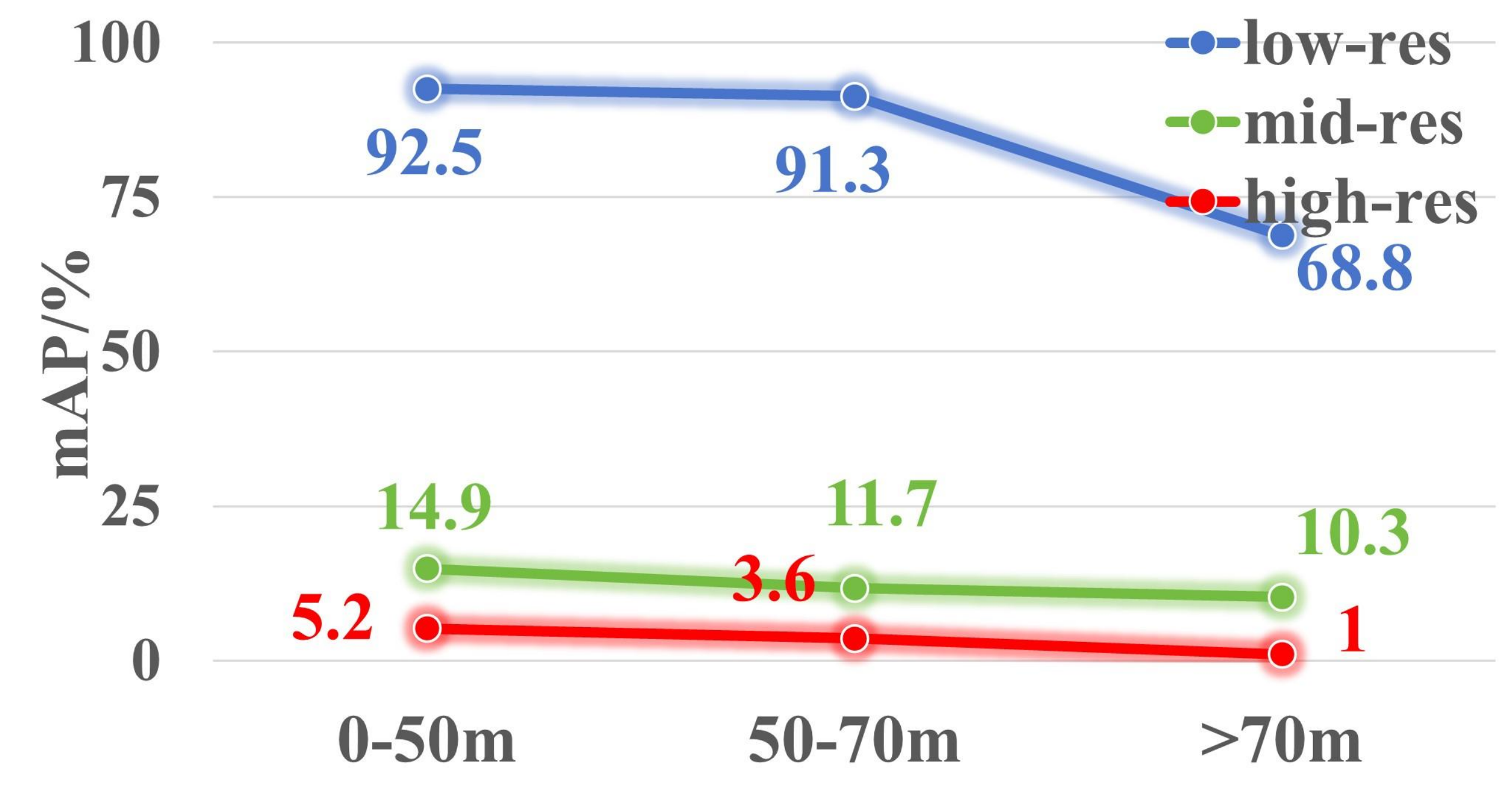}
    \caption{Low-resolution training}
  \end{subfigure}
  \begin{subfigure}[t]{0.32\linewidth}
    \centering
    \includegraphics[width=\linewidth]{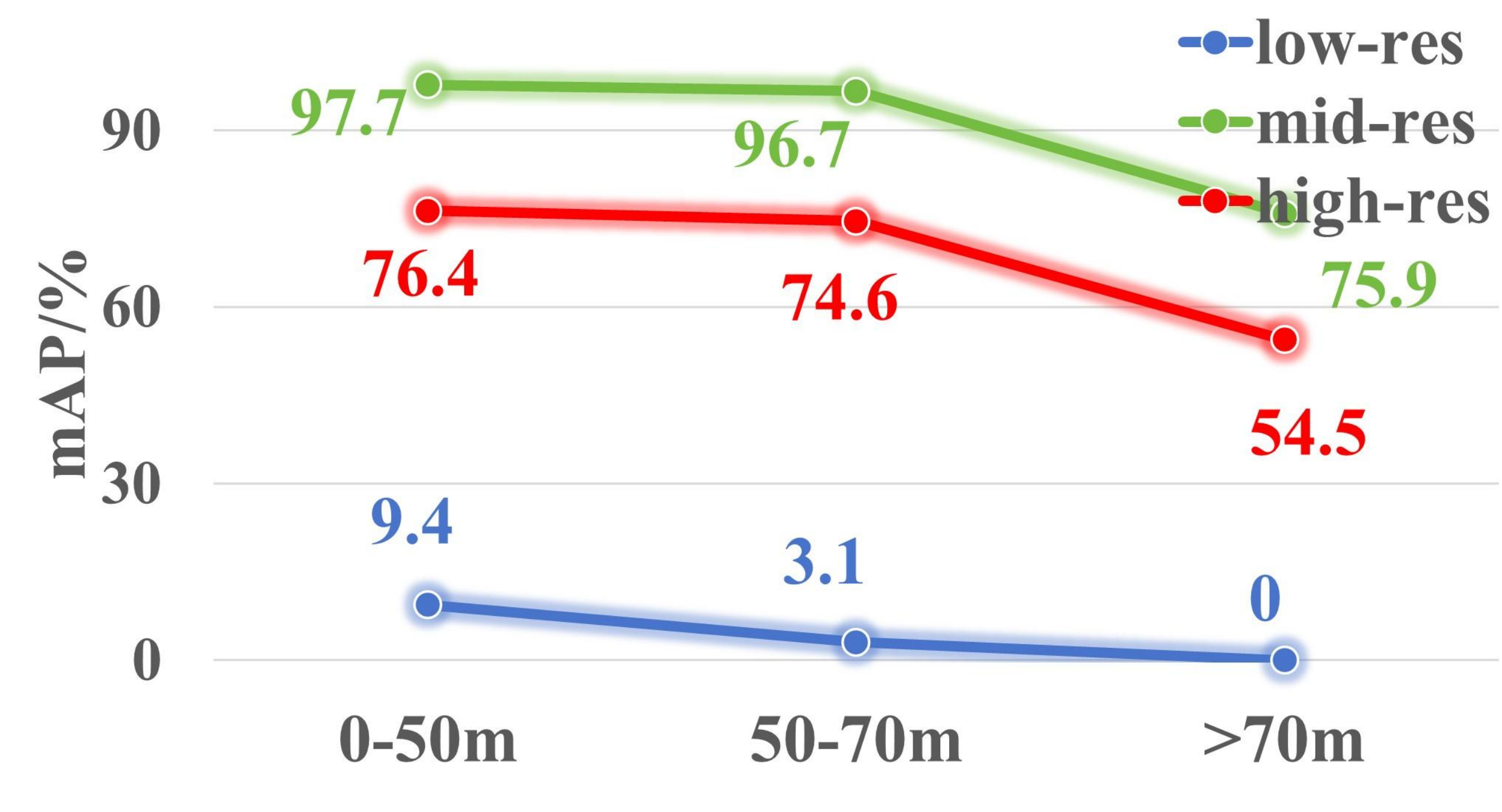}
    \caption{Mid-resolution training}
  \end{subfigure}
  \begin{subfigure}[t]{0.32\linewidth}
    \centering
    \includegraphics[width=\linewidth]{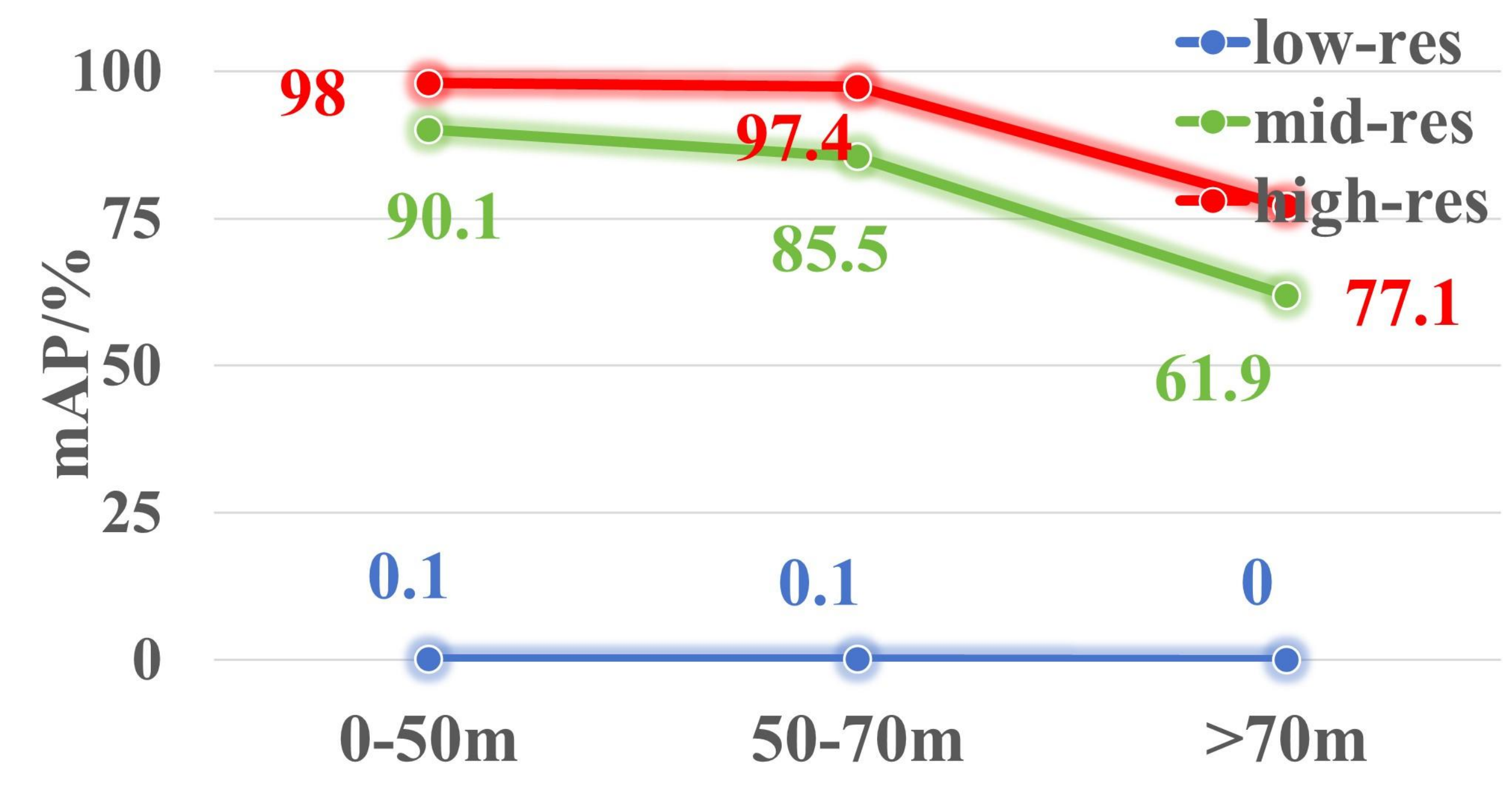}
    \caption{High-resolution training}
  \end{subfigure}
  
  \caption{Multimodal performance degradation caused by different LiDAR resolutions used during training and inference. Top: BEVFusion~\cite{bevfusion} is used as multimodal detector. Bottom: UniTR~\cite{unitr} is used as multimodal detector}
  \label{fig:multi_mismatch}
\end{figure}

\subsection{Ablation Studies}
\subsubsection{Impact of LiDAR Resolution Mismatch.} We compare the robustness of multimodal models with inconsistent LiDAR resolutions during training and inference. As shown in~\cref{fig:multi_mismatch}, the results show that both architectures exhibit a significant resolution domain gap. Performance is optimal when training and inference resolutions are consistent. Once the inference resolution deviates from the training setting, the mAP decreases rapidly with increasing target distance, with the most severe degradation at long distances (>70 m). 

Notably, UniTR~\cite{unitr} is more vulnerable to resolution changes and exhibits significant asymmetry. Catastrophic degradation (even $\mathrm{mAP} \approx 0$)can occur whether the switch is from higher resolution training to low resolution inference or from low resolution training to higher resolution inference. This is primarily because UniTR's Transformer unified token representation is highly sensitive to input discretization sequence. Resolution changes alter token occupancy and sequence distribution, leading to mismatches in attention patterns and cross-modal alignment. Since UniTR relies more heavily on LiDAR tokens as geometric anchors, camera branches struggle to effectively compensate for domain shifts in LiDAR representations, further amplifying performance degradation in long-distance scenes. In contrast, while BEVFusion experiences a significant drop in performance due to resolution mismatch, it still maintains a certain level of usability overall. Overall, the resolution mismatch not only perturbs LiDAR unimodal feature representation but also disrupts cross-modal alignment.

\begin{table}[t]
\centering
\caption{Detection performance of SECOND~\cite{second} under different voxel sizes.}
\label{tab:second_voxel}
\scriptsize
\setlength{\tabcolsep}{4pt}
\renewcommand{\arraystretch}{1.4}
\begin{tabular}{lcccccccccccc}
\hline
\multirow{2}{*}{Voxel size} &
\multicolumn{3}{c}{mAP $\uparrow$} &
\multicolumn{3}{c}{NDS  $\uparrow$} &
\multicolumn{3}{c}{Latency (ms) $\downarrow$} &
\multicolumn{3}{c}{Memory (MB) $\downarrow$} \\
 & Low & Mid & High & Low & Mid & High & Low & Mid & High & Low & Mid & High \\
\hline
$0.1\times0.1$ & 82.0 & 89.5 & 91.0 & 77.0 & 81.8 & 83.2 & 19.9 & 25.9 & 32.9 & 307.8 & 363.4 & 428.0 \\
$0.2\times0.2$ & 74.7 & 83.2 & 85.4 & 71.2 & 76.7 & 78.5 & 17.7 & 22.6 & 29.4 & 96.6 & 125.7 & 156.6 \\
$0.3\times0.3$ & 68.2 & 76.6 & 78.0 & 66.5 & 71.2 & 73.1 & 14.3 & 19.3 & 25.2 & 53.6  & 72.8  & 90.3 \\
\hline
\end{tabular}
\end{table}

\subsubsection{Different Voxel Sizes.} As shown in~\cref{tab:second_voxel}, when the voxel size decreases from 0.3 to 0.1, the model's representation resolution improves, allowing for more thorough preservation of point cloud boundaries and local geometric details. This significantly improves detection accuracy, with mAP increasing by 16.7\%, 16.8\% and 20.2\% at high, medium and low settings, respectively. LiDAR resolution determines the upper limit of observable detail, while voxel size determines the level of detail that the model can utilize. Smaller voxels are helpful for handling dense point clouds, but may generate many empty voxels in sparse inputs, leading to limited gains and higher costs. Therefore, latency increased by 30.6\%, 34.2\%, 39.2\%, and memory usage increased by approximately 4.7x, 5.0x, 5.7x, respectively. Thus, there is a matching relationship between voxel size and LiDAR resolution. Only by reducing the voxel size, which is adapted to the resolution, can a more reasonable balance be achieved between accuracy gain and resource overhead.

\begin{table}[t]
\centering
\caption{Detection performance under real and downsampled 16-beam LiDAR across distances for pedestrians and bicycles.}
\label{tab:ped_bicycle}
\setlength{\tabcolsep}{8pt}
\begin{tabular}{lccccc}
\toprule
\multirow{2}{*}{Source}
& \multicolumn{2}{c}{Pedestrian} 
& \multicolumn{2}{c}{Bicycle} \\
\cmidrule(lr){2-3} \cmidrule(lr){4-5}
 & 
 0--50m & 50--70m 
 & 0--50m & 50--70m \\
\midrule
Real 16-beam                       & 0.95 & 0.95 & 0.96 & 0.89 \\
Real 128-beam$\rightarrow$16-beam  & 0.67 & 0.59 & 0.58 & 0.24 \\
Real 64-beam$\rightarrow$16-beam   & 0.53 & 0.60 & 0.45  & 0.44 \\
\bottomrule
\end{tabular}
\end{table}

\begin{figure}[t]
  \centering
  \begin{subfigure}[t]{0.49\linewidth}
    \centering
    \includegraphics[width=\linewidth]{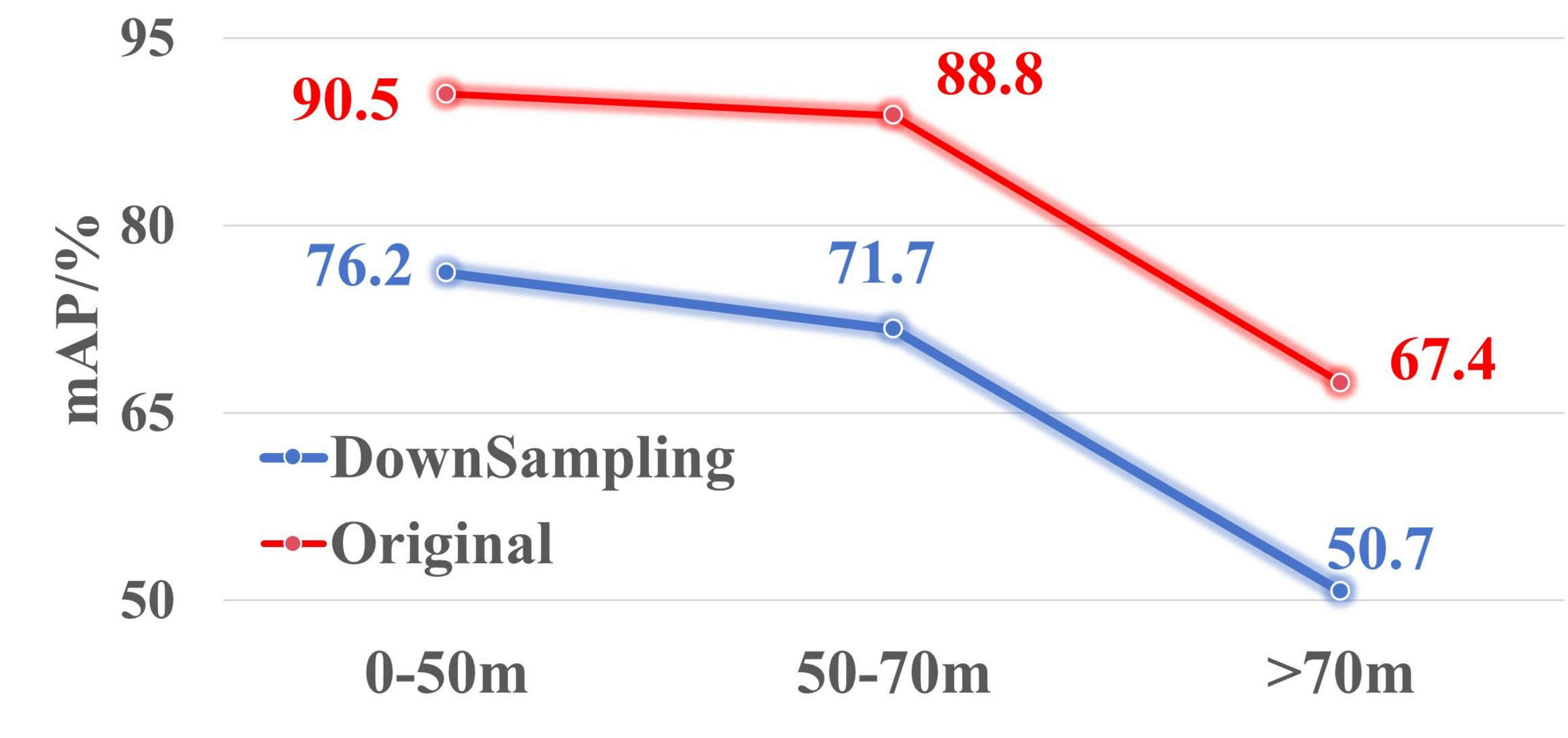}
    \caption{16-beam comparison}
    \label{fig:down16}
  \end{subfigure}
  \begin{subfigure}[t]{0.49\linewidth}
    \centering
    \includegraphics[width=\linewidth]{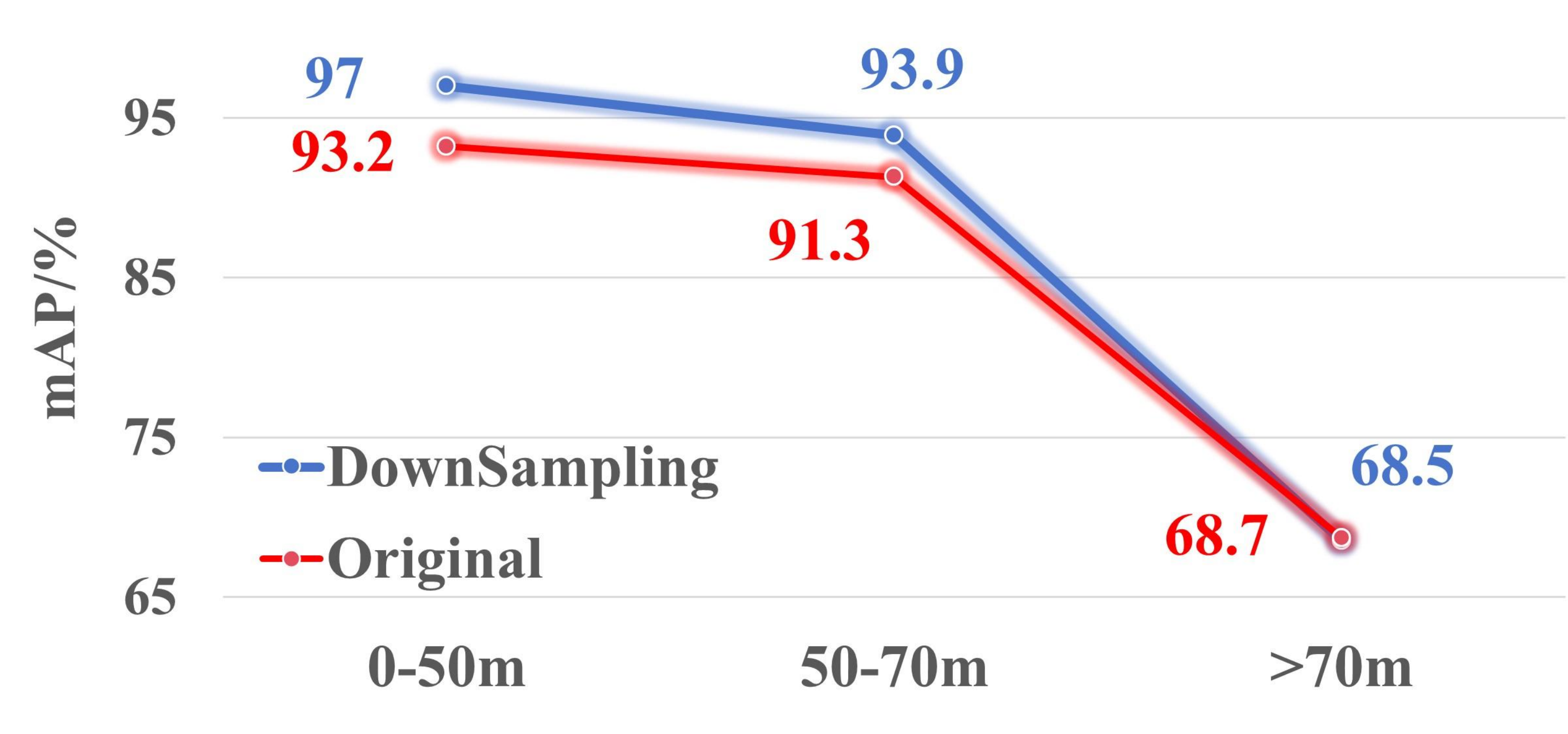}
    \caption{64-beam comparison}
    \label{fig:down64}
  \end{subfigure}
  \caption{Comparison of detection results between 64/16-beam data obtained by downsampling from 128-beam and real-world 64/16-beam data.}
  \label{fig:down_results}
\end{figure}

\subsubsection{Downsampling vs Real-world Beam.} To evaluate whether low-beam point clouds obtained by downsampling from high-beam point clouds can realistically simulate low-resolution LiDAR, we compare the detection results of 64/16-beam data obtained by downsampling from 128-beam with those of real 64/16-beam data. \Cref{fig:down_results} shows that the downsampling strategy exhibits inconsistent and physically unreasonable biases under different beams. In~\cref{fig:down16}, the downsampled 16-beam data is significantly worse than the real 16-beam across the entire range, exhibiting a clearly overly pessimistic estimate. In~\cref{fig:down64}, the downsampled 64-beam data actually achieves higher mAP within 70\,m, indicating an overly optimistic estimate of the real 64-beam. This gap varies across categories and distances, especially for vulnerable road users with smaller physical sizes, as shown in Tab.~\ref{tab:ped_bicycle}. For example, real 16-beam LiDAR achieves 0.95 AP for pedestrians at both distance ranges, whereas 128$\rightarrow$16 and 64$\rightarrow$16 downsampling reduce AP by 29.5\%/37.9\% at 0-50m and 44.2\%/36.8\% at 50-70m, respectively.  The discrepancy is more severe for bicycles, where 128$\rightarrow$16 downsampling causes a 73.0\% AP drop at 50--70m. The same downsampling mechanism produces errors in opposite directions under different beams, indicating that it does not maintain the observation distribution and information loss patterns of a real low-resolution LiDAR. This is because the downsampling only changes the sparsity of points and cannot characterize the system differences in the beam vertical angle distribution, transmit/receive optical characteristics, and structural missing measurements caused by occlusion in a real low-resolution LiDAR. Therefore, simple downsampling from high-beam point clouds cannot reliably reproduce the detection performance of real low-beam LiDAR and is not suitable as a simulation method for real low-resolution sensors.

\begin{figure}[t]
  \centering
  \begin{subfigure}[t]{0.3\linewidth}
    \centering
    \includegraphics[width=\linewidth]{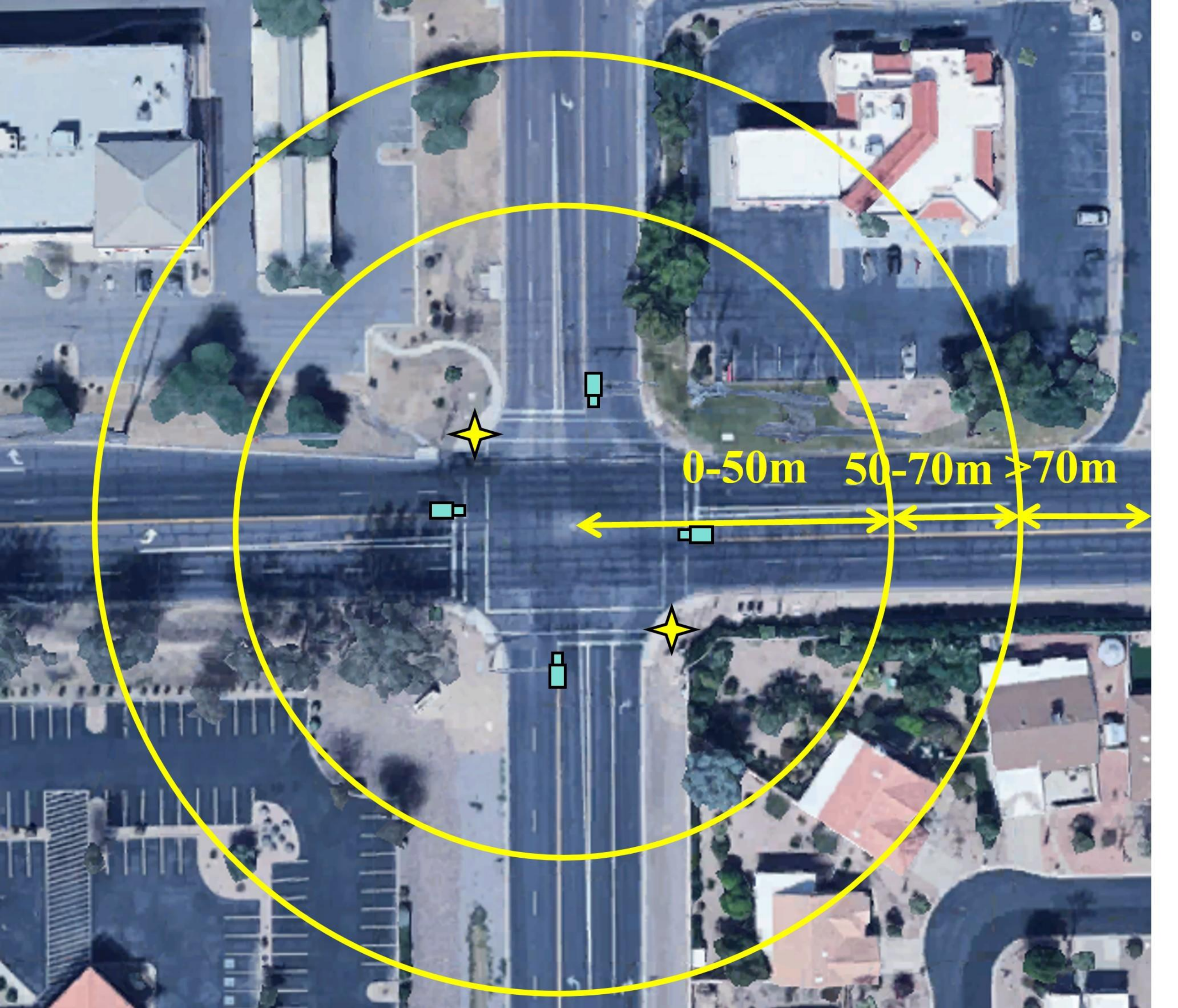}
    \caption{Real-world distances}
  \end{subfigure}
  \begin{subfigure}[t]{0.3\linewidth}
    \centering
    \includegraphics[width=\linewidth]{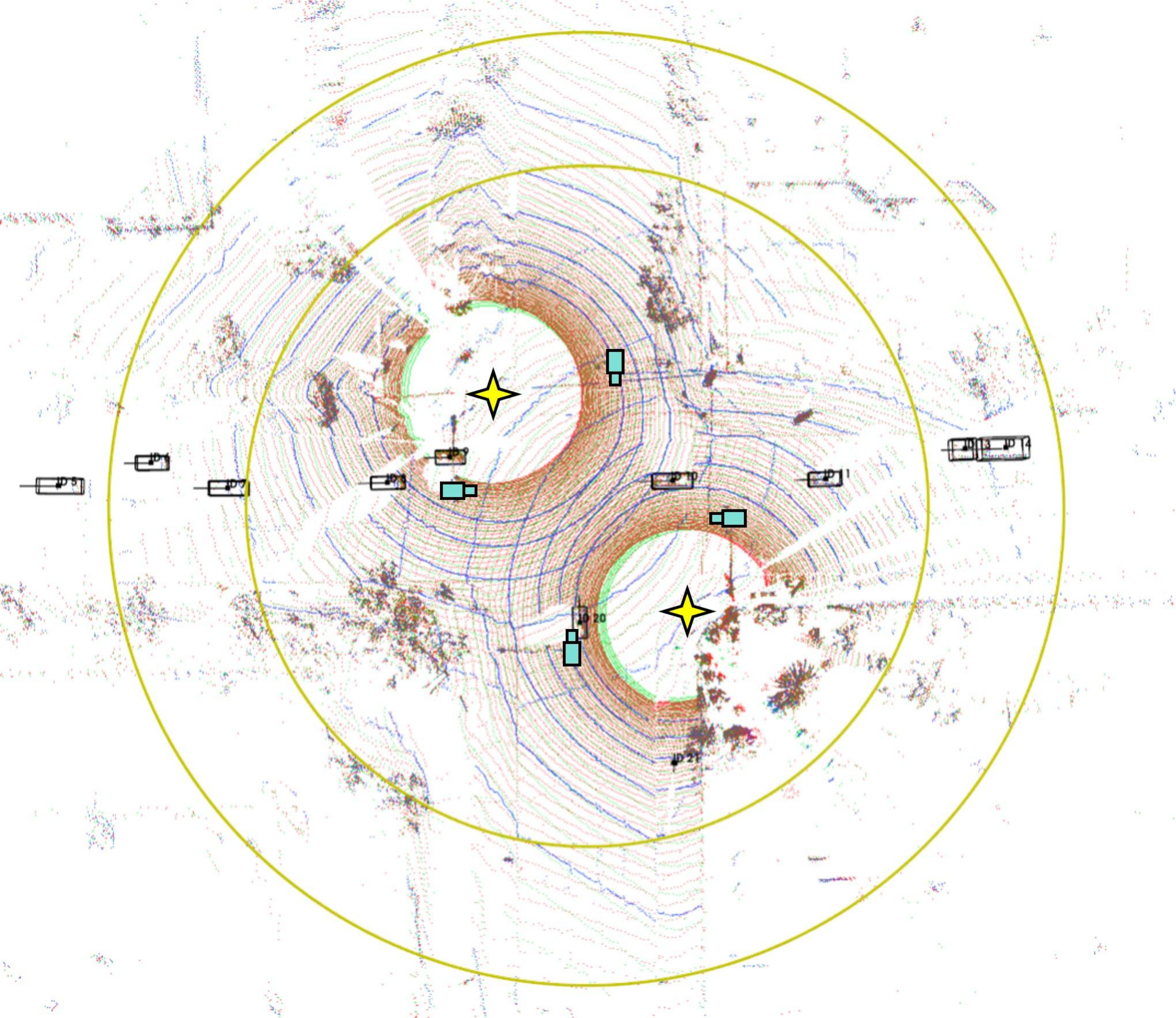}
    \caption{RESOLVE distances}
  \end{subfigure}
  
  \begin{subfigure}[t]{0.3\linewidth}
    \centering
    \includegraphics[width=\linewidth,height=1.9cm]{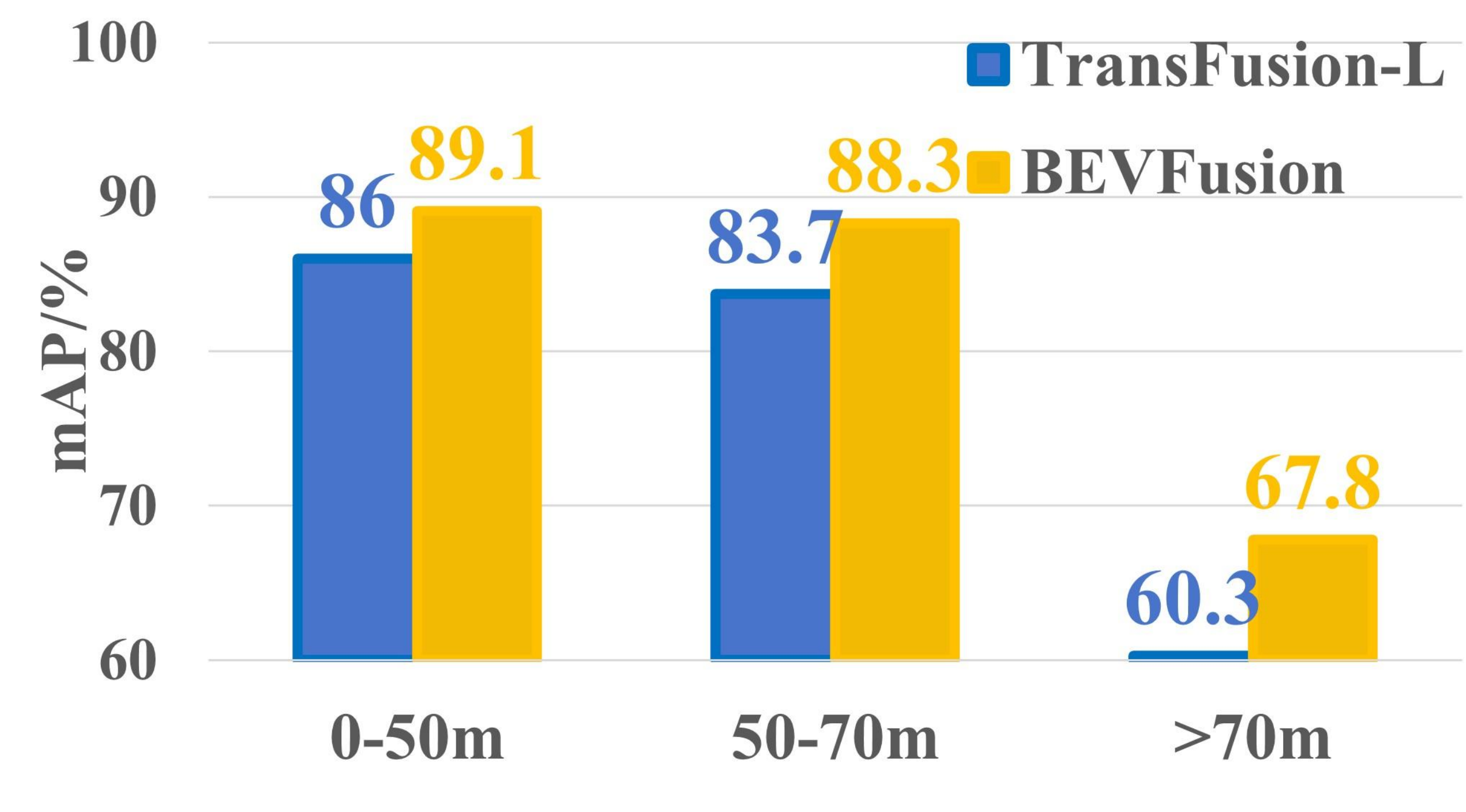}
    \caption{Under low resolution}
  \end{subfigure}
  \begin{subfigure}[t]{0.3\linewidth}
    \centering
    \includegraphics[width=\linewidth,height=1.9cm]{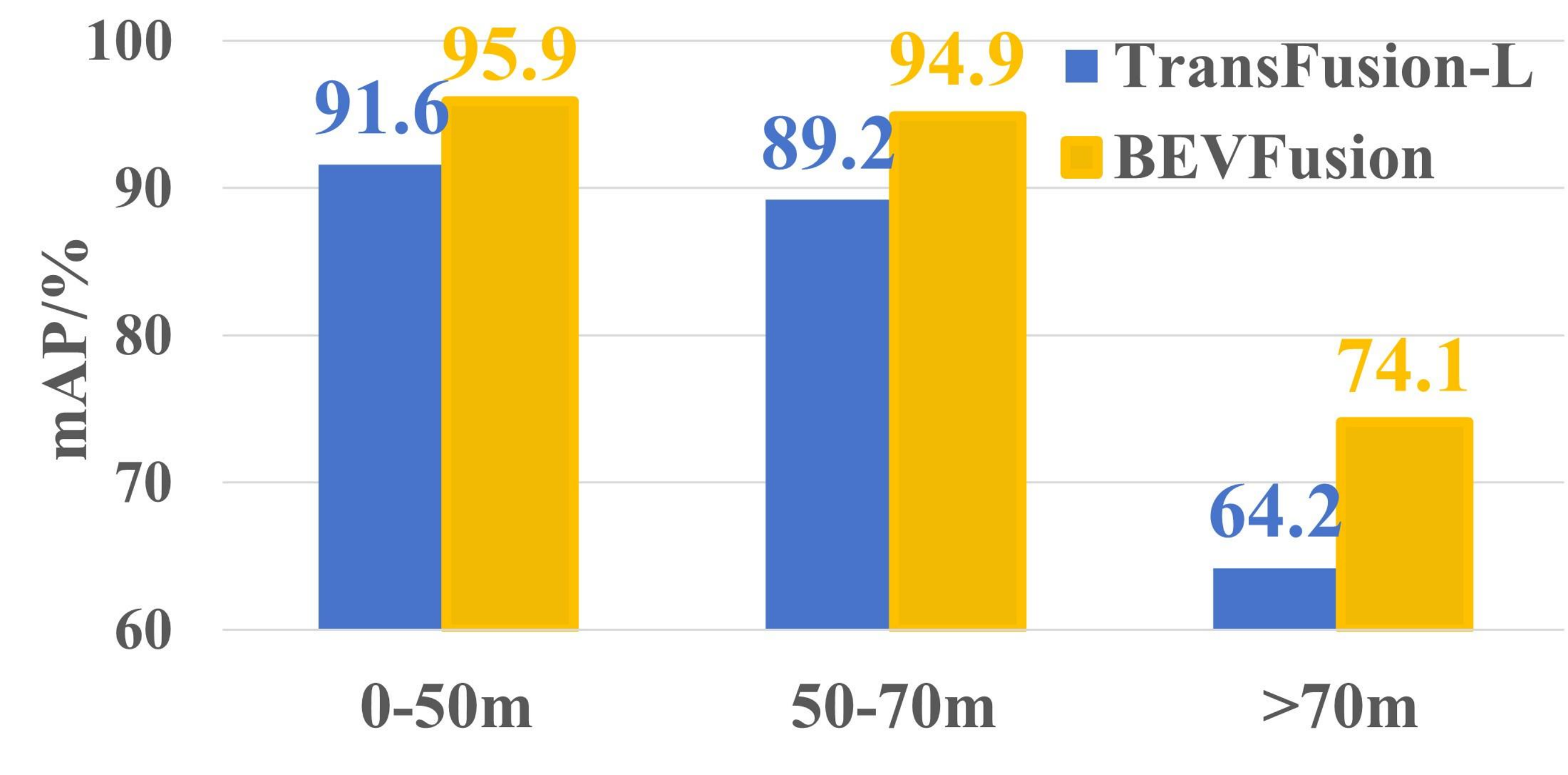}
    \caption{Under mid resolution}
  \end{subfigure}
  \begin{subfigure}[t]{0.3\linewidth}
    \centering
    \includegraphics[width=\linewidth,height=1.9cm]{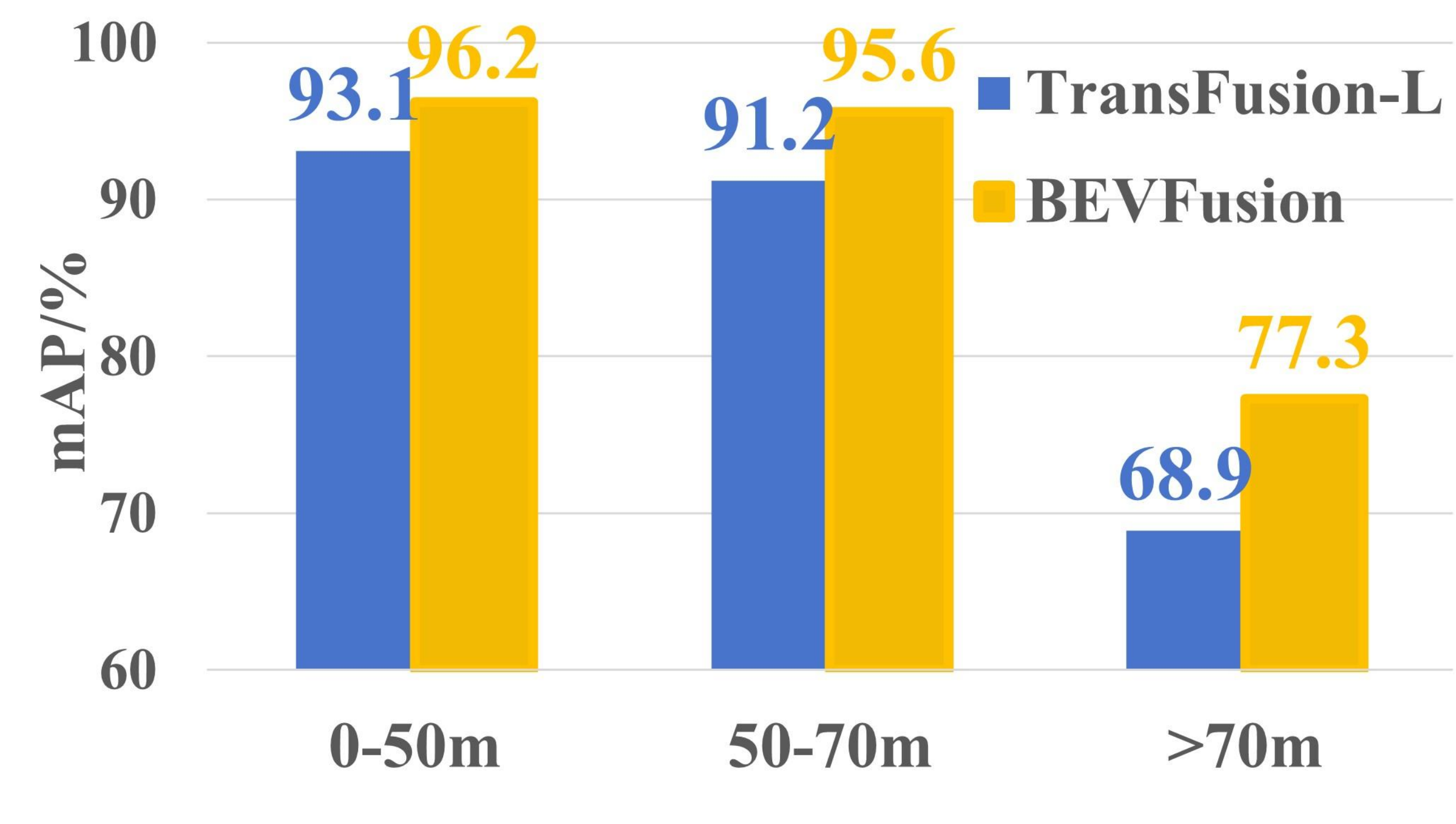}
    \caption{Under high resolution}
  \end{subfigure}
  \caption{Effects of LiDAR resolution on object detection accuracy at different distances. }
  \label{fig:dist_res_comp}
\end{figure}

\subsubsection{Distance-conditioned multimodal perception gain.} We further explore how multimodal perception gains over unimodal vary with object distance. In roadside sensing, LiDAR point density naturally decreases as object distance increases, leading to progressively sparser measurements for distant objects. As show in~\cref{fig:dist_res_comp}, the detection performance of both models decreases with increasing distance, with the most significant drop occurring after 70\,m. Compared to the unimodal model with a decrease of 37\% at long distances, the multimodal model only decreases by 20\%, indicating that fusing visual information can provide supplementary semantics when point clouds are sparse at long distances, thereby improving robustness. 

\begin{figure}[t]
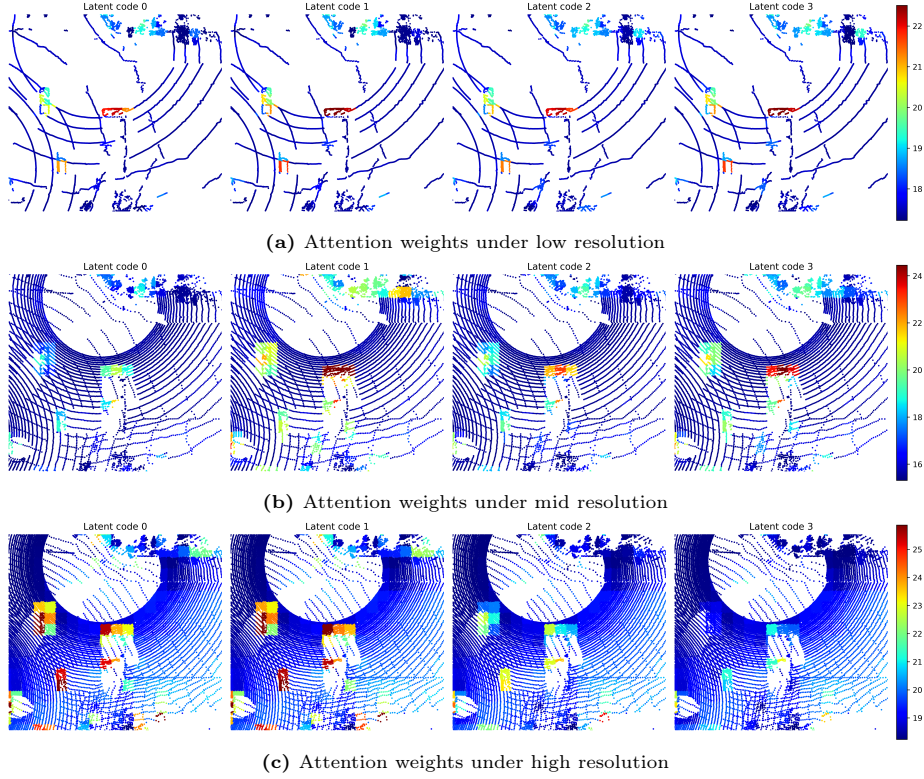

  \centering
  \begin{subfigure}[t]{\linewidth}
    \centering
    \includegraphics[width=\linewidth]{supplement_pictures/low_attention.pdf}
    \caption{Attention weights under low resolution}
  \end{subfigure}
  
  \begin{subfigure}[t]{\linewidth}
    \centering
    \includegraphics[width=\linewidth]{supplement_pictures/mid_attention.pdf}
    \caption{Attention weights under mid resolution}
  \end{subfigure}
  
  \begin{subfigure}[t]{\linewidth}
    \centering
    \includegraphics[width=\linewidth]{supplement_pictures/high_attention.pdf}
    \caption{Attention weights under high resolution}
  \end{subfigure}
  \caption{Comparison of spatial attention maps induced by different latent codes in the last VSA module of VoxSet~\cite{vst} architecture across three LiDAR resolutions.}
  \label{fig:attention_weights}
\end{figure}

\subsubsection{Attention Weights.} \Cref{fig:attention_weights} shows the spatial attention maps of the four latent codes in the last VSA module at three resolutions. As resolution increases, the voxel set input becomes denser, and the attention response evolves from being relatively concentrated on the target center at low resolution to a more refined and structured distribution at high resolution, while simultaneously suppressing background scan lines and non-target structures more stably. VoxSet~\cite{vst} consistently highlights the target region at different resolutions, demonstrating the robustness of its attention mechanism to resolution variations. Furthermore, the different latent codes exhibit clearer complementarity at high resolution, such as focusing on different parts of the target and its surrounding environment, further demonstrating the VSA's characterization capabilities under richer observation conditions.

\section{Visualization}
\subsection{3D Object Detection} 
\begin{figure}[t]
  \centering
  \begin{subfigure}[t]{\linewidth}
    \centering
    \includegraphics[width=\linewidth]{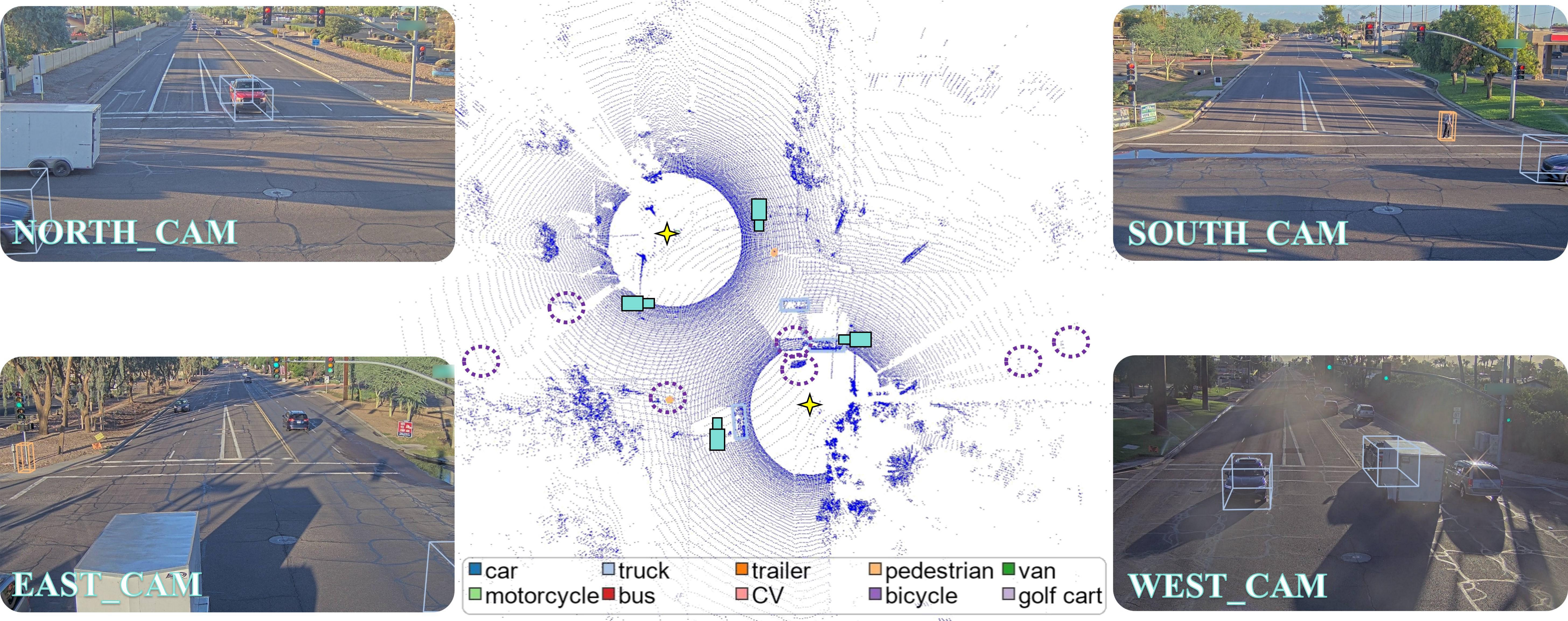}
    \caption{Low-resolution detection result}
    \label{fig:low_det}
  \end{subfigure}
  
  \begin{subfigure}[t]{\linewidth}
    \centering
    \includegraphics[width=\linewidth]{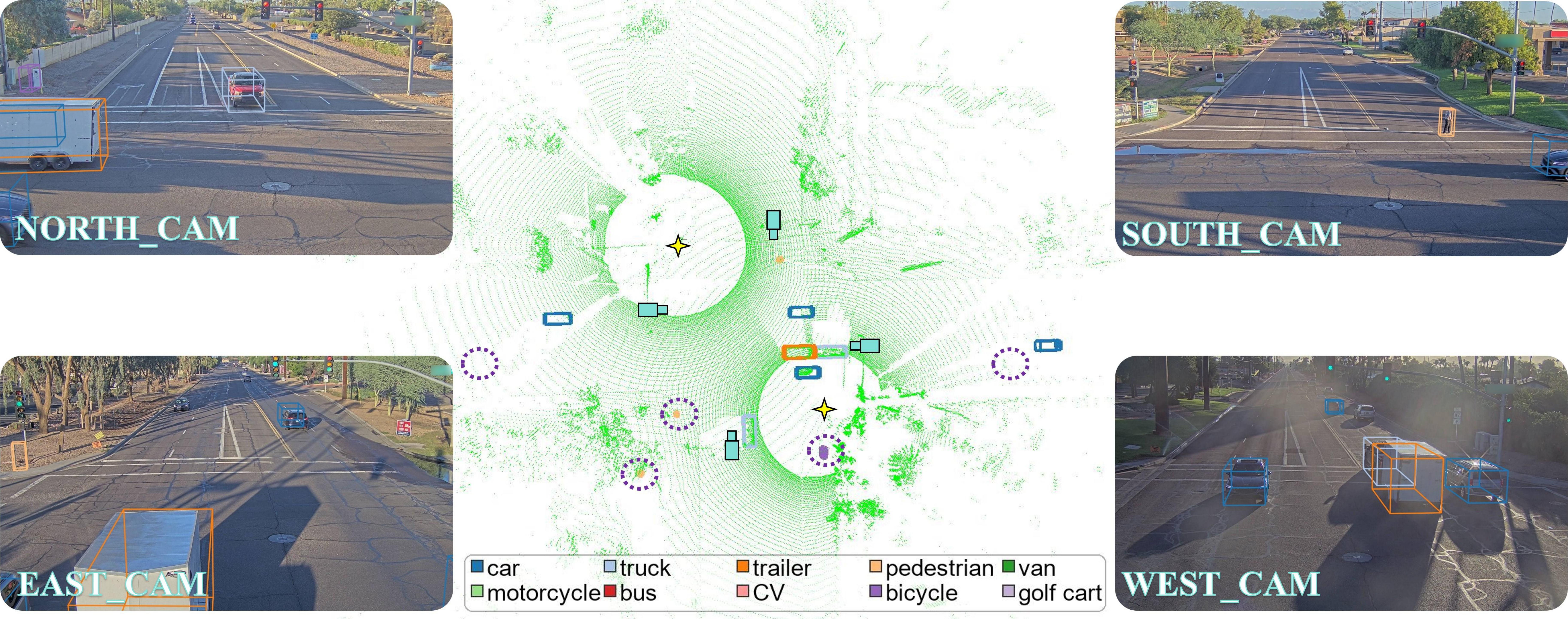}
    \caption{Mid-resolution detection result}
    \label{fig:mid_det}
  \end{subfigure}
  
  \begin{subfigure}[t]{\linewidth}
    \centering
    \includegraphics[width=\linewidth]{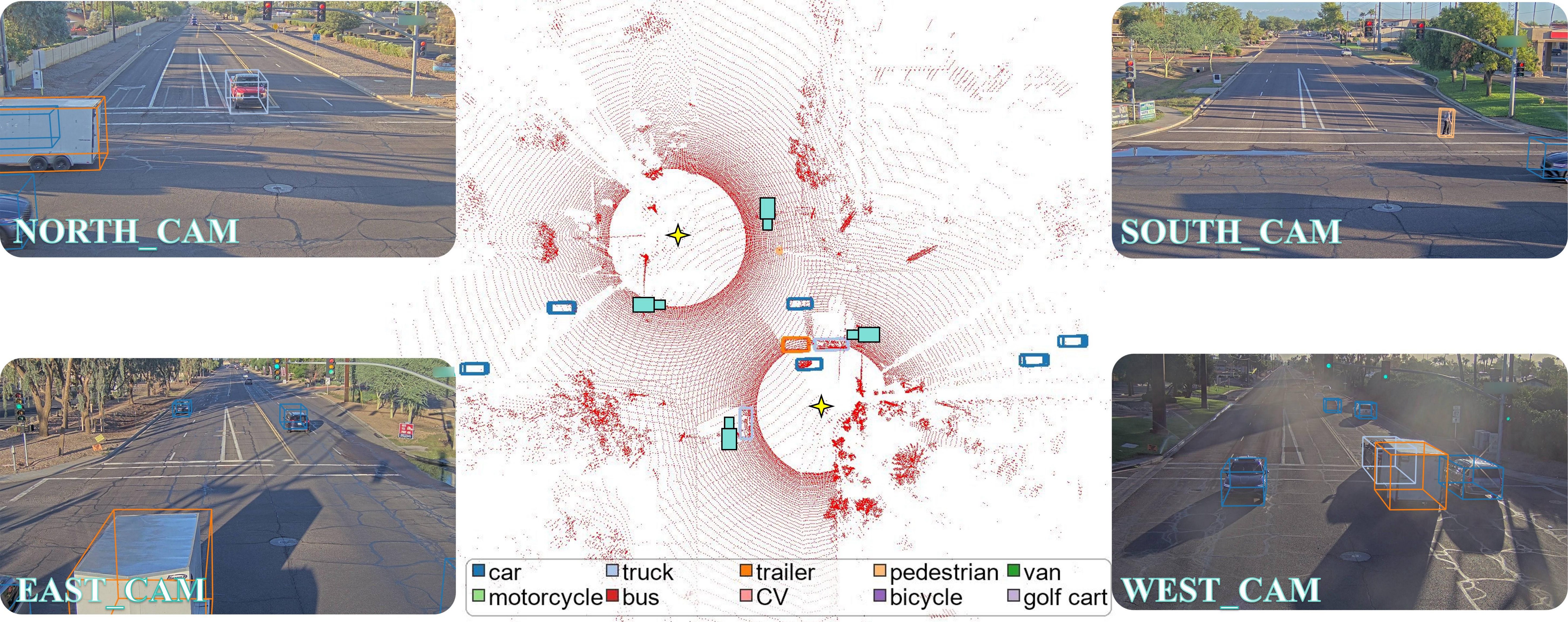}
    \caption{High-resolution detection result}
    \label{fig:high_det}
  \end{subfigure}
  \caption{Qualitative detection results of BEVFusion~\cite{bevfusion} in a sunny scene across three resolution settings, where missed or falsely detected objects are marked with purple circles.}
  \label{fig:det_results}
\end{figure}

\begin{figure}[t]
  \centering
  \begin{subfigure}[t]{\linewidth}
    \centering
    \includegraphics[width=\linewidth]{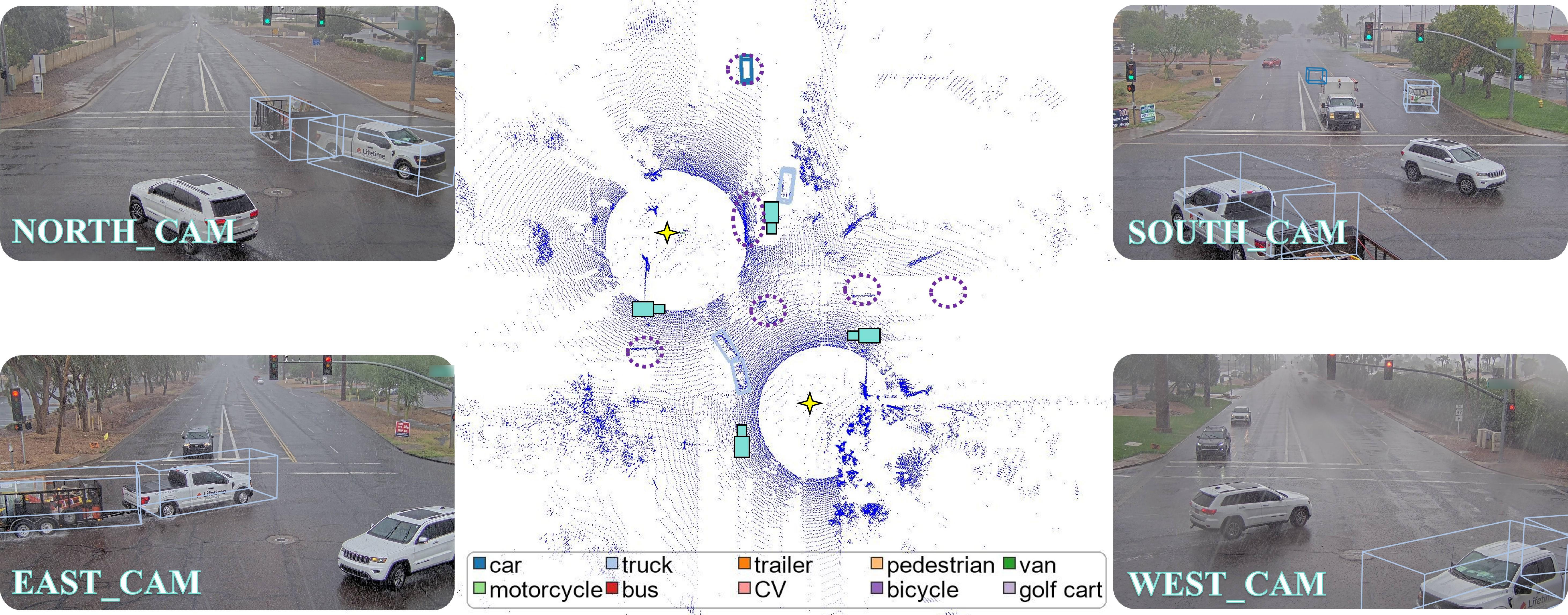}
    \caption{Low-resolution detection result}
    \label{fig:low_det_rain}
  \end{subfigure}
  
  \begin{subfigure}[t]{\linewidth}
    \centering
    \includegraphics[width=\linewidth]{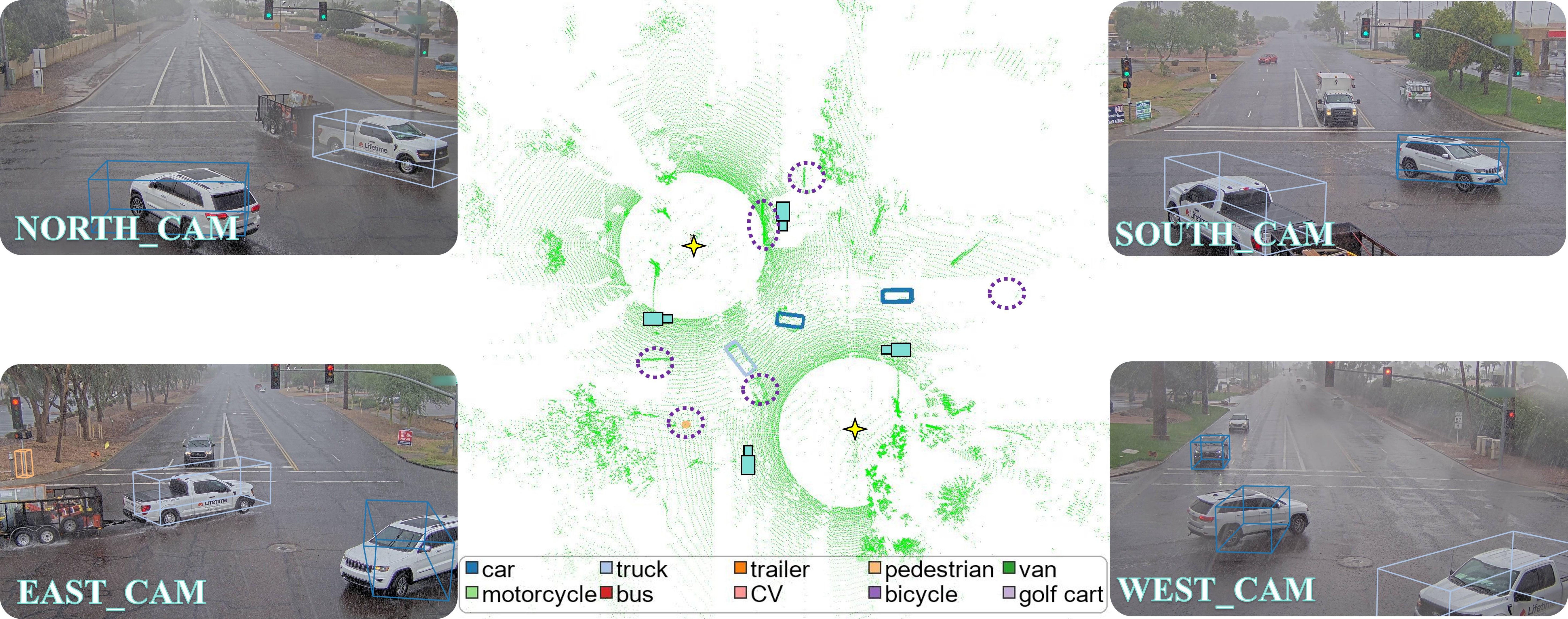}
    \caption{Mid-resolution detection result}
    \label{fig:mid_det_rain}
  \end{subfigure}
  
  \begin{subfigure}[t]{\linewidth}
    \centering
    \includegraphics[width=\linewidth]{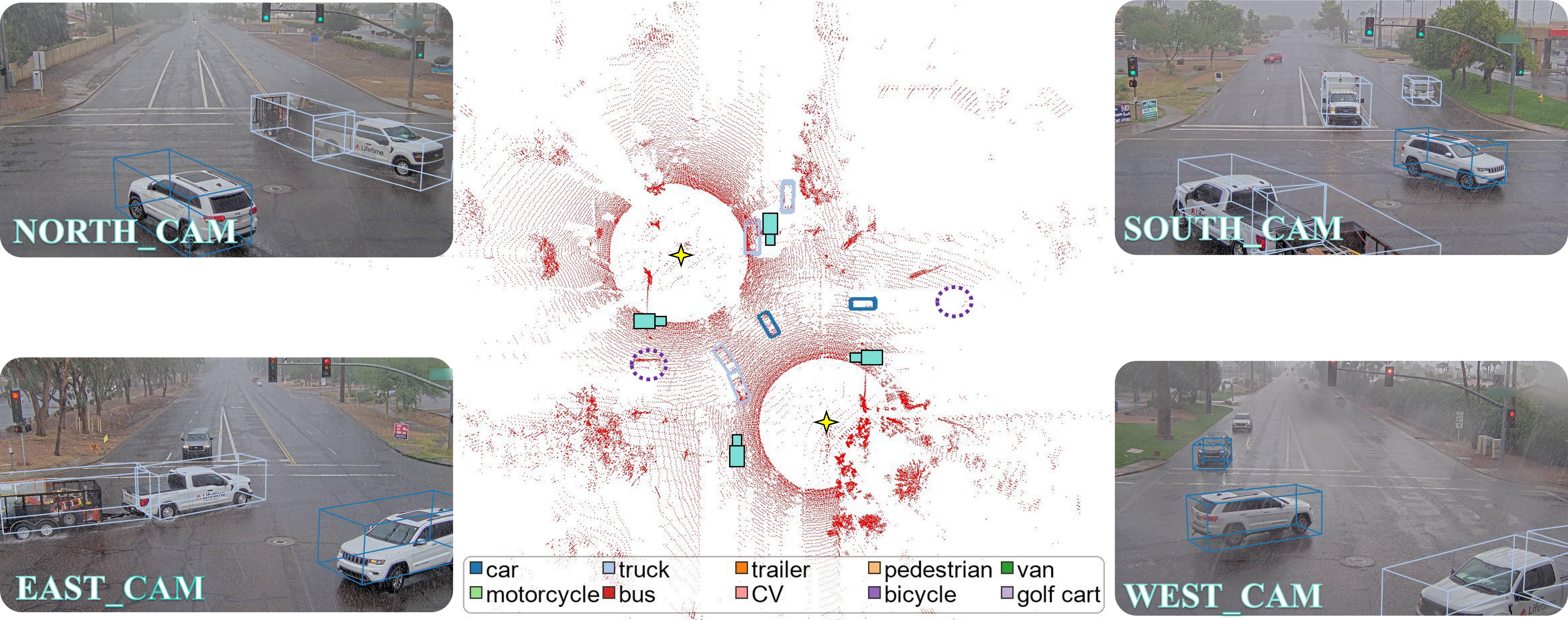}
    \caption{High-resolution detection result}
    \label{fig:high_det_rain}
  \end{subfigure}
  \caption{Qualitative detection results of TransFusion-L~\cite{transfusion} in a rainy scene across three resolution settings, where missed or falsely detected objects are marked with purple circles.}
  \label{fig:det_results_rain}
\end{figure}

\begin{figure}[t]
  \centering
  \begin{subfigure}[t]{\linewidth}
    \centering
    \includegraphics[width=\linewidth]{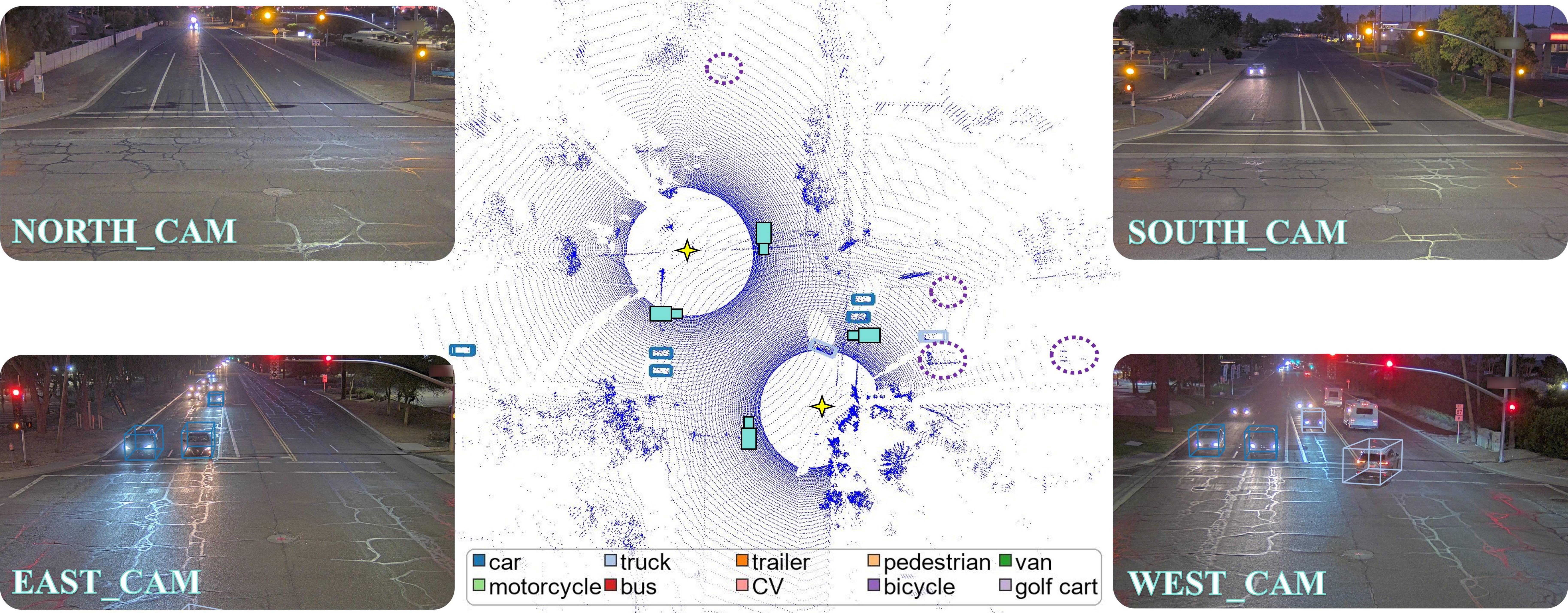}
    \caption{Low-resolution detection result}
    \label{fig:low_det_night}
  \end{subfigure}
  
  \begin{subfigure}[t]{\linewidth}
    \centering
    \includegraphics[width=\linewidth]{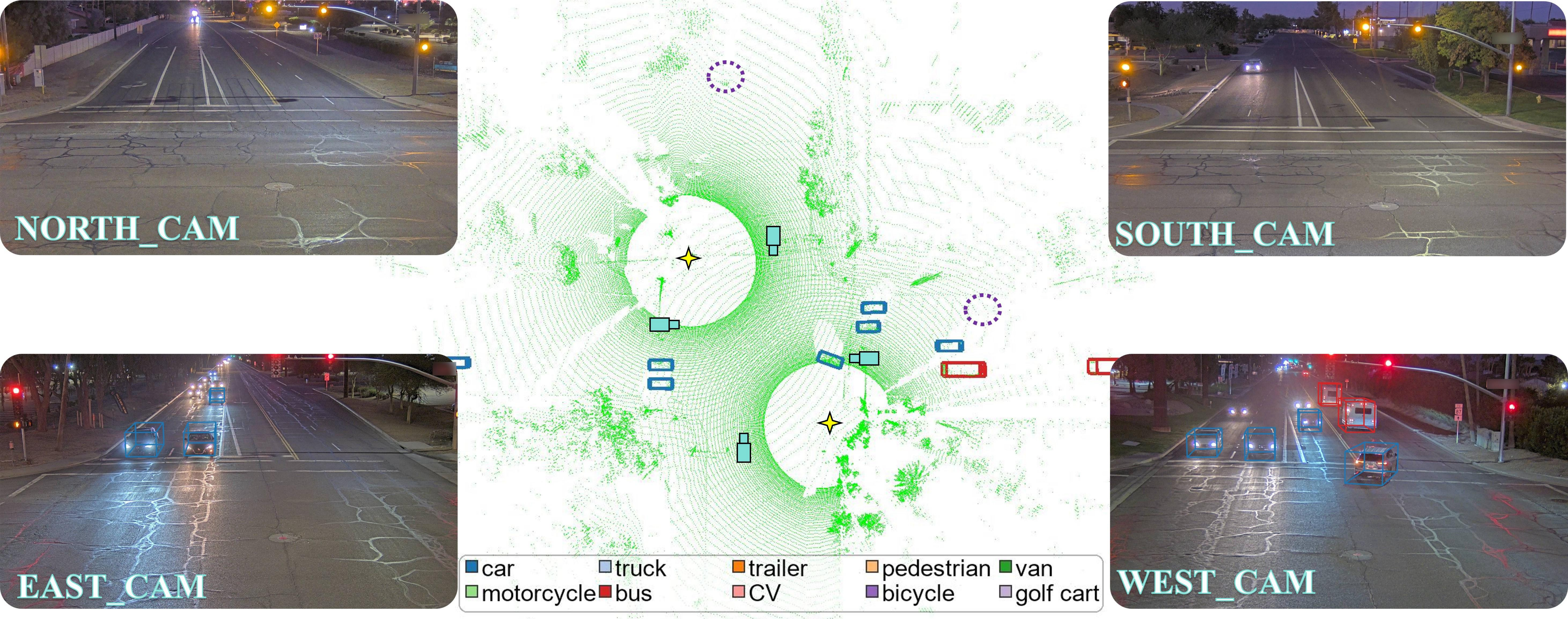}
    \caption{Mid-resolution detection result}
    \label{fig:mid_det_night}
  \end{subfigure}
  
  \begin{subfigure}[t]{\linewidth}
    \centering
    \includegraphics[width=\linewidth]{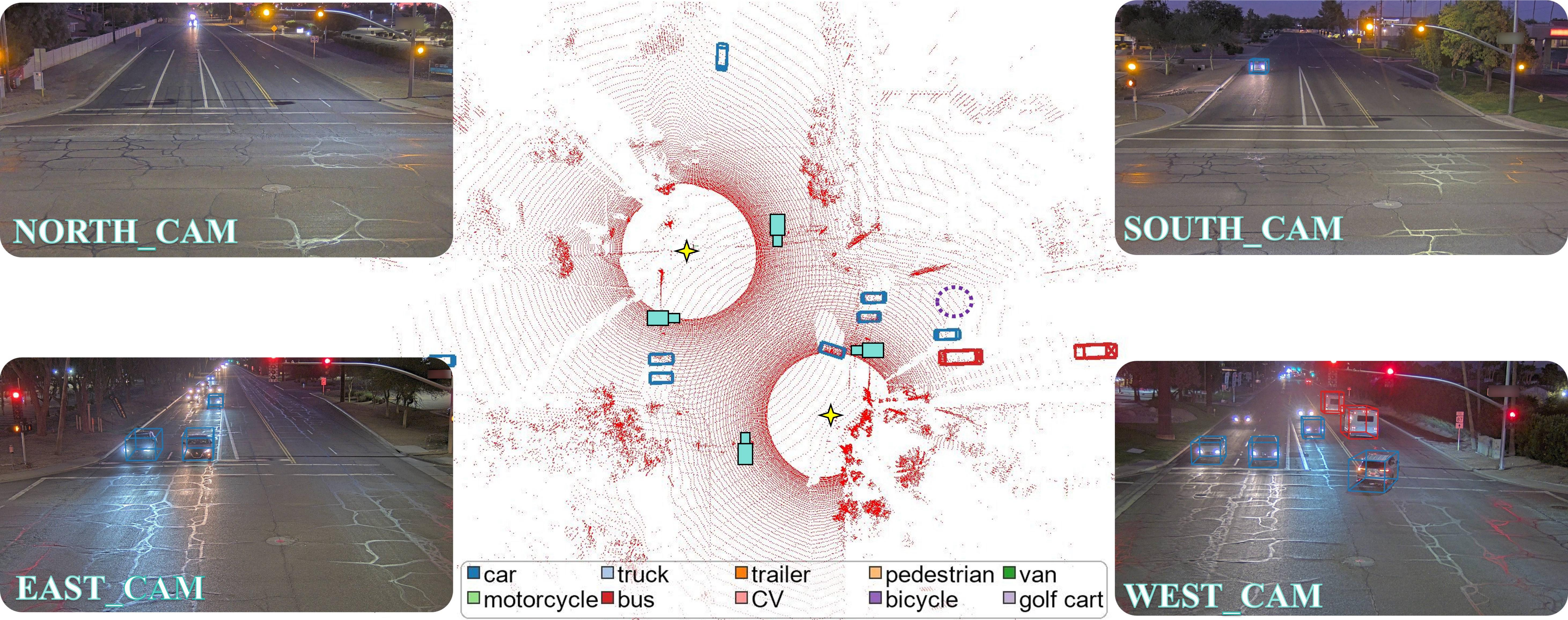}
    \caption{High-resolution detection result}
    \label{fig:high_det_night}
  \end{subfigure}
  \caption{Qualitative detection results of LION~\cite{lion} in a night scene across three resolution settings, where missed or falsely detected objects are marked with purple circles.}
  \label{fig:det_results_night}
\end{figure}

As shown in~\cref{fig:det_results,fig:det_results_rain,fig:det_results_night}, we visualize and compare the detection results of three typical scenarios based on different detection models on our RESOLVE dataset. In sunny scenarios (see~\cref{fig:det_results}), high-resolution models typically exhibit higher detection accuracy and lower false positive rates, showing a significant advantage in detecting distant targets. In contrast, low-resolution models often suffer from limited detection performance due to sparser point clouds at long distances, and are more prone to false positives or class confusion at close range due to insufficient model representation capabilities. In rainy scenarios (see~\cref{fig:det_results_rain}), reflection and scattering caused by raindrops significantly weaken the laser echo intensity and compress the effective detection range, leading to a decline in overall detection performance. Even with high-resolution LiDAR, target misses may still occur at close range, while the performance degradation of low-resolution LiDAR is more significant. In nighttime scenarios (see~\cref{fig:det_results_night}), LiDAR is insensitive to changes in illumination, therefore its detection results are generally more stable than camera models that rely on visible light imaging.

\subsection{Multi-object Tracking}  
\begin{figure}[t]
  \centering
   \includegraphics[width=\linewidth]{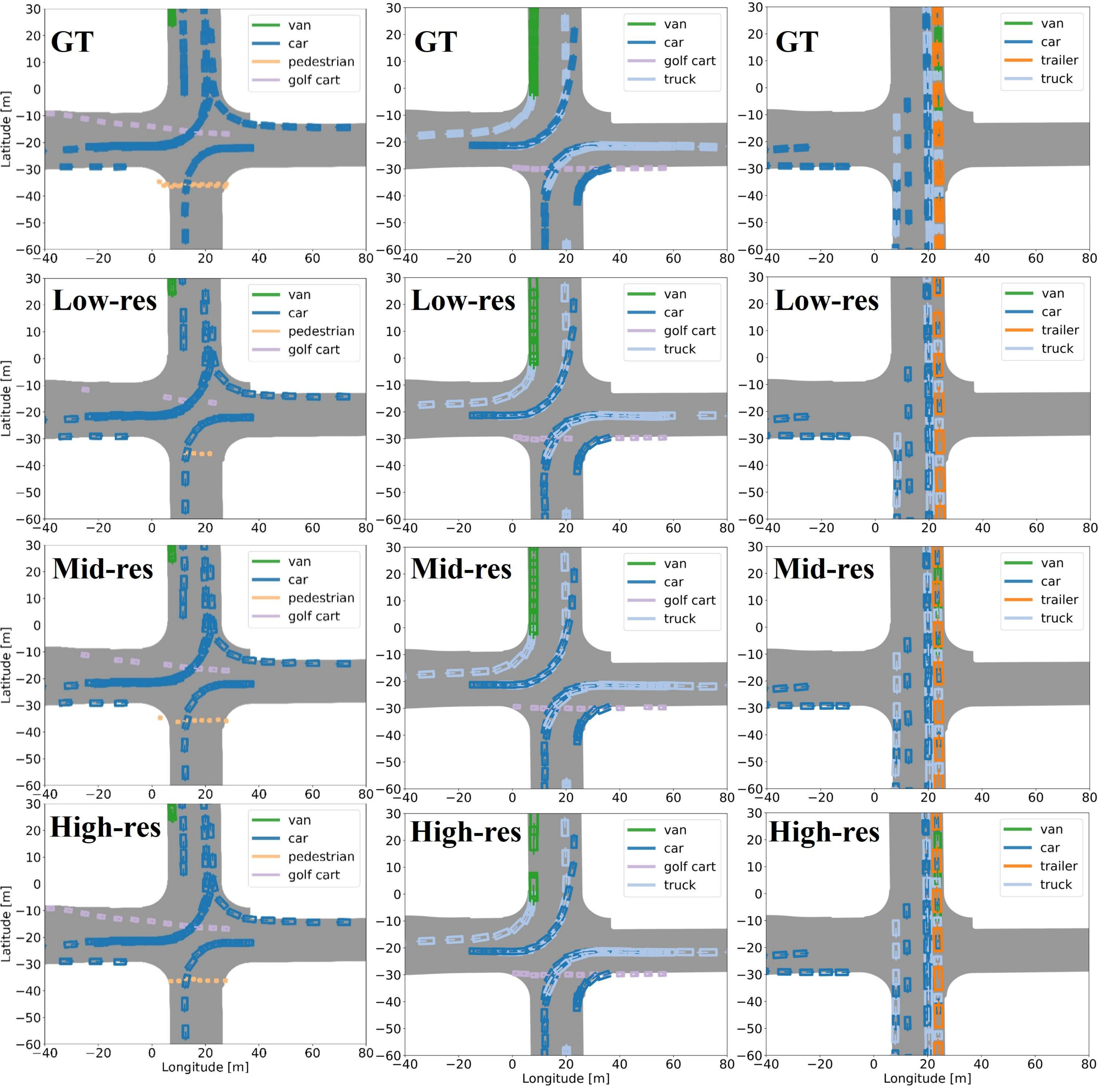}
  \caption{Comparison of tracking results under different detection resolutions across three scenarios. We use Voxel Mamba~\cite{voxelmamba} and SimpleTrack~\cite{simpletrack} as the detector and tracker.}
  \label{fig:track_results_detailer}
\end{figure}

We illustrate the tracking results based on detection at different resolutions in three scenarios in~\cref{fig:track_results_detailer}. Overall, low-resolution detection leads to a higher false negative rate and localization noise, resulting in trajectory fragmentation and exacerbating the uncertainty of data association in areas with dense targets. With increased resolution, the overall trajectory becomes more continuous and smoother. In scenario 1 (first column of~\cref{fig:track_results_detailer}), low resolution has the most significant impact on small targets. The trajectories of pedestrians and golf carts become noticeably discontinuous and local positioning errors increase. High resolution restores continuous tracking of small targets to some extent, but still suffers from insufficient separability when two pedestrians are moving in parallel and approaching each other, making it difficult to stably distinguish the trajectories of two parallel pedestrians. In scenario 2 (second column of~\cref{fig:track_results_detailer}), medium to low resolution struggles to form a complete golf cart trajectory. High resolution improves this, but the van still exhibits some moments of mismatch. In scenario 3 (third column of~\cref{fig:track_results_detailer}), low resolution results in a larger localization deviation, and the tracking start time for the high-speed truck is relatively delayed.

\end{document}